\PassOptionsToPackage{table}{xcolor}
\documentclass[a4, 10pt]{scrartcl}
\usepackage[numbers]{natbib}

\usepackage[utf8]{inputenc}
\usepackage[english]{babel}
\usepackage{cmap}
\usepackage[T1]{fontenc}

\usepackage[
pdfa,
hidelinks,
pdftex, 
pdfdisplaydoctitle,
pdfpagelabels,
pdfauthor={Nils M. Kriege, Christopher Morris, Fredrik D. Johansson},
pdftitle={Survey on Graph Kernels},
pdfsubject={},
pdfkeywords={graph kernels, graph classification, survey},
pdfproducer={Latex with the hyperref package},
pdfcreator={pdflatex}
]{hyperref}

\usepackage{exscale}
\usepackage[l2tabu, orthodox]{nag}

\usepackage{rotating}
\usepackage{graphicx}
\usepackage{booktabs}
\usepackage{url}
\usepackage{todonotes}
\usepackage{multirow}
\usepackage[auth-lg]{authblk}

\usepackage{amsmath}
\usepackage{amssymb}
\usepackage{amsthm}
\usepackage{amsfonts}
\usepackage{thmtools}
\usepackage{mleftright}
\usepackage{dsfont}

\usepackage{thm-restate}
\usepackage[mathic=true]{mathtools}
\usepackage{fixmath}
\usepackage{units}

\usepackage{paralist}
\usepackage{enumitem}
\setlist[enumerate]{noitemsep, topsep=0.5\topsep}
\setlist[description]{noitemsep, topsep=0.5\topsep}
\setlist[itemize]{noitemsep, topsep=0.5\topsep}

\theoremstyle{plain}
\newtheorem{theorem}{Theorem}[section]

\theoremstyle{definition}
\newtheorem{definition}[theorem]{Definition}

\usepackage{microtype}
\usepackage{ellipsis}
\usepackage{xspace}
\usepackage{hfoldsty}

\usepackage[draft]{fixme}
\fxsetup{inline,nomargin,theme=color}
\usepackage{lmodern}
\usepackage{algorithm}
\usepackage{algorithmicx}

\usepackage{tikz}
\usepackage{todonotes}

\makeatletter
\def\thmt@refnamewithcomma #1#2#3,#4,#5\@nil{%
	\@xa\def\csname\thmt@envname #1utorefname\endcsname{#3}%
	\ifcsname #2refname\endcsname
	\csname #2refname\expandafter\endcsname\expandafter{\thmt@envname}{#3}{#4}%
	\fi
}
\makeatother
\usepackage[capitalise,noabbrev]{cleveref}

\DeclareSymbolFont{usualmathcal}{OMS}{cmsy}{m}{n}
\DeclareSymbolFontAlphabet{\mathcal}{usualmathcal}

\DeclarePairedDelimiter{\norm}{\lVert}{\rVert}
\newcommand{\ra}[1]{\renewcommand{\arraystretch}{#1}}

\newcommand{\new}[1]{\emph{#1}}

\newcommand{\cH}{\ensuremath{{\mathcal H}}\xspace}

\newcommand{\cO}{\ensuremath{{\mathcal O}}\xspace}

\newcommand{\bbR}{\ensuremath{\mathbb{R}}}

\newcommand{\bbN}{\ensuremath{\mathbb{N}}}

\newcommand{\cGI}[0]{\ensuremath{\mathsf{GI}}\xspace}

\newcommand{\RR}{\mathbb{R}}

\newcommand{\GG}{\mathbb{G}}

\usepackage{bm}
\let\vec\bm

\providecommand{\varitem}{}
\makeatletter
\newenvironment{axioms}[1]
 {\renewcommand\varitem[1]{\item[\textbf{#1\arabic{enumi}\rlap{$##1$}}]%
    \edef\@currentlabel{#1\arabic{enumi}{$##1$}}}%
  \enumerate[topsep=.3em,parsep=0pt,itemsep=.3em,leftmargin=2em+\widthof{#1},label=\normalfont\textbf{#1\arabic*}, ref=#1\arabic*]}
 {\endenumerate}
\makeatother

\addtokomafont{disposition}{\rmfamily}
\addtokomafont{descriptionlabel}{\rmfamily}

\begin{document}

		\title{A Survey on Graph Kernels}

		\author[1]{Nils M.~Kriege}
		\author[2]{Fredrik D.~Johansson}
		\author[1]{Christopher Morris}
		\affil[1]{Department of Computer Science \authorcr 
		TU Dortmund University, Dortmund, Germany \authorcr
      \texttt{\{nils.kriege,christopher.morris\}@tu-dortmund.de}}
		\affil[2]{Institute for Medical Engineering and Science, MIT \authorcr
		\texttt{fredrikj@mit.edu}}
		
		\date{\vspace{-40pt}}

\maketitle

		\begin{abstract}
			Graph kernels have become an established and widely-used technique for solving classification tasks on graphs.
			This survey gives a comprehensive overview of techniques for kernel-based graph classification developed in the past 15 years.
			We describe and categorize graph kernels based on properties inherent to their design, such as the nature of their extracted graph features, their method of computation and their applicability to problems in practice.
			In an extensive experimental evaluation, we study the classification accuracy of a large suite of graph kernels on established benchmarks as well as new datasets.
			We compare the performance of popular kernels with several baseline methods and study the effect of applying a Gaussian RBF kernel to the
			metric induced by a graph kernel. In doing so, we find that simple baselines become competitive after this transformation on some datasets.
			Moreover, we study the extent to which existing graph kernels agree in their predictions (and prediction errors) and obtain a data-driven categorization of kernels as result.
			Finally, based on our experimental results, we derive a practitioner's guide to kernel-based
			graph classification.
		\end{abstract}

\section{Introduction}
\label{sec:introduction}
Machine learning analysis of large, complex datasets has become an integral part of research in both the natural and social sciences. Largely, this development was driven by the empirical success of supervised learning of  vector-valued data or image data.
However, in many domains, such as chemo- and bioinformatics, social network analysis or computer vision, observations describe relations between objects or individuals and cannot be interpreted as vectors or fixed grids; instead, they are naturally represented by graphs.
This poses a particular challenge in the application of traditional data mining
and machine learning approaches.
In order to learn successfully from such data, it is necessary for algorithms to exploit the rich
information inherent to the graphs' structure and annotations associated with
their vertices and edges.

A popular approach to learning with graph-structured data is to make use of graph kernels---functions which
measure the similarity between graphs---plugged into a kernel machine, such as a support
vector machine.
Due to the prevalence of graph-structured data and the empirical success of kernel-based methods for classification, a large body of work in this area exists. In particular, in the past 15 years, numerous graph kernels have been proposed, motivated either by their theoretical properties or by their suitability and specialization to particular application domains.
Despite this, there are no review articles aimed at comprehensive comparison between different graph kernels nor at giving practical guidelines for choosing between them. As the number of methods grow, it is becoming increasingly difficult for both non-expert practitioners and researchers new to the field to identify an appropriate set of candidate kernels for their application.

This survey is intended to give an overview of the graph kernel literature,
targeted at the active researcher as well as the practitioner.
First, we describe and categorize graph kernels according to their design paradigm, the used graph features and their method of computation. We discuss theoretical approaches to measure the expressivity of graph kernels and their applicability to problems in practice.
Second, we perform an extensive experimental evaluation of state-of-the-art graph kernels on a
wide range of benchmark datasets for \emph{graph classification} stemming from chemo- and bioinformatics as well
as social network analysis and computer vision. Finally, we provide guidelines for the practitioner for the
successful application of graph kernels.

\subsection{Contributions}
\label{sec:contributions}

We summarize our contributions below.

\begin{itemize}
	\item We give a comprehensive overview of the graph kernel literature, categorizing kernels according to several properties. Primarily, we distinguish graph kernels by their mathematical definition and which graph features they use to measure similarity.
	      Moreover, we discuss whether kernels are applicable to
	      \begin{inparaenum}[(i)]
	      	\item graphs annotated with continuous attributes, or
	      	\item discrete labels, or
	      	\item unlabeled graphs only.
	      \end{inparaenum}
	      Additionally, we describe which kernels rely on the \emph{kernel trick} as opposed to being computed from feature vectors and what effects this
	      has on the running time and flexibility. 
	\item We give an overview of applications of graph kernels in different domains
	      and review theoretical work on the expressive power of graph kernels.
	\item We compare state-of-the-art graph kernels in an extensive experimental
	      study across a wide range of established and new benchmark datasets.
	      Specifically, we show the strengths and weaknesses of
	      the individual kernels or classes of kernels for specific datasets.
	      \begin{itemize}
	        \item We compare popular kernels to simple baseline methods in order to assess the need for more
	              sophisticated methods which are able to take more structural features into account.
	              To this end, we analyze the ability of graph kernels to distinguish the graphs
	              in common benchmark datasets.
	        \item Moreover, we investigate the effect of combining a  Gaussian RBF kernel  with the metric induced by
	              a graph kernel in order to learn non-linear decision boundaries in the feature space of the graph kernel. We observe that with this approach simple baseline methods become competitive to state-of-the-art kernels for some datasets, but fail for others.
	        \item We study the similarity between graph kernels in terms of their classification predictions and errors on
                 graphs from the chosen datasets. This analysis provides a qualitative, data-driven means of assessing the similarity of different kernels in terms of which graphs they deem similar.
	      \end{itemize}
	\item Finally, we provide guidelines for the practitioner and new researcher for the successful application of graph kernels.
\end{itemize}

\subsection{Related Work}
\label{sec:related}
The most recent surveys of graph kernels are the works of~\citet{Gho+2018} and~\citet{Zha+2018b}. \citet{Gho+2018} place a strong emphasis on covering the fundamentals of kernel methods in general and summarizing known experimental results for graph kernels. The article does not, however, cover the most recent contributions to the literature. Most importantly, the article does not provide a detailed experimental study comparing the discussed kernels. That is, the authors do not perform (nor reproduce) original experiments on graph classification and solely report numbers found in the corresponding original paper.
The survey by \citet{Zha+2018b} focuses on kernels for graphs without attributes which is a small subset of the scope of this survey. Moreover, it does not discuss the most recent developments in this area. Another survey was published in 2010 by \citet{Vis+2010} but its main topic are random walk kernels and it does not include recent advances. Moreover, various PhD theses give (incomplete or dated) overviews, see, e.g., \cite{Bor+2007,Kri2015,Neu+2015,She+2012}.
None of the papers provides compact guidelines for choosing a kernel for a particular dataset.

Compared to the existing surveys, we provide a more complete overview covering a larger number of kernels, categorizing them according to their design, the extracted graph features and their computational properties. The validity of comparing results from different papers depends on whether these were obtained using comparable experimental setups (e.g., choices for hyperparameters, number of folds used for cross-validation, etc.), which is not the case across the entire spectrum of the graph kernel literature. Hence, we conducted an extensive experimental evaluation comparing a large number of graph kernels and datasets going beyond comparing kernels just by their classification accuracy. Another unique contribution of this article is a practitioner's guide for choosing between graph kernels.

\subsection{Outline}

In~\cref{fun}, we introduce notation and provide mathematical definitions necessary to understand the rest of the paper. \cref{main} gives an overview of the graph kernel literature. We start off by introducing kernels based on neighborhood aggregation techniques. Subsequently, we describe kernels based on assignments, substructures, walks and paths, and neural networks, as well as  approaches that do not fit into any of the former categories. In~\cref{sec:theory}, we survey theoretical work on the expressivity of kernels and in~\cref{sec:applications} we describe applications of graph kernels in four domain areas. Finally, in~\cref{exp} we introduce and analyze the results of a large-scale experimental study of graph kernels in classification problems, and provide guidelines for the successful application of graph kernels.

\section{Fundamentals}\label{fun}

In this section, we cover notation and definitions of fundamental concepts pertaining to graph-structured data, kernel methods, and graph kernels. In Section~\ref{main}, we use these concepts to define and categorize popular graph kernels.

\subsection{Graph Data}
\label{sec:graphdata}
A \new{graph} $G$ is a pair $(V,E)$ of a finite set of \new{vertices} $V$ and a set of \new{edges} $E \subseteq \{ \{u,v\} \subseteq V \mid u \neq v \}$. A vertex is typically used to represent an object (e.g., an atom) and an edge a relation between objects (e.g., a molecular bond). We denote the set of vertices and the set of edges of $G$ by $V(G)$ and $E(G)$, respectively.  We restrict our attention to \emph{undirected} graphs in which no two edges with identical (unordered) end points, nor any self-cycles exist.
For ease of notation we denote the edge $\{u,v\}$ in $E(G)$ by $(u,v)$ or $(v,u)$.
A \new{labeled graph} is a graph $G$ endowed with a \new{label function} $l \colon V(G) \to \Sigma$, where $\Sigma$ is some alphabet, e.g., the set of natural or real numbers. We say that $l(v)$ is the \new{label} of $v$. In the case $\Sigma=\bbR^d$ for some $d>0$, $l(v)$ is the \new{(continuous) attribute} of $v$. In Section~\ref{sec:applications}, we give examples of applications involving graphs with vertex labels and attributes. The edges of a graph may also be assigned labels or attributes (e.g., weights representing vertex  similarity), in which case the domain of the labeling function $l$ may be extended to the edge set.

\begin{figure}
	\centering
	\includegraphics[width=.8\textwidth]{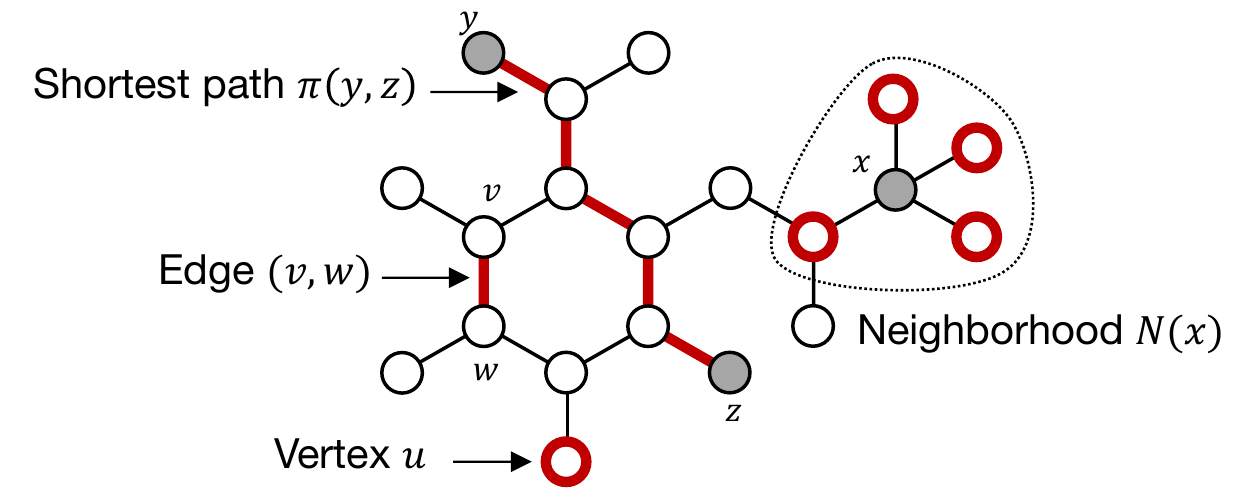}
	\caption{{Graph representation fundamentals.} Illustration of a graph $G$ in which each circle represents a different vertex and each line connecting two circles an edge. Some edges and vertices are highlighted to illustrate specific graph concepts. Here, $\pi(y, z)$ represents the shortest path (sequence of vertices) between vertices $y$ and $z$. The neighborhood $N(x)$ of a vertex $x$ is the set of vertices adjacent to $x$. \label{fig:fundamentals}}
\end{figure}

We let $N(v)$ denote the \new{neighborhood} of a vertex $v\in G$ in $V(G)$, i.e., $N(v) = \{ u \in V(G) \mid (v, u) \in E(G) \}$. The degree of a vertex is the size of its neighborhood, $\mbox{deg}(u) = |N(u)|$. A \new{walk} $\omega$ in a graph is an ordered sequence of vertices $\omega = (u, \dots, v)$ such that any two subsequent vertices are connected by an edge. A \new{$(u,v)$-path} is a walk that starts in $u$ and ends in $v$ with no repeated vertices. A graph $G$ is called \new{connected} if there is a path between any pair of vertices in $V(G)$ and \new{disconnected} otherwise. Paths, vertices, edges and neighborhoods are illustrated in Figure~\ref{fig:fundamentals}.

We say that two unlabeled graphs $G$ and $H$ are \new{isomorphic}, denoted by $G \simeq H$, if there exists a bijection $\varphi: V(G) \to V(H)$, such that $(u,v) \in E(G)$ if and only if $(\varphi(u),\varphi(v)) \in E(H)$ for all $u,v$ in $V(G)$. For labeled graphs, isomorphism holds only if the bijection maps only vertices and edges with the same label.
Finally, a graph $G'=(V',E')$ is a \emph{subgraph} of a graph $G=(V,E)$ if $V' \subseteq V$ and $E' \subseteq E$. Let $S \subseteq V(G)$ be a subset of vertices in $G$. Then $G[S] = (S,E_S)$ denotes the \new{subgraph induced} by $S$ with $E_S = \{ (u,v) \in E(G) \mid u,v \in S \}$.

Graphs are often represented in matrix form. Perhaps most frequent is the  \new{adjacency matrix} $\vec{A}$ with binary elements $a_{uv} = \{1 \mbox{ iff } (u,v) \in E\}$.\footnote{Weighted graphs are represented by their corresponding edge weight matrix.} An alternative representation is the \new{graph Laplacian} $\vec{L}$, defined as $\vec{L} = \vec{D}-\vec{A}$, where $\vec{D}$ is the diagonal degree matrix, such that $d_{uu} = \mbox{deg}(u)$. Finally, the \new{incidence matrix} $\vec{M}$ of a graph is the binary $n \times n^2$ matrix with vertex-edge-pair elements $m_{ue} = \{1 \mbox{ iff } e = (u, v) \in E\}$ representing the event that the vertex $u$ is incident on the edge $e$. It holds that $\vec{L} = \vec{M}\vec{M}^\top$. The matrices $\vec{A}, \vec{L}$, and $\vec{M}$ all carry the same information.

\subsection{Kernel Methods}
\label{sec:kernelmethods}

\new{Kernel methods} refer to machine learning algorithms that learn by comparing pairs of data points using particular similarity measures---\emph{kernels}. We give an overview below; for an in-depth treatment, see~\citep{Scholkopf2001,Shawe-Taylor2004}. Consider a non-empty set of data points $\chi$, such as $\mathbb{R}^d$ or a finite set of graphs, and let $k \colon \chi \times \chi \to \RR$ be a function. Then, $k$ is a kernel on $\chi$ if there is a Hilbert space $\cH_k$ and a feature map $\phi \colon \chi \to \cH_k$ such that $k(x,y) = \langle \phi(x), \phi(y) \rangle$ for $x, y \in \chi$, where $\langle \cdot, \cdot \rangle$ denotes the inner product of $\cH_k$. Such a feature map exists if and only if $k$ is a \new{positive-semidefinite} function. A trivial example is where $\chi = \mathbb{R}^d$ and $\phi(\vec{x}) = \vec{x}$, in which case the kernel equals the dot product, $k(\vec{x},\vec{y}) = \vec{x}^\top \vec{y}$.

An important concept in kernel methods is the \new{Gram matrix} $\vec{K}$, defined with respect to a finite set of data points $x_1, ..., x_m \in \chi$. The Gram matrix of a kernel $k$ has elements $K_{ij}$, for $i, j \in \{0, ..., m\}$ equal to the kernel value between pairs of data points, i.e., $K_{ij} = k(x_i, x_j)$. If the Gram matrix of $k$ is positive semidefinite for every possible set of data points, $k$ is a kernel~\citep{scholkopf1997kernel}. Kernel methods have the desirable property that they do not rely on explicitly characterizing the vector representation $\phi(x)$ of data points, but access data only via the Gram matrix $\vec{K}$.
The benefit of this is often illustrated using the Gaussian radial basis function (RBF) kernel on
$\mathbb{R}^d$, $d \in \mathbb{N}$, defined as
\begin{equation}\label{eq:rbf}
	k_{\text{RBF}}(\vec{x}, \vec{y}) = \exp \left(-\frac{\norm{\vec{x} - \vec{y}}^2}{2 \sigma^2} \right),
\end{equation}
where $\sigma$ is a bandwidth parameter.
The Hilbert-space associated with the Gaussian RBF kernel has infinite dimension but the kernel may be readily computed for any pair of points $(\vec{x}, \vec{y})$ (see~\cite{Moh+2012} for further details).
Kernel methods have been developed for most machine learning paradigms, e.g., \new{support vector machines} (SVM) for classification~\citep{cortes1995support}, \new{Gaussian processes} (GP) for regression~\citep{rasmussen2004gaussian}, \new{kernel PCA}, \new{k-means} for unsupervised learning and clustering~\citep{scholkopf1997kernel}, and \new{kernel density estimation} (KDE) for density estimation~\citep{silverman2018density}. In this work, we restrict our attention to \emph{classification} of objects in a non-empty set of graphs $\GG$. In this setting, a kernel $k\colon \GG \times \GG \to \bbR$ is called a \new{graph kernel}. Like kernels on vector spaces, graph kernels can be calculated either \new{explicitly} (by computing $\phi$) or \new{implicitly} (by computing only $k$). Traditionally, learning with implicit kernel representations means that the value of the chosen kernel applied to every pair of graphs in the training set must be computed and stored. Explicit computation means that we compute a finite dimensional feature vector for each graph; the values of the kernel can then be computed on-the-fly during learning as the inner product of feature vectors. If explicit computation is possible, and the dimensionality of the resulting feature vectors is not too high, or the vectors are sparse, then it is usually faster and more memory efficient than implicit computation, see also~\cite{Kri+2014,Kriege2019dami}.

\subsection{Design Paradigms For Kernels on Structured Data}
\label{sec:convkernel}
When working with vector-valued data, it is common practice for kernels to compare objects $\vec{x}, \vec{y} \in \mathbb{R}^d$ using differences between vector components (see for example the Gaussian RBF kernel in \cref{eq:rbf}). The structure of a graph, however, is invariant to permutations of its representation---the ordering by which vertices and edges are enumerated does not change the structure---and vector distances between, e.g., adjacency matrices are typically uninformative. For this reason, it is important to compare graphs in ways, that are themselves permutation invariant. As mentioned previously, two graphs with identical structure (irrespective of representation) are called \new{isomorphic}, a concept that could in principle be used for learning. However, not only is there no known polynomial-time algorithm for testing graph isomorphism~\citep{Johnson2005} but isomorphism is also typically too strict for learning---it is akin to learning with the equality operator. In practice, it is often desirable to have \emph{smoother} metrics of comparison in order to gain generalizable knowledge from the comparison of graphs.

The vast majority of graph kernels proposed in the literature are instances of so-called \new{convolution kernels}.
Given two discrete structures, e.g., two graphs, the idea of Haussler's Convolution Framework~\cite{Hau1999} is to decompose these two structures into substructures, e.g., vertices or subgraphs, and then evaluate a kernel between each pair of such substructures. The convolution kernel is defined below.

\begin{definition}[Convolution Kernel]\label{conv}
	Let $\mathcal{R} = \mathcal{R}_1 \times \cdots \times \mathcal{R}_d$ denote a space of components such that a composite object $X \in \mathcal{X}$ decomposes into elements of $\mathcal{R}$. Further, let $R : \mathcal{R} \rightarrow \mathcal{X}$ denote the mapping from components to objects, such that $R(x) = X$ if and only if the components $x \in \mathcal{R}$ make up the object $X \in \mathcal{X}$, and let $R^{-1}(X) = \{x \in \mathcal{R} : R(x) = X\}$.
	Then, the \emph{$R$-convolution kernel} is
	\begin{equation}\label{conk}
		k_{\text{CV}}(X,Y) =
		\sum_{x \in R^{-1}(X)}
		\sum_{y \in R^{-1}(Y)}
		\underbrace{\prod_{i=1}^{d} k_i(x_i,y_i)}_{k(x,y)},
	\end{equation}
	where $k_i$ is a kernel on $\mathcal{R}_i$ for $i$ in $\{1,\dots,d\}$.
\end{definition}

In our context, we may view the inverse map $R^{-1}(G)$ of the convolution kernel as the set of all components of a graph $G$ that we wish to compare. A simple example of the $R$-convolution kernel is the \new{vertex label kernel} for which the mapping $R$ takes the attributes $x_u \in \mathcal{R}$ of each vertex $u \in G \cup H$ and maps them to the graph that $u$ is a member of. We expand on this notion in \cref{vertexlabel}.
A benefit of the convolution kernel framework when working with graphs is that if the kernels on substructures are invariant to orderings of vertices and edges, so is the resulting graph kernel.

A property of convolution kernels often regarded as unfavorable is that the sum
in \cref{conk} applies to all pairs of components.
When the considered components become more and more specific, each object becomes
increasingly similar to itself, but no longer to any other objects.
This phenomenon is referred to as the \new{diagonal dominance problem}, since the
entries on the main diagonal of the Gram matrix are much higher than the others
entries.
This problem was observed for graph kernels, for which weights between the
components were introduced to alleviate the problem~\cite{Yan+2015,Aio+2015}.
In addition, the fact that convolution kernels compare all pairs of components may be unsuitable in situations where each component of one object corresponds
to exactly one component of the other (such as the features of two faces).
\citet{shin2008generalization} studied \new{mapping kernels}, where the sum moves over a predetermined subset of pairs rather than the entire cross product.
It was shown that, for general primitive kernels $k$, a valid mapping kernel is obtained if and only if the considered subsets of pairs are transitive on
$\mathcal{R}$.
This does not necessarily hold, when assigning the components of two objects to each other such that a correspondence of maximum total similarity w.r.t.\@ $k$ is obtained.
As a consequence, this approach does not lead to valid kernels in general.
However, graph kernels following this approach have been studied in detail and
are often referred to as \new{optimal assignment kernels}, see \cref{sec:assignment}.

\section{Graph Kernels}\label{main}
The first methods for graph comparison referred to as \emph{graph kernels} were proposed in 2003~\cite{Gae+2003, Kas+2003}. However, several approaches similar to graph kernels had been developed in the field of chemoinformatics, long before the term graph kernel was coined. The timeline in \cref{timeline} shows milestones in the development of graph kernels and related learning algorithms for graphs. We postpone the discussion of the latter to \cref{sec:app:chem}.
Following the introduction of graph kernels, subsequent work focused for a long time on making kernels computationally tractable for large graphs with (predominantly) discrete vertex labels. Since 2012, several kernels specifically designed for graphs with continuous attributes have been proposed. 
It remains a current challenge in research to develop neural techniques for graphs that are able to learn feature representations that are clearly superior to the fixed feature spaces used by graph kernels.

In the following, we give an overview of the graph kernel literature in order of popular design paradigms. We begin our treatment with kernels that are based on neighborhood aggregation techniques. The subsequent subsections deal with \new{assignment-} and \new{matching-based} kernels, and kernels based on the extraction of subgraph patterns, respectively. The final subsections deal with kernels based on \new{walks and paths}, and kernels that do not fall into either of the previous categories. \cref{rlw} gives an overview of the discussed graph kernels and their properties.

\begin{table}
	\ra{1.1}
	\caption{Summary of selected graph kernels: Computation by explicit
		(EX) and implicit (IM) feature mapping and support for
		attributed
		graphs. The column 'Labels' refers to whether the kernels support comparison of graphs with discrete vertex and edge labels in a way that depends on the interplay between structure and labels. The column 'Attributes' refer to the same capability but for continuous or more general vertex attributes. $^\star$ --- not considered in publication,
		but method can be extended; $^\dagger$ --- vertex annotations only.}
	\label{rlw}
	\centering
		\resizebox{1.0\textwidth}{!}{
		\rowcolors{2}{gray!15}{white}
		\setlength{\tabcolsep}{3pt}
		\renewcommand{\arraystretch}{1.1}
		\begin{tabular}{lccc}
			\toprule
			\textbf{Graph Kernel}                                                  & \textbf{Computation} & \textbf{Labels}   & \textbf{Attributes}   \\
			\midrule
			Shortest-Path~\cite{Borgwardt2005}                              & IM                            & $\;\,$+$^\dagger$ & $\;\,$+$^\dagger$   \\
			Generalized Shortest-Path~\cite{Her+2015}                       & IM                            & +                 & $\;\,$+$^\dagger$   \\
			Graphlet~\cite{She+2009}                                        & EX                            & --                & --                  \\
			Cycles and Trees~\cite{Hor+2004}                                & EX                            & {$\;\,$+$^\star$} & --                  \\
			Tree Pattern Kernel~\cite{Ram+2003,Mah+2009}                    & IM                            & +                 & $\;\,$+$^\star$     \\
			Ordered Directed Acyclic Graphs~\cite{Mar+2012,Mar+2012a}       & EX                            & +                & --                  \\
			GraphHopper \cite{Fer+2013}                                     & IM                            & $\;\,$+$^\dagger$ & +                   \\
			Graph Invariant \cite{Ors+2015}                                 & IM                            & +                 & +                   \\
			Subgraph Matching \cite{Kri+2012}                               & IM                            & +                 & +                   \\
			Weisfeiler-Lehman Subtree~\cite{She+2011}                       & EX                            & +                 & --                  \\
			Weisfeiler-Lehman Edge~\cite{She+2011}                          & EX                            & +                 & --                  \\
			Weisfeiler-Lehman Shortest-Path~\cite{She+2011}                 & EX                            & +                 & --                  \\
			k-dim.~Local Weisfeiler-Lehman Subtree~\cite{Mor+2017}          & EX                            & +                 & --                  \\
			Neighborhood Hash Kernel~\cite{Hid+2009}                        & EX                            & +                 & --                  \\
			Propagation Kernel~\cite{Neu+2016}                              & EX                            & +                 & +                   \\
			Neighborhood Subgraph Pairwise Distance Kernel~\cite{Cos+2010}  & EX                            & +                 & --                  \\
			Random Walk~\cite{Gae+2003,Kas+2003,Mah+2004,Vis+2010,Sug+2015,Kan+2012} & IM                   & +                 & +                   \\
			Optimal Assignment Kernel~\cite{Fro+2005}                       & IM                            & +                 & +                   \\
			Weisfeiler-Lehman Optimal Assignment~\cite{Kri+2016}            & IM                            & +                 & --                  \\
			Pyramid Match~\cite{Nik+2017}                                   & IM                            & +                 & --                  \\
			Matchings of Geometric Embeddings~\cite{Joh+2015}               & IM                            & +                 &  $\;\,$+$^{\star}$                 \\
			Descriptor Matching Kernel~\cite{Su+2016}                       & IM                            & +                 & $\;\,$+$^\dagger$                   \\
			Graphlet Spectrum~\cite{Kon+2009}                               & EX                            & +                 & --                  \\
			Multiscale Laplacian Graph Kernel~\cite{Kon+2016}               & IM                            & +                 & $\;\;\,$+$^{\star\dagger}$                   \\
			Global Graph Kernel~\cite{Joh+2014}                             & EX                            & --                & --                  \\
			Deep Graph Kernels~\cite{Yan+2015}                              & IM                            & +                 & --                  \\
			Smoothed Graph Kernels~\cite{Yan+2015a}                         & IM                            & $\;\,$+$^\star$   & --                  \\
			Hash Graph Kernel~\cite{Mor+2016}                               & EX                            & +                 & +                   \\
			Depth-based Representation Kernel~\citep{bai2014graph}          & IM                            & --                & --                  \\
			Aligned Subtree Kernel~\cite{Bai+2015}                          & IM                            & +                 & --                  \\
			\bottomrule
		\end{tabular}}
\end{table}

\usetikzlibrary{chains}
\makeatletter
\tikzset{west below/.code=\tikz@lib@place@handle@{#1}{north west}{0}{-1}{south west}{1}}
\makeatother
\tikzset{
	typnode/.style={anchor=north west, text width=12cm, inner sep=0mm, node distance=2.4mm},
	data/.style={draw=gray, rectangle, font=\scriptsize, inner sep=.8mm},
}
\newcommand{\embed}[1]{{\color{gray}{#1}}}
\newcommand{\chem}[1]{{\textit{#1}}}
\newcommand{\kernDisc}[1]{{#1}}
\newcommand{\kernAttr}[1]{{\textbf{#1}}}
\newcommand{\neural}[1]{{\color{brown}{#1}}}
\begin{figure}[]
	\begin{tikzpicture}[x=1cm, y=-7mm]
		\path (0,0) edge[-latex] ++(0,24)
		edge[dashed] ++ (down:1);

		\foreach \y [evaluate=\y as \xear using int(1972+\y*3)] in {0,1,...,15}{
			\node[left=2pt,anchor=east,xshift=0,font=\scriptsize] at (0,1.5*\y) {$\xear$};
			\draw (-0.1,1.5*\y) -- (0.1,1.5*\y);
			\draw (0,1.5*\y+.5) -- (0.1,1.5*\y+.5);
			\draw (0,1.5*\y+1) -- (0.1,1.5*\y+1);
		}

		\begin{scope}[start chain=ch1 going west below, node distance=+1em]
			\foreach \Year/\Text in {%
				1973/{\embed{\chem{Fingerprints for chemical similarity}}~\cite{Adamson1973}},
				1986/{\embed{\chem{Systematic evaluation of fingerprint similarities}}~\cite{Willett1986}},
				2000/{\embed{\chem{Extended connectivity fingerprints}}~\cite{Rogers2010}},
				2003/{\kernAttr{Random walk kernels}~\cite{Gae+2003,Kas+2003}},
				2003/{\kernDisc{Tree pattern kernels}~\cite{Ram+2003,Mah+2009}},
				2004/{\kernDisc{Cycles and Trees kernel}~\cite{Hor+2004}},
				2005/{\kernAttr{Shortest-path kernel}~\cite{Borgwardt2005}},
				2005/{\kernDisc{\chem{Kernels from chemical similarities}}~\cite{Ralaivola2005}},
				2005/{\kernAttr{\chem{Optimal assignment kernels}}~\cite{Fro+2005}},
				2005/{\neural{\chem{Molecular graph networks}}~\cite{Mer+2005}},
				2009/{\kernDisc{Graphlet kernels}~\cite{She+2009}},
				2009/{\kernDisc{Neighborhood Hash Kernel}~\cite{Hid+2009}},
				2009/{\kernDisc{Weisfeiler-Lehman kernels}~\cite{She+2011}},
				2010/{\kernDisc{Neighborhood subgraph pairwise distance kernel} \cite{Cos+2010}},
				2012/{\kernDisc{Ordered Directed Acyclic Graphs}~\cite{Mar+2012}},
				2012/{\kernAttr{Subgraph matching kernel}~\cite{Kri+2012}},
				2013/{\kernAttr{GraphHopper kernel}~\cite{Fer+2013}},
				2015/{\kernAttr{Generalized shortest-path kernel}~\cite{Her+2015}},
				2015/{\kernAttr{Graph Invariant}~\cite{Ors+2015}},
				2015/{\neural{\chem{Neural molecular fingerprints}}~\cite{Duv+2015}},
				2016/{\kernAttr{Descriptor matching kernel}~\cite{Su+2016}},
				2016/{\kernAttr{Hash graph kernels}~\cite{Mor+2016}},
				2016/{\kernDisc{Valid optimal assignment kernels}~\cite{Kri+2016}},
				2017/{\neural{Graph convolutional networks}~\cite{Kip+2017}},
				2017/{\neural{Neural message passing}~\cite{Gil+2017}},
				2017/{\neural{GraphSAGE}~\cite{Ham+2017}},
				2018/{\neural{SplineCNN}~\cite{Fey2018}},
				2019/{\neural{$k$-GNN}~\cite{Mor+2019}}
				}{
				\node[typnode, at=(right:3.4cm), on chain=ch1, alias=Text] {\Text};
				\node[data,    base left=+2em of Text, alias=Year] {\Year};
				\draw[-|] (Year.east) -- ++(right:3mm);
				\draw     (Year.west) -- ++(left:3mm)
				-- ([shift=(right:3mm)] 0,{(\Year-1972)/2})
				--                     (0,{(\Year-1972)/2});
			}
		\end{scope}
	\end{tikzpicture}
	\caption{{Timeline.} 
		Selected techniques for graph classification with a focus on kernels.
		Techniques based on fingerprints are marked in \embed{gray} and methods
		using neural networks in \neural{brown}. Methods proposed for cheminformatics
		are shown in \chem{italics}, kernels for attributed graphs in \textbf{bold}.}\label{timeline}
\end{figure}
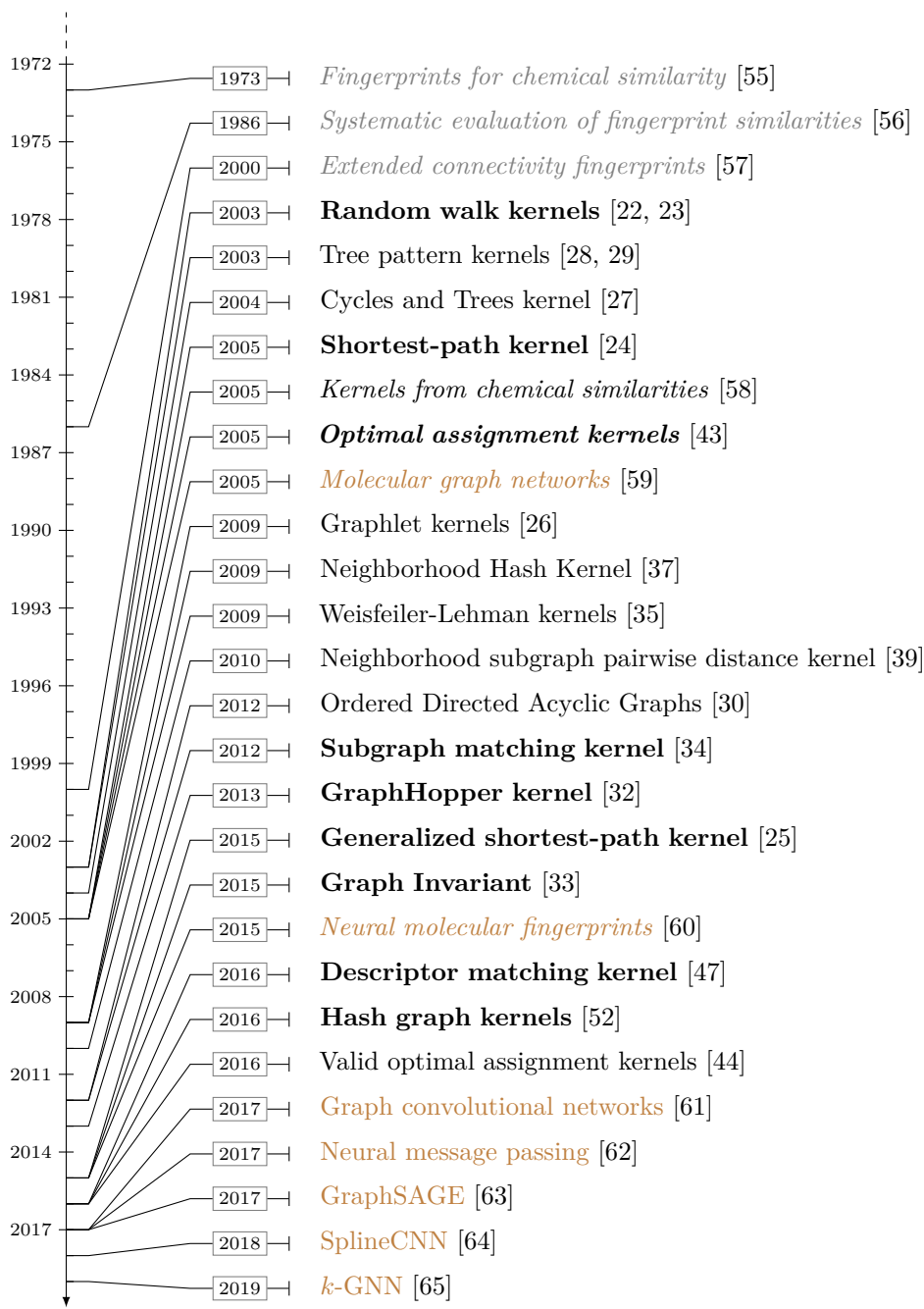

\subsection{Neighborhood Aggregation Approaches}\label{sec:naa}
One of the dominating paradigms in the design of graph kernels is representation and comparison of \emph{local} structure. Two vertices are considered similar if they have identical labels---even more so if their neighborhoods are labeled similarly. Expanding on this notion, two graphs are considered similar if they are composed of vertices with similar neighborhoods, i.e., that they have similar local structure. The different ways by which local structure is defined, represented and compared form the basis for several influential graph kernels. We describe a first example next.

\begin{figure}[h!]
	\centering
	\includegraphics[width=.95\textwidth]{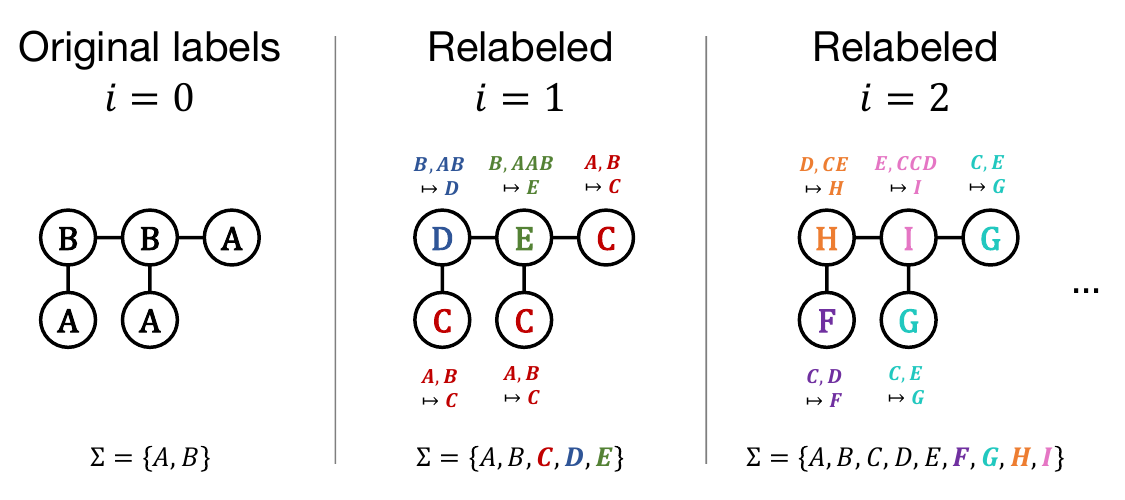}
	\caption{{Weisfeiler-Lehman (WL) relabeling.} Two iterations of Weisfeiler-Lehman vertex relabeling for a graph with discrete labels in $\{A, B\}$. At initialization (left), vertex labels are left in their original state. In the first iteration (middle), a new label is computed for each vertex, determined by the unique combination of its own and its neighbors' labels. For example, the top-left vertex with label $B$ has neighbors with labels $A$ and $B$. This combination is renamed $D$ and assigned to the top-left vertex in the first iteration. The second iteration (right) proceeds analogously.}\label{fig:wl}
\end{figure}

Neighborhood aggregation approaches work by assigning an attribute to each vertex based on a summary of the local structure around them. Iteratively, for each vertex, the attributes of its immediate neighbors are aggregated to compute a new attribute for the target vertex, eventually representing the structure of its extended neighborhood. \citet{She+2011} introduced a highly influential class of neighborhood aggregation kernels for graphs with discrete labels based on the 1-dimensional \new{Weisfeiler-Lehman} ($1$-WL) or \new{color refinement} algorithm---a well-known heuristic for the graph isomorphism problem, see, e.g.,~\cite{Bab+1979}. We illustrate an application of the $1$-WL algorithm in \cref{fig:wl}.

Let $G$ and $H$ be graphs, and let $l \colon V(G) \cup V(H)\to \Sigma$ be the observed vertex label function of $G$ and $H$.\footnote{If the graph is unlabeled, let $l$ map to a constant.} In a series of iterations $i = 0, 1, \ldots$, the \textsc{$1$-WL} algorithm computes new label functions $l^i \colon V(G) \cup V(H)\to \Sigma$, each of which can be used to compare $G$ and $H$. In iteration $0$ we set $l^0 = l$ and in subsequent iterations $i>0$, we set
\begin{equation}\label{na}
	l^{i}(v) = \textsf{\small relabel} ((l^{i-1}(v),\textsf{\small sort}(\{\!\!\{ l^{i-1}(u) \mid u \in N(v) \}\!\!\}))),
\end{equation}
for $v \in V(G) \cup V(H)$, where $\textsf{\small sort}(S)$ returns a sorted tuple of the multiset $S$ and the injection $\textsf{\small relabel}(p)$ maps the pair $p$ to a unique value in $\Sigma$ which has not been used in previous iterations. Now if $G$ and $H$ have an unequal number of vertices with label  $\sigma \in \Sigma$, we can conclude that the graphs are not isomorphic. Moreover, if the cardinality of the image of $l^{i-1}$ equals the cardinality of the image of $l^{i}$, the algorithm terminates.

The idea of the \new{Weisfeiler-Lehman subtree kernel} is to compute the above algorithm for $h \geq 0$ iterations, and after each iteration $i$ compute a feature vector $\phi^i(G) \in \RR^{|\Sigma_i|}$ for each graph $G$, where $\Sigma_i \subseteq \Sigma$ denotes the image of $l^i$. Each component $\phi^i(G)_{\sigma^i_j}$ counts the number of occurrences of vertices labeled with $\sigma^i_j \in \Sigma_i$. The overall feature vector $\phi_{\text{WL}}(G)$ is defined as the concatenation of the feature vectors of all $h$ iterations, i.e.,
\begin{equation*}
	\mleft(\phi^0(G)_{\sigma^0_1}, \dotsc, \phi^0(G)_{\sigma^0_{|\Sigma_0|}}, \dotsc,\phi^h(G)_{\sigma^h_1}, \dotsc \phi^h(G)_{\sigma^h_{|\Sigma_h|}} \mright)\,.
\end{equation*}
Then the Weisfeiler-Lehman subtree kernel for $h$ iterations is $k_{\text{WL}}(G,H) = \langle \phi_{\text{WL}}(G), \phi_{\text{WL}}(H) \rangle$. The running time for a single feature vector computation is in $\cO(hm)$ and $\cO(Nhm+N^2hn)$ for the computation of the Gram matrix for a set of $N$ graphs~\cite{She+2011}, where $n$ and $m$ denote the maximum number of vertices and edges over all $N$ graphs, respectively.

The WL subtree kernel suggests a general paradigm for comparing graphs at different levels of resolution: iteratively relabel graphs using the WL algorithm and construct a graph kernel based on a base kernel applied at each level. Indeed, in addition to the subtree kernel, \citet{She+2011} introduced two other variants, the \new{Weisfeiler-Lehman edge} and the \new{Weisfeiler-Lehman shortest-path kernel}. Instead of counting the labels of vertices after each iteration the Weisfeiler-Lehman edge kernel counts the colors of the two endpoints for all edges. The Weisfeiler-Lehman shortest-path kernel is the sum of shortest-path kernels applied to the graphs with refined labels $l^i$ for $i \in \{0,\dots,h\}$.

\citet{Mor+2017} introduced a graph kernel based on higher dimensional variants of the Weisfeiler-Lehman algorithm. Here, instead of iteratively labeling vertices, the algorithm labels $k$-tuples or sets of cardinality $k$. \citet{Mor+2017} also provide efficient approximation algorithm to scale the algorithm up to large datasets.  In~\cite{Hid+2009}, a graph kernel similar to the $1$-WL was introduced which replaces the neighborhood aggregation function~\cref{na} by a function based on binary arithmetic. Similarly, in~\cite{Neu+2016} the propagation kernel is defined which propagates labels, and real-valued attributes for several iterations while tracking their distribution for every vertex. A randomized approach based on $p$-stable locality-sensitive hashing is used to obtain unique features after each iteration. 
In recent years, graph neural networks (GNNs) have emerged as an alternative to graph kernels. Standard GNNs can be viewed as a feed-forward neural network version of the $1$-WL algorithm, where colors (labels) are replaced by continuous feature vectors and network layers are used to aggregate over vertex neighborhoods~\cite{Ham+2017,Kip+2017}. Recently, a connection between the $1$-WL and GNNs has been established~\cite{Mor+2019}, showing that any possible GNN architecture cannot be more powerful than the $1$-WL in terms of distinguishing non-isomorphic graphs.

\citet{bai2014graph,Bai+2015} proposed graph kernels based on \new{depth-based representations}, which can be seen as a different form of neighborhood aggregation. For a vertex $v$ the \new{$m$-layer expansion subgraph} is the subgraph induced by the vertices of shortest-path distance at most $m$ from the vertex $v$. In order to obtain a vertex embedding for $v$ the Shannon entropy of these subgraphs is computed for all $m\leq h$, where $h$ is a given parameter~\citep{bai2014graph}.
A similar concept is applied in \cite{Bai+2015}, where depth-based representations are used to compute strengthened vertex labels. Both methods are combined with matching-based techniques to obtain a graph kernel.

\subsection{Assignment- and Matching-based Approaches}\label{sec:assignment}

\begin{figure}
	\centering
	\includegraphics[width=.95\textwidth]{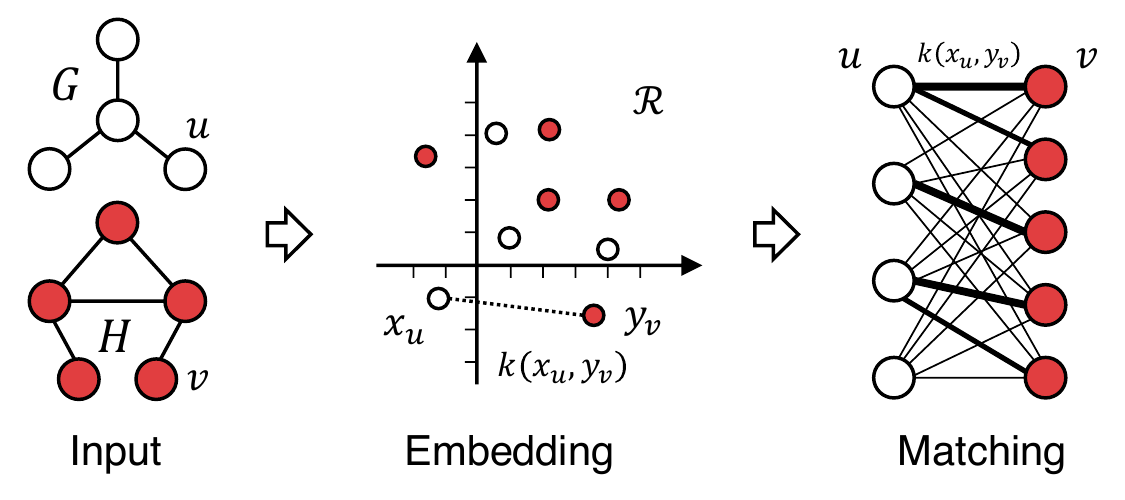}
	\caption{{Assignment kernels.} Illustration of optimal assignment kernels with vertex embeddings. The vertices of two different graphs (left), $G$ and $H$ are embedded in a common space $\mathcal{R}$ (middle). For example, vertices $u\in V_G, v\in V_H$ are given embeddings $x_u, y_v \in \mathcal{R}$. Finally, a bipartite graph with weights determined by the distances between the vertex embeddings of the two graphs is constructed and used to compute an optimal matching between the vertex sets. The weight of the matching is used to compute the kernel value $k(G, H)$. }\label{fig:matching}
\end{figure}

A common approach to comparing two composite or structured objects is to identify the best possible matching of the components making up the two objects.
For example, when comparing two chemical molecules it is instructive to map each atom in one graph to the atom in the other graph that is most similar in terms of, for example, neighborhood structure and attached chemical and physical measurements. This idea has been used also in graph kernels, an early example of which was proposed by~\citet{Fro+2005} in the \new{optimal assignment} (OA) kernel. In the OA kernel, each vertex is endowed with a representation (e.g., a label) that is compared using a base kernel. Then, a similarity value for a pair of graphs is computed based on a mapping between their vertices such that the total similarity between the matched vertices with respect to a base kernel is maximized.  An illustration of the optimal assignment kernel can be seen in Figure~\ref{fig:matching}. The OA kernel can be defined as follows. 

\begin{definition}[Optimal assignment kernel]\label{def:oakernel}
Let $X = \{x_1,\dots, x_n \}$ and $Y= \{y_1,\dots, y_n\}$ be sets of components
from $\mathcal{R}$ and $k : \mathcal{R} \times \mathcal{R} \to \bbR$ a base kernel on components.
The \emph{optimal assignment kernel} is
\begin{equation}\label{eq:assignment}
	K_A(X,Y) = \max_{\pi \in \Pi_n} \sum_{i=1}^{n} k(x_i, y_{\pi(i)}),
\end{equation}
where $\Pi_n$ is the set of all possible permutations of $\{1,\dots,n\}$. In order to apply the assignment kernel to sets of different cardinality, we fill the smaller set with objects $z$ and define $k(z, x) = 0$ for all $x \in \mathcal{R}$.
\end{definition}

The careful reader may have noticed a superficial similarity between the OA kernel and the $R$-convolution and mapping kernels (see \cref{sec:convkernel}). However, instead of summing the base kernel over a fixed ordering of component pairs, the OA kernel searches for the optimal mapping between components of two objects $X, Y$.
Unfortunately, this means that Equation~\eqref{eq:assignment} is not a positive-semidefinite kernel in general~\cite{Ver2008,Vis+2010}. This fact complicates the use of assignment similarities in kernel methods, although generalizations of SVMs for arbitrary similarity measures have been developed, see, e.g.,\@ \cite{Loosli2015} and references therein. Moreover, kernel methods, such as SVMs, have been found to work well empirically also with indefinite kernels~\citep{Joh+2015}, without enjoying the guarantees that apply to positive definite kernels.

Several different approaches to obtain positive definite graph kernels from indefinite assignment similarities have been proposed. \citet{Woznica2010} derived graph kernels from set distances and employed a
matching-based distance to compare graphs, which was shown to be a metric~\cite{Ramon2001}.
In order to obtain a valid kernel, the authors use so-called \emph{prototypes}, an idea prevalent also in the theory of learning with (non-kernel) similarity functions under the name \new{landmarks}~\citep{balcan2008theory}. Prototypes are a selected set of instances (e.g., graphs) to which all other instances are compared. Each graph is then represented by a feature vector in which each component is the distance to a different prototype. Prototypes were used also by \citet{Joh+2015} who proposed to embed the vertices of a graph into the $d$-dimensional real vector space in order to compute a matching between the vertices of two graphs with respect to the Euclidean distance. Several methods for the embedding were proposed; in particular, the authors used Cholesky decompositions of matrix representations of graphs including the graph Laplacian and its pseudo-inverse. The authors found empirically that the indefinite graph similarity matrix from the matching worked as well as prototypes. In Section~\ref{exp}, we use this, indefinite version.

Instead of generating feature vectors from prototypes, \citet{Kri+2016} showed
that Equation~\eqref{eq:assignment} is a valid kernel for a restricted class of
base kernels $k$. These, so-called \emph{strong base kernels}, give rise to
hierarchies from which the optimal assignment kernels are computed in linear
time by histogram intersection. For graph classification, a base kernel was
obtained from Weisfeiler-Lehman refinement. The derived \new{Weisfeiler-Lehman
optimal assignment kernel} often provides better classification accuracy on
real-world benchmark datasets than the Weisfeiler-Lehman subtree kernel (see~\cref{exp}).
The weights of the hierarchy associated with a strong base kernel can be optimized
via multiple kernel learning~\cite{Kriege2019}.

\citet{Pachauri2013} studied a generalization of the assignment problem to
more than two sets, which was used to define \new{transitive assignment kernels}
for graphs~\citep{Schiavinato2015}.
The method is based on finding a single assignment between the vertices of all
graphs of the dataset instead of finding an optimal assignment for each pairs
of graphs.
This approach satisfies the transitivity constraint of mapping kernels and
therefore leads to positive-semidefinite kernels. However, non-optimal assignments
between individual pairs of graphs are possible.
\citet{Nik+2017} proposed a matching-based approach based on the Earth Mover's
Distance, which results in an indefinite kernel function. In order to deal with
this they employ a variation of the SVM algorithm, specialized for learning with indefinite kernels. Additionally, they propose an alternative solution based on the \new{pyramid match kernel}, a generic kernel for comparing sets of features~\citep{grauman2007pyramid}. The pyramid match kernel avoids the indefiniteness of other assignment kernels by comparing features through a multi-resolution histograms (with bins determined globally, rather than for each pair of graphs).

\subsection{Subgraph Patterns}

\begin{figure}
	\centering
	\includegraphics[width=.8\textwidth]{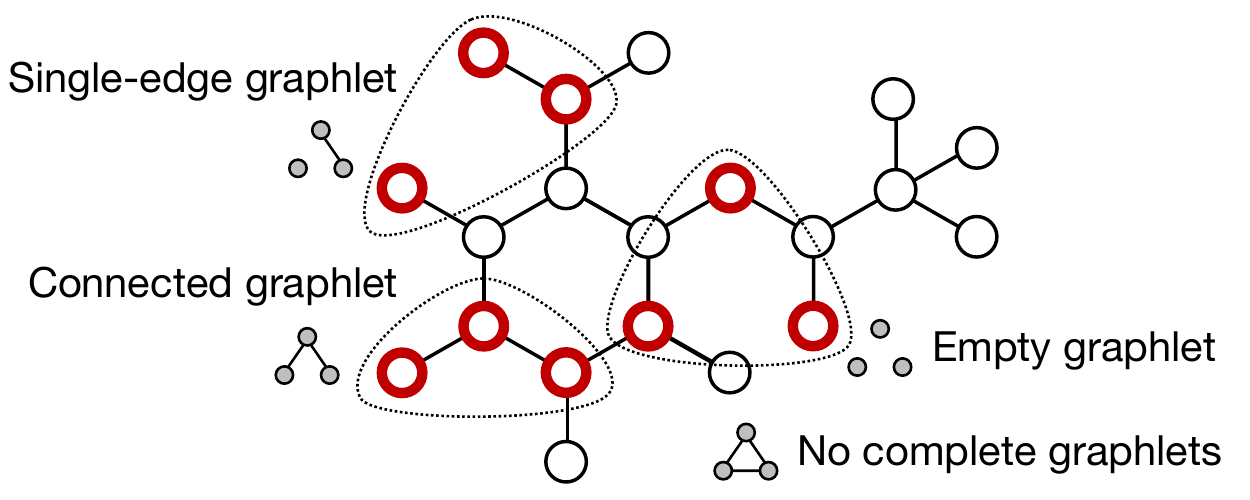}
	\caption{{Graphlets.} Illustration of graphlets on 3 vertices in a graph $G$. Each circle represents a vertex and each line connecting two circles an edge. A 3-graphlet is an instance of an edge pattern on the induced subgraph of 3 vertices. We highlight examples of empty (right), single-edge (top-left), and double-edge (bottom-left) 3-graphlets. No complete graphlets are present in the graph. The graphlet kernel is computed by comparing the number of instances of each pattern in two graphs.%
	}\label{fig:graphlet}
\end{figure}

\label{vertexlabel}
In many applications, a strong baseline for representations of composite objects such as documents, images or graphs is one that ignores the structure altogether and represents objects as \new{bags of components}. A well-known example is the so-called bag-of-words representation of text---statistics of word occurrences without context---which remains a staple in natural language processing. For additional specificity, it is common to compare statistics also of bigrams (sequences of two words), trigrams, etc. A similar idea may be used to compare graphs by ignoring large-scale structure and viewing graphs as bags of vertices or edges. The \new{vertex label kernel} does precisely this by comparing graphs only at the level of similarity between all pairs of vertex labels from two different graphs,
$$
k_{\text{VL}}(G, H) = \sum_{u \in V(G)}\sum_{v \in V(G)} k(l(u), l(v))\,.
$$
With the base kernel $k$ the equality indicator function, $k_{\text{VL}}$ is a linear kernel on the (unnormalized) distributions of vertex labels in $G$ and $H$. Similar in spirit, the \new{edge label kernel} is defined as the sum of base kernel evaluations on all pairs of edge labels (or triplets of the edge label and  incident vertex labels).
Note that such kernels are a paramount example for instances of the convolution kernel framework, see~\cref{sec:convkernel}.

A downside of vertex and edge label kernels is that they ignore the interplay between structure and labels and are almost completely uninformative for unlabeled graphs. Instead of viewing graphs as bags of vertices or edges, we may view them as bags of \new{subgraph patterns}. To this end, \citet{She+2009} introduced a kernel based on counting occurrences of subgraph patterns of a fixed size---so called~\new{graphlets} (see Figure~\ref{fig:graphlet}). Every graphlet is an instance of an isomorphism type---a set of graphs that are all isomorphic---such as a graph on three vertices with two edges. While there are three graphs that connect three vertices with two edges, they are all isomorphic and considered equivalent as graphlets.

Graphlet kernels count the isomorphism types of all induced (possibly disconnected) subgraphs on $k > 0$ vertices of a graph $G$. Let $\phi(G)_{\sigma_i}$ for $1 \leq i \leq N$ denote the number of instances of  isomorphism type $\sigma_i$ where $N$ denotes the number of different types. The kernel computes a feature map $\phi_{\text{GR}}(G)$ for $G$,
\begin{equation*}
	\phi_{\text{GR}}(G) = \mleft(\phi(G)_{\sigma_1}, \dotsc, \phi(G)_{\sigma_N} \mright)\,.
\end{equation*}
The graphlet kernel is finally defined as $k_{\text{GR}}(G,H) = \langle \phi_{\text{GR}}(G), \phi_{\text{GR}}(H) \rangle$ for two graphs $G$ and $H$.

The time required to compute the graphlet kernel scales exponentially with the size of the considered graphlets. To remedy this, \citet{She+2009} proposed two algorithms for speeding up the computation time of the feature map for $k$ in $\{3,4\}$. In particular, it is common to restrict the kernel to connected graphlets (isomorphism types). Additionally, the statistics used by the graphlet kernel may be estimated approximately by subgraph sampling, see, e.g.,~\cite{johansson2015classifying,Ahm+2016,Che+2016,Bre+2017}.
Please note that the graphlet kernel as proposed by \citet{She+2009} does not consider any labels or attributes. However, the concept (but not all speed-up tricks) can be extended to labeled graphs by using labeled isomorphism types as features, see, e.g., \citep{Wale2008a}.
Mapping (sub)graphs to their isomorphism type is known as \emph{graph canonization problem}, for which no polynomial time algorithm is known~\citep{Johnson2005}. However, this is not a severe restriction for small graphs such as graphlets and, in addition, well-engineered algorithms solving most practical instances in a short time exist~\cite{McKay2014}.
\citet{Hor+2004} proposed a kernel which decomposes graphs into cycles and tree patterns, for which the canonization problem can be solved in polynomial time and simple practical algorithms for this are known.

\citet{Cos+2010} introduced the \emph{neighborhood subgraph pairwise distance kernel}
which associates a string with every vertex representing its neighborhood up to
a certain depth. In order to avoid solving the graph canonization problem, they
proposed using a graph invariant that may, in rare cases, map
non-isomorphic neighborhood subgraphs to the same string. Then, pairs of these
neighborhood graphs together with the shortest-path distance between their central vertices are counted as features. The approach is similar to the Weisfeiler-Lehman shortest-path kernel (see \cref{sec:naa}).

An alternative to subgraph patterns, tree patterns may contain repeated vertices just like random walks and were initially proposed for use in graph comparison by \citet{Ram+2003} and later refined by \citet{Mah+2009}. Tree pattern kernels are similar to the Weisfeiler-Lehman subtree kernel, but do not
consider all neighbors in each step, but also all possible
subsets~\citep{She+2011}, and hence do not scale to larger datasets.
\citet{Mar+2012} proposed decomposing a graph into trees and applying a kernel defined on trees. In~\citep{Mar+2012a}, a fast hashing-based computation scheme for the aforementioned graph kernel is proposed.

\subsection{Walks and Paths}

A downside of the subgraph pattern kernels described in the previous section is that they require the specification of a set of patterns, or subgraph size, in advance. To ensure efficient computation, this often restricts the patterns to a fairly small scale, emphasizing local structure. A popular alternative is to compare the sequences of vertex or edge attributes that are encountered through traversals through graphs. In this section, we describe two families of traversal algorithms which yield different attribute sequences and thus different kernels---shortest paths and random walks.

\subsubsection{Shortest-path kernels}
One of the very first, and most influential, graph kernels is the \new{shortest-path} (SP) kernel~\cite{Borgwardt2005}. The idea of the SP kernel is to compare the attributes and lengths of the shortest paths between all pairs of vertices in two graphs. The shortest path between two vertices is illustrated in Figure~\ref{fig:fundamentals}. Formally, let $G$ and $H$ be graphs with label function $l\colon V(G) \cup V(H) \to \Sigma$ and let $d(u,v)$ denote the shortest-path distance between the vertices $u$ and $v$ in the same graph. Then, the kernel is defined as
\begin{equation}\label{spk}
	k_{\text{SP}}(G,H) = \sum_{\substack{(u,v) \in V(G)^2\\ u \neq v}} \sum_{\substack{(w,z) \in V(H)^2\\ w \neq z}} k((u,v),(w,z)),
\end{equation}
where
\begin{equation*}\label{ksp}
	k((u,v),(w,z)) = k_{\text{L}}(l(u), l(w)) \cdot k_{\text{L}}(l(v), l(z)) \cdot k_{\text{D}}(d(u,v), d(w,z))\,.
\end{equation*}
Here, $k_{\text{L}}$ is a kernel for comparing vertex labels and $k_{\text{D}}$ is a kernel to compare shortest-path distances, such that $k_{\text{D}}(d(u,v), k(w,z)) = 0$ if $d(u,v) = \infty$ or $d(w,z) = \infty$.

The running time for evaluating the general form of the SP kernel for a pair of graphs is in $\cO(n^4)$. This is prohibitively large for most practical applications. However, in the case of discrete vertices and edge labels, e.g., a finite subset of the natural numbers, and $k$ the indicator function, we can compute the feature map $\phi_{\text{SP}}(G)$ corresponding to the kernel explicitly. In this case, each component of the feature map counts the number of triples $(l(u), l(v), d(u,v))$ for $u$ and $v$ in $V(G)$ and $u \neq v$. Using this approach, the time complexity of the SP kernel is reduced to the time complexity of the Floyd-Warshall algorithm, which is in $O(n^3)$. In~\cite{Her+2015} the shortest-path is generalized by considering all shortest paths between two vertices.

\subsubsection{Random walk kernels}
\citet{Gae+2003} and \citet{Kas+2003} simultaneously proposed graph kernels
based on random walks, which count the number of (label sequences along) walks that two graphs have in
common. The description of the random walk kernel by \citet{Kas+2003} is motivated by
a probabilistic view of kernels and based on the idea of so-called
\emph{marginalized kernels}. The feature space of the kernel comprises all possible label sequences produced by random walks; since the length of the walks is unbounded, the space is of infinite dimension.
A method of computation is proposed based on a recursive reformulation of
the kernel, which at the end boils down to finding the stationary state of a
discrete-time linear system. Since this kernel was later generalized
by~\cite{Vis+2010} we do not go into the mathematical
details of the original publication.
The approach fully supports attributed graphs, since vertex and edge labels
encountered on walks are compared by user-specified kernels.

\citet{Mah+2004} extended the original formulation of random walk kernels
with a focus on application in cheminformatics~\cite{Mah+2005} to improve the
scalability and relevance as similarity measure. A mostly unfavorable characteristic of random walks is that they may visit the same
vertex several times. Walks are even allowed to traverse an edge from $u$ to $v$ and instantly return to $u$ via the same edge, a problem referred to as
\new{tottering}. These repeated consecutive vertices do not provide useful
information and may even harm the validity as similarity measure.
Hence, the marginalized graph kernel was extended to avoid tottering by
replacing the underlying first-order Markov random walk model by a
second-order Markov random walk model. This technique to prevent tottering only eliminates walks $(v_1, \dots, v_n)$
with $v_i = v_{i+2}$ for some $i$, but it does not require the considered walks
to be paths, i.e., repeated vertices still occur.

Like other random walk kernels, \citet{Gae+2003} define the feature space of their kernel as the label sequences derived from walks, but propose a different
method of computation based on the \new{direct product graph} of two labeled input graphs.
\begin{definition}[Direct Product Graph] \label{def:direct-productgraph}
	For two labeled graphs
	$G=(V,E)$ and $H=(V',E')$ the \emph{direct product graph} is
	denoted by $G \times H = (\mathcal{V}, \mathcal{E})$ and defined as
	\begin{align*}
		\mathcal{V} = & \left\{ (v,v') \in V \times V' \mid l(v) = l(v') \right\}                                                                 \\
		\mathcal{E} = & \left\{ ((u,u'),(v,v')) \in {\mathcal{V}} \mid (u,v) \in E \wedge (u',v') \in E' \wedge l((u,v)) = l((u',v')) \right\}\,.
	\end{align*}
	A vertex (edge) in $G \times H$ has the same label as the corresponding vertices
	(edges) in $G$ and $H$.
\end{definition}
\begin{figure}
	\includegraphics{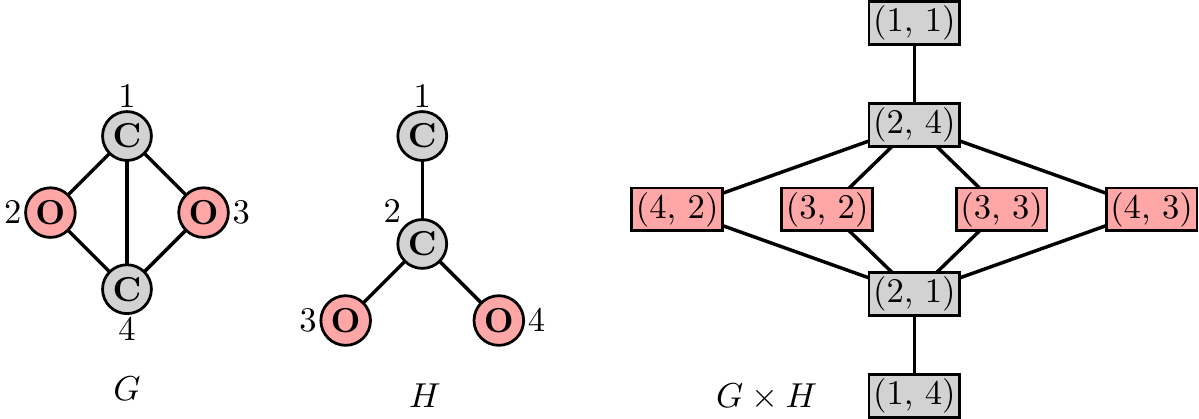}
	\caption{{Direct product graph.} Two labeled graphs $G$, $H$ and their
   direct product graph $G \times H$. The vertices of $G$ and $H$ are labeled
   with 'C' (gray) and 'O' (red). In the direct product graph, there is a vertex
   for all pairs of vertices of $G$ and $H$ with the same label. Two vertices in
   the direct product graph are adjacent if and only if the associated pairs of
   vertices are adjacent in $G$ and $H$.
   }\label{fig:directproductgraph}
\end{figure}
The concept is illustrated in Figure~\ref{fig:directproductgraph}.
There is a one-to-one correspondence between walks in $G \times H$ and walks in
the graphs $G$ and $H$ with the same label sequence.
The \emph{direct product kernel} is then defined as
\begin{equation}\label{rw}
	K_{\text{RW}}(G, H) = \sum_{i,j=1}^{|\mathcal{V}|} \left[\sum_{l=0}^\infty \lambda_l \vec{A}_\times^l \right]_{ij},
\end{equation}
where $\vec{A}_\times$ is the adjacency matrix of $G \times H$ and $\lambda=(\lambda_0,
\lambda_1,\dots)$ a sequence of weights such that the above sum converges. This is the case for $\lambda_i = \gamma^i$, $i \in \bbN$, and $\gamma < \frac{1}{a}$ with $a \geq \Delta$, where $\Delta$ is the maximum degree of $G \times H$.
For this choice of weights and with $\vec{I}$ the identity matrix, there exists a closed-form expression,
\begin{equation}\label{eq:kernel:geometric-rw}
	K_{\text{GRW}}(G, H) = \sum_{i,j=1}^{|\mathcal{V}|}\left[ (\vec{I} - \gamma \vec{A}_\times)^{-1} \right]_{ij}
\end{equation}
which can be computed by matrix inversion. Since the expression reminds of the geometric series transferred
to matrices, Equation~\eqref{eq:kernel:geometric-rw} is referred to as
\emph{geometric random walk kernel}.
The running time to compute the geometric random walk kernel between two graphs
is dominated by the inversion of the adjacency matrix associated with the direct
product graph. The running time is given as roughly
$\cO(n^6)$~\cite{Vis+2010}.

\citet{Vis+2010} propose a generalizing framework for random walk based
graph kernels and argue that the approach by~\citet{Kas+2003} and~\citet{Gae+2003} can be considered special cases of this kernel. The paper does not address vertex labels and makes extensive use of the Kronecker product between matrices denoted by $\otimes$ and lifts it to the feature space associated with an (edge) kernel.
Given an edge kernel $\kappa_{\text{E}}$ on attributes from the set $\mathcal{A}$, let
$\phi \colon \mathcal{A} \to \cH$ be a feature map.
For an attributed graph $G$, the feature matrix $\Phi(G)$ is then defined as
$\Phi_{ij}(G) = \phi((v_i,v_j))$ if $(v_i,v_j) \in E(G)$ and
$\vec{0}$ otherwise. Then, $\vec{W}_\times = \Phi(G) \otimes \Phi(H)$
yields a weight matrix of the direct product graph $G \times H$.\footnote{Here
	vertex labels are ignored, i.e., $V(G \times H) = V(G) \times V(H)$.}
The proposed kernel is defined as
\begin{equation}
	K_{\text{RW}}(G,H) = \sum_{l=0}^{\infty} \mu_l \vec{q}^T_\times \vec{W}^l_\times \vec{p}_\times,
\end{equation}
where $\vec{p}_\times$ and $\vec{q}_\times$ are initial and stopping probability
distributions and $\mu_l$ coefficients such that the sum converges.
Several methods of computation are proposed, which yield different running times
depending on a parameter $l$, specific to that approach. The parameter $l$ either
denotes the number of fixed-point iterations, power iterations or the effective
rank of $\vec{W}_\times$.
The running times to compare graphs of order $n$ also depend on the edge labels
of the input graphs and the desired edge kernel:
For unlabeled graphs the running time $\cO(n^3)$ is achieved and $\cO(dln^3)$
for labeled graphs, where $d = |\mathcal{L}|$ is the size of the label
alphabet.
The same running time is attained by edge kernels with a $d$-dimensional
feature space, while $\cO(ln^4)$ time is required in the infinite case. For
sparse graphs, $\cO(ln^2)$ is achieved in all cases, where a graph $G$ is said
to be sparse if $|E(G)|=\cO(|V(G)|)$.
Further improvements of the running time were subsequently achieved by non-exact
algorithms based on low rank approximations~\cite{Kan+2012}.
Recently, the phenomenon of \emph{halting} in random walk kernels has been
studied~\citet{Sug+2015}, which refers to the fact that
walk-based graph kernels may down-weight longer walks so much that their value
is dominated by walks of length $1$.

The classical random walk kernels described above in theory take all walks
without a limitation in length into account, which leads to a high-dimensional
feature space. Several application-related papers used walks up to a certain
length only, e.g., for the prediction of protein functions \citep{Bor+2005a} or
image classification \citep{Har+2007}. These walk based kernels are not
susceptible to the phenomenon of halting.
\citet{Kri+2014,Kriege2019dami} systematically studied kernels based on all the
walks of a predetermined fixed length $\ell$, referred to as \emph{$\ell$-walk
kernel}, and all the walks with length at most $\ell$, called
\emph{Max-$\ell$-walk kernel}, respectively. For these, computation schemes based
on implicit and explicit feature maps were proposed and compared experimentally.
Computation by explicit feature maps provides a better performance for graphs
with discrete labels with a low label diversity and small walk lengths.
Conceptually different, \citet{Zha+2018RetGK} derived graph kernels based on return
probabilities of random walks.

\subsection{Kernels for Graphs with Continuous Labels}

Most real-world graphs have attributes, mostly real-valued vectors, associated with their vertices and edges. For example, atoms of chemical molecules have physical and chemical properties; individuals in social networks have demographic information; and words in documents carry semantic meaning. Kernels based on pattern counting or neighborhood aggregation are of a discrete nature, i.e., two vertices are regarded as similar if and only if they exactly match, structure-wise as well as attribute-wise. However, in most applications it is desirable to compare real-valued attributes with more nuanced similarity measures such as the Gaussian RBF kernel of~\cref{eq:rbf}.

Kernels suitable for attributed graphs typically rely on user-defined kernels for the comparison of vertex and edge labels. These kernels are then combined with kernels on structure through operations that yield a valid kernel on graphs, such as addition or multiplication. Two examples of this, the recently proposed kernels for attributed graphs,
\new{GraphHopper}~\cite{Fer+2013} and \new{GraphInvariant}~\cite{Ors+2015}, can be expressed as
\begin{equation}
	k_{\text{WV}}(G,H) =\! \sum_{v \in V(G)} \sum_{v' \in V(H)} k_W(v,v') \cdot k_V(v,v').
\end{equation}
Here, $k_V$ is a user-specified kernel comparing vertex attributes and $k_W$ is
a kernel that determines a weight for a vertex pair based on the individual graph
structures.
Kernels belonging to this family are easily identifiable as instances of
$R$-convolution kernels, cf.~Definition~\ref{conv}.

For graphs with real-valued attributes, one could set $k_V$ to the Gaussian RBF kernel. The selection of the kernel $k_W$ is essential to take the graph structure into account and allows to obtain different instances of weighted vertex kernels. One implementation of $k_W$ motivated along the lines of
\new{GraphInvariant}~\cite{Ors+2015} is
\begin{equation*}
	k_W(v,v') = \sum_{i=0}^h k_\delta(\tau_i(v),\tau_i(v')),
\end{equation*}
where $\tau_i(v)$ denotes the discrete label of the vertex $v$ after the $i$-th
iteration of Weisfeiler-Lehman label refinement of the underlying unlabeled
graph. Intuitively, this kernel reflects to what extent the two vertices have a
structurally similar neighborhood.

Another graph kernel, which fits into the framework of weighted vertex kernels,
is the GraphHopper kernel~\cite{Fer+2013} with
\begin{equation*}
	k_W(v,v') = \langle \vec{M}(v), \vec{M}(v') \rangle_F\,.
\end{equation*}
Here $\vec{M}(v)$ and $\vec{M}(v')$ are $\delta \times \delta$ matrices, where
the entry $\vec{M}(v)_{ij}$ for $v$ in $V(G)$ counts the number of times the
vertex $v$ appears as the $i$-th vertex on a shortest path of discrete length $j$
in $G$, where $\delta$ denotes the maximum diameter over all graphs, and
$\langle \cdot, \cdot \rangle_F$ is the Frobenius inner product.

\citet{Kri+2012} proposed the \new{subgraph matching kernel} which is computed by considering all bijections between all subgraphs on at most $k$ vertices, and allows to compare vertex attributes using a custom kernel. Moreover, in~\cite{Su+2016} the \new{Descriptor Matching kernel} is defined, which captures the graph structure by a propagation mechanism between neighbors, and uses a variant of the pyramid match kernel~\cite{Gra+2006} to compare attributes between vertices. The kernel can be computed in time linear in the number of edges.

\citet{Mor+2016} introduced a scalable framework to compare attributed graphs. The idea is to iteratively turn the continuous attributes of a graph into discrete labels using randomized hash
functions. This allows to apply fast explicit graph feature maps, which are limited to graphs with discrete annotations such as the one associated with the Weisfeiler-Lehman subtree kernel~\cite{She+2011}. For special hash functions, the authors obtain approximation results for several state-of-the-art kernels which can handle continuous information. Moreover, they derived a variant of the Weisfeiler-Lehman subtree kernel which can handle continuous attributes.

\subsection{Other Approaches}
\label{sec:other}
\citet{Kon+2009} derived a graph kernel using graph invariants based on group representation theory. In~\cite{Kon+2016}, a graph kernel is proposed which is able to capture the graph structure at multiple scales, i.e., neighborhoods around vertices of increasing depth, by using ideas from spectral graph theory. Moreover, the authors provide a low-rank approximation algorithm to scale the kernel computation to large graphs. \citet{Joh+2014} define a graph kernel based on the the Lovász number~\cite{Lov+1979} and provide algorithms to approximate this kernel.

In~\cite{Li+2015}, a kernel for dynamic graphs is proposed, where vertices and edges are added or deleted over time. The kernel is based on eigen decompositions. \citet{Kri+2014,Kriege2019dami} investigated under which conditions it is possible and more efficient to compute the feature map corresponding to a graph kernel explicitly. They provide theoretical as well as empirical results for walk-based kernels. \citet{Li+2012} proposed a streaming version of the Weisfeiler-Lehman algorithm using a hashing technique. \citet{Aio+2015} and \citet{Massimo2016} applied multiple kernel learning to the graph kernel domain.
\citet{Nikolentzos2018} proposed to first build the $k$-core decomposition of
graphs to obtain a hierarchy of nested subgraphs, which are then individually
compared by a graph similarity measure. The approach has been combined with
several graph kernels such as the Weisfeiler-Lehman subtree kernel and was shown
to improve the accuracy on some datasets.

\citet{Yan+2015} uses recent neural techniques from neural language modeling, such as \new{skip-gram}~\cite{Mik+2013}. The authors build on known state-of-the-art kernels, but allow to respect relationships between
their features. This is demonstrated by hand-designed matrices encoding the similarities between features for selected graph kernels such as the graphlet and Weisfeiler-Lehman subtree kernel.
Similar ideas were used in~\cite{Yan+2015a} where smoothing methods for multinomial distributions were applied to the graph domain.

\section{Expressivity of Graph Kernels}
\label{sec:theory}
While a large literature has studied the empirical performance of various graph kernels, there exists comparatively few works that deal with graph kernels exclusively from a theoretical point of view. Most works that provide learning guarantees for graph kernels attempt to formalize their \new{expressivity}.

The \new{expressivity} of a graph kernel refers broadly to the kernel's ability to distinguish certain patterns and properties of graphs. In an early attempt to formalize this notion, \citet{Gae+2003} introduced the concept a \new{complete graph kernel}---kernels for which the corresponding feature map is an injection. If a kernel is not complete, there are non-isomorphic graphs $G$ and $H$ with $\phi(G) = \phi(H)$ that cannot be distinguished by the kernel. In this case there is no way any classifier based on this kernel can separate these two graphs. However, computing a complete graph kernel is \cGI-hard, i.e., at least as hard as deciding whether two graphs are isomorphic~\citep{Gae+2003}. For this problem no polynomial time algorithm for general graphs is known~\cite{Johnson2005}. Therefore, none of the graph kernels used in practice are complete. Note however, that a kernel may be injective with respect to a finite or restricted family of graphs.

As no practical kernels are complete, attempts have been made to characterize expressivity in terms of which graph properties can be distinguished by existing graph kernels. In~\cite{Kri+2018}, a framework to measure the expressivity of graph kernels based on ideas from property testing was introduced. The authors show that graph kernels such as the Weisfeiler-Lehman subtree, the shortest-path and the graphlet kernel are not able to distinguish basic graph properties such as planarity or connectedness. Based on these results they propose a graph kernel based on frequency counts of the isomorphism type of subgraphs around each vertex up to a certain depth. This kernel is able to distinguish the above properties and computable in polynomial time for graphs of bounded degree. Finally, the authors provide learning guarantees for 1-nearest neighborhood classifiers. Similarly, \cite{Joh+2015} gave bounds on the classification margin obtained when using the optimal assignment kernel, with Laplacian embeddings, to classify graphs with different densities or random graphs with and without planted cliques. In~\citet{Joh+2014}, the authors studied global properties of graphs such as girth, density and clique number and proposed kernels based on vertex embeddings associated with the Lovász-$\vartheta$ and SVM-$\vartheta$ numbers which have been shown to capture these properties.

The expressivity of graph kernels has been studied also from statistical perspectives. In particular, \citet{One+2017} use well-known results from statistical learning theory to give results which bound measures of expressivity in terms of Rademacher complexity and stability theory. Moreover, they apply their theoretical findings in an experimental study comparing the estimated expressivity of popular graph kernels, confirming some of their known properties. Finally, \citet{johansson2015classifying} studied the statistical tradeoff between expressivity and differential privacy~\citep{dwork2014algorithmic}.

\section{Applications of Graph Kernels}
\label{sec:applications}
The following section outlines a \emph{non-exhaustive} list of applications of the kernels described in Section~\ref{main}, categorized by scientific area.

\paragraph{Chemoinformatics.}\label{sec:app:chem}
Chemoinformatics is the study of chemistry and chemical compounds using statistical and computational resources~\citep{Brown2009}. An important application is drug development in which  new, untested medical compounds are modeled in silico before being tested in vitro or in animal tests. The primary object of study---the molecule---is well represented by a graph in which vertices take the places of atoms and edges that of bonds. The chemical properties of these atoms and bonds may be represented as vertex and edge attributes, and the properties of the molecule itself through features of the  structure and attributes.
The graphs derived from small molecules have specific characteristics. They typically have
less than 50 vertices, their degree is bounded by a small constant ($\leq 4$ with few
exceptions), and the distribution of vertex labels representing atom types is specific
(e.g., most of the atoms are carbon).
Almost all molecular graphs are planar, most of them even outerplanar~\citep{Horvath2010a},
and they have a tree-like structure~\citep{Yamaguchi2003a}.
Molecular graphs are not only a common benchmark dataset for
graph kernels, but several kernels were specifically proposed for this
domain, e.g.,~\citep{Hor+2004,Swa+2005,Ceroni2007,Mah+2009,Fro+2005}.
The \emph{pharmacophore kernel} was introduced by \citet{Mah+2006} to compare chemical compounds based on characteristic features of vertices together with their relative spatial arrangement. As a result, the kernel is designed to handle with continuous distances. The pharmacophore kernel was shown to be an instance of the more general subgraph matching kernel~\citep{Kri+2012}.
\citet{Mah+2009} developed new tree pattern kernels for molecular graphs, which
were then applied in toxicity and anti-cancer activity prediction tasks.
Kernels for chemical compounds such as this have been successfully employed for various tasks
in cheminformatics including the prediction of mutagenicity, toxicity and
anti-cancer activity~\cite{Swa+2005}.

However, such tasks have been addressed by computational methods long before the
advent of graph kernels, cf.~Figure~\ref{timeline}. So-called \emph{fingerprints}
are a well-established classical technique in cheminformatics to represent
molecules by feature vectors~\citep{Brown2009}.
Commonly features are obtained by
\begin{inparaenum}[(i)]
 \item enumeration of all substructures of a certain class contained in the
   molecular graphs,
 \item taken from a predefined dictionary of relevant substructures or
 \item generated in a preceding data-mining phase.
\end{inparaenum}
Fingerprints are then used to encode the number of occurrences of a feature or
only its presence or absence by a single bit per feature.
Often hashing is used to reduce the fingerprint length to a fixed size at the
cost of information loss \citep[see, e.g.,][]{Daylight2008}.
Such fingerprints are typically compared using similarity measures such as
the \emph{Tanimoto coefficient}, which are closely related to kernels~\citep{Ralaivola2005}.
Approaches of the first category are, e.g., based on all paths contained in a
graph~\citep{Daylight2008} or all subgraphs up to a certain size~\citep{Wale2008a},
similar to graphlets.
\citet{Ralaivola2005} experimentally compared random walk kernels to kernels
derived from path-based fingerprints and has shown that these reach similar
classification performance on molecular graph datasets.
\emph{Extended connectivity fingerprints} encode the neighborhood of atoms
iteratively similar to the graph kernels discussed in Section~\ref{sec:naa} and
can be considered a standard tool in cheminformatics for decades~\citep{Rogers2010}.
Predefined dictionaries compiled by experts with domain-specific knowledge exist,
e.g., MACCS/MDL Keys for drug discovery~\citep{Durant2002}.

\paragraph{Bioinformatics.}
Understanding proteins, one of the fundamental building blocks of life, is a central goal in bioinformatics. Proteins are complex molecules which are often represented in terms of larger components such as helices, sheets and turns.
\citet{Bor+2005a} model protein data as graphs where each vertex represents such a component, and each edge indicates proximity in space or in amino acid sequence. Both vertices and edges are annotated by categorical and real-valued attributes. The authors used a modified random walk kernel to classify proteins as enzymes or non-enzymes.
In related work, \citet{borgwardt2007graph} predict disease outcomes from protein-protein interaction networks. Here, each vertex is a protein and each edge the physical interaction between a protein-protein pair. In order to take missing edges into account, which is crucial for studying protein-protein-interaction networks, the kernel
\begin{equation*}\label{eq:kernel:ppi}
	K_{\text{CP}} (G,H) = K_{\text{RW}}(G,H) + K_{\text{RW}}(\overline{G},\overline{H}),
\end{equation*}
was proposed, which is the sum of a random walk kernel $K_{\text{RW}}$ applied
to the original graphs $G$ and $H$ as well as to their complement graphs $\overline{G}$ and $\overline{H}$. Studying pairs of complement graphs may be useful also in other applications.

\paragraph{Neuroscience.}
The connectivity and functional activity between neurons in the human brain are indicative of diseases such as Alzheimer's disease as well as subjects' reactions to sensory stimuli. For this reason, researchers in neuroscience have studied the similarities of brain networks among human subjects to find patterns that correlate with known differences between them. Representing parts of the brain as vertices and the strength of connection between them as edges, several authors have applied graph kernels for this purpose~\citep{Vega-Pons2014,Takerkart2014,Vega-Pons2013,Wang2016,Jie2016}. Unlike many other applications, the vertices in brain networks often have an identity, representing a specific part of the brain. \citet{Jie2016} exploited this fact in learning to classify mild cognitive impairments (MCI). They find that their proposed kernel, based on iterative neighborhood expansion (similar to the Weisfeiler-Lehman kernel), which exploits the one-to-one mapping of vertices (brain regions) between different graphs consistently outperforms baseline kernels in this task.

\paragraph{Natural language processing.}
Natural language processing is ripe with relational data: words in a document relate through their location in text, documents relate through their publication venue and authors, named entities relate through the contexts in which they are mentioned. Graph kernels have been used to measure similarity between all of these concepts. For example, \citet{Nik+2017a} use the shortest-path kernel to compute document similarity by converting each document to a graph in which vertices represent terms and two vertices are connected by an edge if the corresponding terms appear together in a fixed-size window.
\citet{hermansson2013entity} used the co-occurrence network of person names in a large news corpus to classify which names belong to multiple individuals in the database. Each name was represented by the subgraph corresponding to the  neighborhood of co-occuring names and labeled by domain experts. The output of the system was intended for use as preprocessing to an entity disambiguation system.
In~\cite{Li+2016} the Weisfeiler-Lehman subtree kernel was used to define a similarity function for call graphs of Java programs to identify similar call graphs.
\citet{Vri+2013} extended the Weisfeiler-Lehman subtree kernel so that it can handle RDF data.

\paragraph{Computer vision.}
\citet{Har+2007} applied kernels based on walks of a fixed length to image
classification and developed a dynamic programming approach for their computation.
The also modified tree pattern kernels for image classification, where
graphs typically have a fixed embedding in the plane.
\citet{Wu2014} proposed graph kernels for human action recognition in video
sequences. To this end, they encode the features of each frame as well as the
dynamic changes between successive frames by separate graphs. These graphs are
compared by a linear combination of random walk kernels using multiple kernel
learning, which leads to an accurate classification of human actions.
The propagation kernel was applied to predict object categories in order to
facilitate robot grasping~\cite{Neumann2013a}. To this end, 3D point cloud data
was represented by $k$-nearest neighbor graphs.

\section{Experimental Study}\label{exp}

In our experimental study, we investigate various kernels considered to be
state-of-the-art in detail and compare them to simple baseline methods using
vertex and edge label histograms.
We would like to answer the following research questions.

\begin{axioms}{Q}
	\item \textbf{Expressivity.} Are the proposed graph kernels sufficiently expressive to distinguish the
	graphs of common benchmark datasets from each other according to their labels and
	structure? \label{h:complete}
	\item \textbf{Non-linear decision boundaries.} Can the classification accuracy of graph kernels be improved by
	finding non-linear decision boundaries in their feature space? \label{h:rbf}
	\item \textbf{Accuracy.} Is there a graph kernel that is superior over the other graph kernels
	in terms of classification accuracy?
	Does the answer to \ref{h:complete} explain the differences in prediction accuracy? \label{h:sup}
	\item \textbf{Agreement.} Which graph kernels predict similarly? Do different graph kernels succeed and fail for the same graphs? \label{h:qsim}
	\item \textbf{Continuous attributes.} Is there a kernel for graphs with continuous attributes that is superior over the other graph kernels in terms of classification accuracy? \label{h:cont}
\end{axioms}

\subsection{Methods}
We describe the methods we used to answer the research questions and summarize
our experimental setup.

\subsubsection{Classification Accuracy}
In order to answer several of our research questions, it is necessary to determine
the prediction accuracy achieved by the different graph kernels.
We performed classification experiments using the $C$-SVM implementation
LIBSVM~\cite{Cha+2011}.
We used nested cross-validation with $10$ folds in the inner and outer loop.
In the inner loop the kernel parameters and the regularization parameter $C$ were
chosen by cross-validation based on the training set for the current fold.
In the same way it was determined whether the kernel matrix should be normalized.
The parameter $C$ was chosen from $\{10^{-3}, 10^{-2}, \dots, 10^3\}$.
We repeated the outer cross-validation ten times with different random folds, and
report average accuracies and standard deviations.

\subsubsection{Complete Graph Kernels}
The theoretical concept of complete graph kernels has little practical relevance
and is not suitable for answering \ref{h:complete}.
Therefore we generalize the concept of complete graph kernels.
For a given dataset $\mathcal{D} = \{(G_1,y_1),\dots,(G_n,y_n)\}$ of graphs
$G_i$ with class labels $y_i \in \mathcal{Y}$ for all $1 \leq i \leq n$, we say
a graph kernel $K$ with a feature map $\phi$  is \emph{complete for $\mathcal{D}$}
if for all graphs  $G_i,G_j$ the implication
$\phi(G_i) = \phi(G_j) \Longrightarrow i = j$ holds;
it is \emph{label complete for $\mathcal{D}$} if for all graphs  $G_i,G_j$
the implication $\phi(G_i) = \phi(G_j) \Longrightarrow y_i = y_j$ holds.
Note that we may test whether $\phi(G_i) = \phi(G_j)$ holds using the kernel trick
without constructing the feature vectors.
For a kernel $K$ on $\mathcal{X}$ with a feature map $\phi : \mathcal{X} \to \cH$
the \emph{kernel metric} is
\begin{align}
	d_K(x,y) & = \norm{\phi(x)-\phi(y)} \label{eq:kerneldistance:expl}            \\
	         & = \sqrt{K(x,x) + K(y,y) - 2K(x,y)}. \label{eq:kerneldistance:impl}
\end{align}
Therefore, $\phi(G) = \phi(H)$ if and only if $K(G,G) + K(H,H) - 2K(G,H) = 0$.
We define the \emph{(label) completeness ratio} of a graph kernel w.r.t.\@ a
dataset as the fraction of graphs in the dataset that can be distinguished from
all other graphs (with different class labels) in the dataset.

We investigate how these measures align with the observed prediction accuracy.
Note that the label completeness ratio limits the accuracy of a kernel on a
specific dataset.
Vice versa, classifiers based on complete kernels not necessarily achieve a high
accuracy.
A kernel that is one for two isomorphic graphs and zero otherwise, for example,
would achieve the highest possible completeness ratio, but is too strict for
learning, cf.\@ Section~\ref{sec:convkernel}.
Moreover, a complete graph kernel not necessarily maps graphs in different classes
to feature vectors that are linearly separable.
In this case (an additional) mapping in a high-dimensional feature space might
improve the accuracy.

\subsubsection{Non-linear Decision Boundaries in the Feature Space of Graph Kernels}\label{sec:rbf}
Many graph kernels explicitly compute feature vectors and thus essentially
transform graph data to vector data, cf.~Section~\ref{main}.
Typically, these kernels then just apply the linear kernel to these vectors to
obtain a graph kernel. This is surprising since it is well-known
that for vector data often better results can be obtained by a polynomial
or Gaussian RBF kernel. These, however, are usually not used in combination with
graph kernels.
\citet{Sug+2015} observed that applying a Gaussian RBF kernel to vertex and edge
label histograms leads to a clear improvement over linear kernels. Moreover, for
some datasets the approach was observed to be competitive with random walk kernels.
Going beyond the application of standard kernels to graph feature vectors,
\citet{Kri2015} proposed to obtain modified graph kernels also from those based
on implicit computation schemes by employing the kernel trick, e.g., by
substituting the Euclidean distance in the Gaussian RBF kernel by the metric
associated with a graph kernel.
Since the kernel metric can be computed without explicit feature maps, any graph
kernel can thereby be modified to operate in a different (high-dimensional) feature
space.
However, the approach was generally not employed in experimental evaluations of
graph kernels. Only recently, \citet{Nikolentzos2018a} presented first
experimental results of the approach for the shortest-path, Weisfeiler-Lehman and
pyramid match graph kernel using a polynomial and Gaussian RBF kernel for
successive embedding. Promising experimental results were presented, in particular,
for the Gaussian RBF kernel. We present an in detail evaluation  of the approach
on a wide range of graph kernels and datasets.

We apply the Gaussian RBF kernel to the feature vectors associated with graph kernels by
substituting the Euclidean distance in \cref{eq:rbf} by the metric associated
with graph kernels.
Note that the kernel metric can be computed from feature vectors according to
\cref{eq:kerneldistance:expl} or by employing the kernel trick according to
\cref{eq:kerneldistance:impl}.
In order to study the effect of this modification experimentally, we have
modified the computed kernel matrices as described above.
The parameter $\sigma$ was selected from $\{2^{-7},2^{-6},\dots,2^7\}$ by
cross-validation in the inner cross-validation loop based on the training data
sets.

\subsection{Datasets}
In our experimental evaluation, we have considered graph data from various
domains, most of which has been used previously to compare graph kernels.
Moreover, we derived new large datasets from the data published by the National
Center for Advancing Translational Sciences in the context of the
\emph{Tox21 Data Challenge 2014}\footnote{\url{https://tripod.nih.gov/tox21/challenge/}}
initiated with the goal to develop better toxicity assessment methods for small
molecules. These datasets each contain more than 7000 graphs and thus exceed the
size of the datasets typically used to evaluate graph kernels.
We have made all datasets publicly available~\cite{KKMMN2016}.
Their statistics are summarized in \cref{tab:datasets}.

The datasets AIDS, BZR, COX2, DHFR, Mutagenicity, MUTAG, NCI1, NCI109, PTC and
Tox21 are graphs derived from small molecules, where class labels encode a
certain biological property such as toxicity and activity against cancer cells.
The vertices and edges of the graphs represent the atoms and their chemical
bonds, respectively, and are annotated by their atom and bond type.
The datasets DD, ENZYMES and PROTEINS represent macromolecules using different
graph models. Here, the vertices either represent protein tertiary structures or amino
acids and the edges encode spatial proximity. The class labels are the 6
EC top-level classes or encode whether a protein is an enzyme.
The datasets REDDIT-BINARY, IMDB-BINARY and IMDB-MULTI are derived
from social networks.
The MSRC datasets are associated with computer vision tasks. Images are encoded
by graphs, where vertices represent superpixels with a semantic label and edges
their adjacency.
Finally, SYNTHETICnew and Synthie are synthetically generated graphs with
continuous attributes. FRANKENSTEIN contains graphs derived from small molecules,
where atom types are represented by high dimensional vectors of pixel intensities
of associated images.

\begin{table*}
	\setlength{\tabcolsep}{2.4pt}
	\begin{center}
		\caption{Dataset statistics and properties.}
		\label{tab:datasets}
		\rowcolors{4}{gray!15}{white}
		\renewcommand{\arraystretch}{1.1}
		\begin{tabular}{lrrrrccllc}\toprule
			\multirow{2}{*}{\textbf{Dataset}}  &\multicolumn{4}{c}{\textbf{Properties}}&\multicolumn{2}{c}{\textbf{Labels}}&\multicolumn{2}{c}{\textbf{Attributes}}&\multirow{2}{*}{\textbf{Ref.}}\\\cmidrule{2-9}
			&  Graphs& Clas.  & Avg. $|V|$ & Avg. $|E|$ & Vertex & Edge & Vertex & Edge\\\midrule
			{AIDS}          & 2000 & 2  & 15.69  & 16.20  & +  & +  & + (4)   & -- & \cite{Riesen2008}              \\
			{BZR}           & 405  & 2  & 35.75  & 38.36  & +  & -- & + (3)   & -- & \cite{Sutherland2003}          \\
			{COX2}          & 467  & 2  & 41.22  & 43.45  & +  & -- & + (3)   & -- & \cite{Sutherland2003}          \\
			{DHFR}          & 467  & 2  & 42.43  & 44.54  & +  & -- & + (3)   & -- & \cite{Sutherland2003}          \\
			{DD}            & 1178 & 2  & 284.32 & 715.66 & +  & -- & --      & -- & \cite{Dob+2003,She+2011}       \\
			{ENZYMES}       & 600  & 6  & 32.63  & 62.14  & +  & -- & + (18)  & -- & \cite{Bor+2005a,Schomburg2004} \\
			{FRANKENSTEIN}  & 4337 & 2  & 16.90  & 17.88  & -- & -- & + (780) & -- & \cite{Ors+2015}                \\
			{IMDB-BINARY}   & 1000 & 2  & 19.77  & 96.53  & -- & -- & --      & -- & \cite{Yan+2015}                \\
			{IMDB-MULTI}    & 1500 & 3  & 13.00  & 65.94  & -- & -- & --      & -- & \cite{Yan+2015}                \\
			{Mutagenicity}  & 4337 & 2  & 30.32  & 30.77  & +  & +  & --      & -- & \cite{Riesen2008,Kaz+2005}     \\
			{MSRC-9}       & 221  & 8  & 40.58  & 97.94  & +  & -- & --      & -- & \cite{Neu+2016}                \\
			{MSRC-21}      & 563  & 20 & 77.52  & 198.32 & +  & -- & --      & -- & \cite{Neu+2016}                \\
			{MSRC-21C}     & 209  & 20 & 40.28  & 96.60  & +  & -- & --      & -- & \cite{Neu+2016}                \\
			{MUTAG}         & 188  & 2  & 17.93  & 19.79  & +  & +  & --      & -- & \cite{Deb+1991,Kri+2012}       \\
			{NCI1}          & 4110 & 2  & 29.87  & 32.30  & +  & -- & --      & -- & \cite{She+2011}                \\
			{NCI109}        & 4127 & 2  & 29.68  & 32.13  & +  & -- & --      & -- & \cite{She+2011}                \\
			{PTC-FM}       & 349  & 2  & 14.11  & 14.48  & +  & +  & --      & -- & \cite{Helma2001,Kri+2012}      \\
			{PTC-FR}       & 351  & 2  & 14.56  & 15.00  & +  & +  & --      & -- & \cite{Helma2001,Kri+2012}      \\
			{PTC-MM}       & 336  & 2  & 13.97  & 14.32  & +  & +  & --      & -- & \cite{Helma2001,Kri+2012}      \\
			{PTC-MR}       & 344  & 2  & 14.29  & 14.69  & +  & +  & --      & -- & \cite{Helma2001,Kri+2012}      \\
			{PROTEINS}      & 1113 & 2  & 39.06  & 72.82  & +  & -- & + (1)   & -- & \cite{Bor+2005a,Dob+2003}      \\
			{REDDIT-BINARY} & 2000 & 2  & 429.63 & 497.75 & -- & -- & --      & -- & \cite{Yan+2015a}               \\
			{SYNTHETICnew}  & 300  & 2  & 100.00 & 196.25 & -- & -- & + (1)   & -- & \cite{Fer+2013}                \\
			{Synthie}       & 400  & 4  & 95.00  & 173.92 & -- & -- & + (15)  & -- & \cite{Mor+2016}                \\
			{Tox21-AR}     & 9362 & 2  & 18.39  & 18.84  & +  & +  & --      & -- & \cite{Tox21}                   \\
			{Tox21-MMP}    & 7320 & 2  & 17.49  & 17.83  & +  & +  & --      & -- & \cite{Tox21}                   \\
			{Tox21-AHR}    & 8169 & 2  & 18.09  & 18.50  & +  & +  & --      & -- & \cite{Tox21}                   \\
			\bottomrule
		\end{tabular}
	\end{center}
\end{table*}

\subsection{Graph Kernels}
As a baseline we included the \emph{vertex label kernel} (VL) and \emph{edge label
kernel} (EL), which are the dot products on vertex and edge label
histograms, respectively. An edge label is a triplet consisting of the labels of
the edge and the label  of its two endpoints.
We used the Weisfeiler-Lehman subtree (WL) and Weisfeiler-Lehman optimal assignment
kernel (WL-OA), see Section~\ref{sec:naa}. For both the number of refinement
operations was chosen from $\{0,1,\dots,8\}$ by cross-validation.
In addition we implemented a \emph{graphlet kernel} (GL3) and the shortest-path
kernel (SP)~\citep{Borgwardt2005}.
GL3 is based on connected subgraphs with three vertices taking labels into account
similar to the approach used by~\citet{She+2011}. For SP we used the indicator function to compare path lengths and computed the kernel by explicit feature maps
in case of discrete vertex labels, cf.~\citep{She+2011}.
These kernels were implemented in Java based on the same common data structures and
support both vertex labels and---with exception of VL and SP---edge labels.

We compare three kernels based on matching of vertex embeddings, the matching kernel of~\citet{Joh+2015} with inverse Laplacian (MK-IL) and Laplacian (MK-L) embeddings and the Pyramid Match (PM) kernel of~\citep{Nik+2017}. The MK kernels lack hyperparameters and for the PM-kernel, we used the default settings---vertex embedding dimension ($d=6$) and matching levels $(L=3)$---in the implementation by~\citet{Nik+2017code}.
Finally, we include the shortest-path variant of the Deep Graph Kernel (DeepGK)~\citep{Yan+2015} with parameters as suggested in~\citet{Yan+2015code} (SP feature type, MLE kernel type, window size 5, 10 dimensions)\footnote{We did not perform a parameter search for the parameters of the Deep Graph kernel and the accuracy of the kernel may improve with a more tailored choice.}, the DBR kernel of~\citet{bai2014graph} (no parameters, code obtained through correspondence) and the propagation kernel (Prop)~\citep{Neu+2016,Neu+2016code} for which we select the number of diffusion iterations by cross-validation and use the settings recommended by the authors for other hyperparameters.

In a comparison of kernels for graphs with continuous vertex attributes we use the shortest-path kernel~\cite{Borgwardt2005} with a Gaussian RBF base kernel to compare vertex attributes, see also~\cref{spk}, the GraphHopper kernel~\cite{Fer+2013}, the GraphInvariant kernel~\cite{Ors+2015}, the Propagation kernel (P2K)~\cite{Neu+2016}, and the Hash Graph kernel~\cite{Mor+2016}.
We set the parameter $\sigma$ of the Gaussian RBF kernel to $\sqrt{\nicefrac{D}{2}}$ for the GraphHopper and the GraphInvariant kernel, as reported in~\cite{Fer+2013,Ors+2015}, where $D$ denotes the number of components of the vertex attributes. For datasets that do not have vertex labels, we either used the vertex degree instead or uniform labels (selected by (double) cross-validation). Following~\cite{Mor+2016}, we set the number of iteration for the Hash Graph kernel to 20 for all datasets, excluding the Sythnie datasets where we used 100.

\subsection{Results and Discussion}
We present our experimental results and discuss the research questions.

\paragraph{\ref{h:complete} Expressivity.}
For these experiments we only considered kernels that are permutation-invariant
and guarantee that two isomorphic graphs are represented by the same feature
vector. This is not the case for the MK-* and PM kernels because of the vertex
embedding techniques applied.

\begin{figure}
	\includegraphics{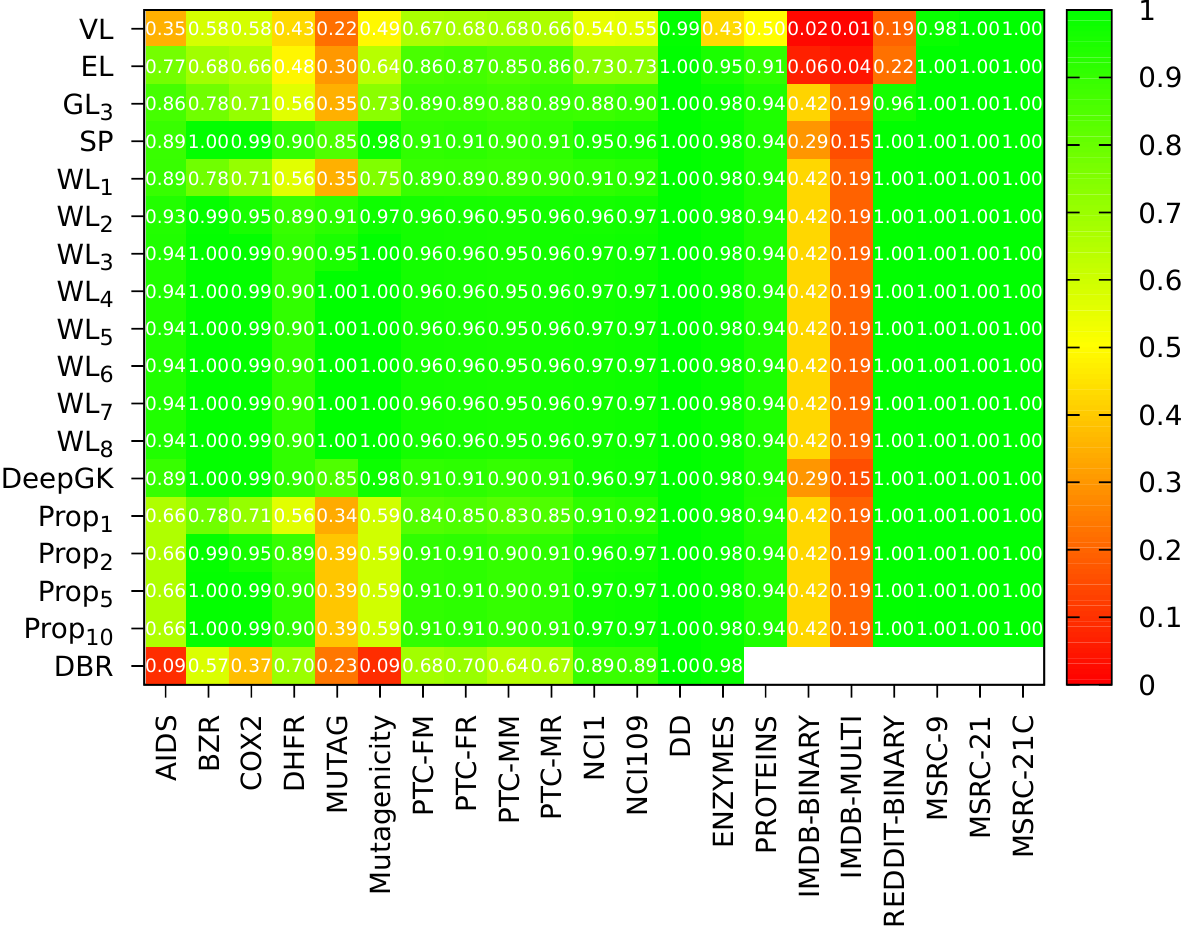}
	\caption{Completeness ratio.}\label{fig:completeness}
\end{figure}

\begin{figure}
	\includegraphics{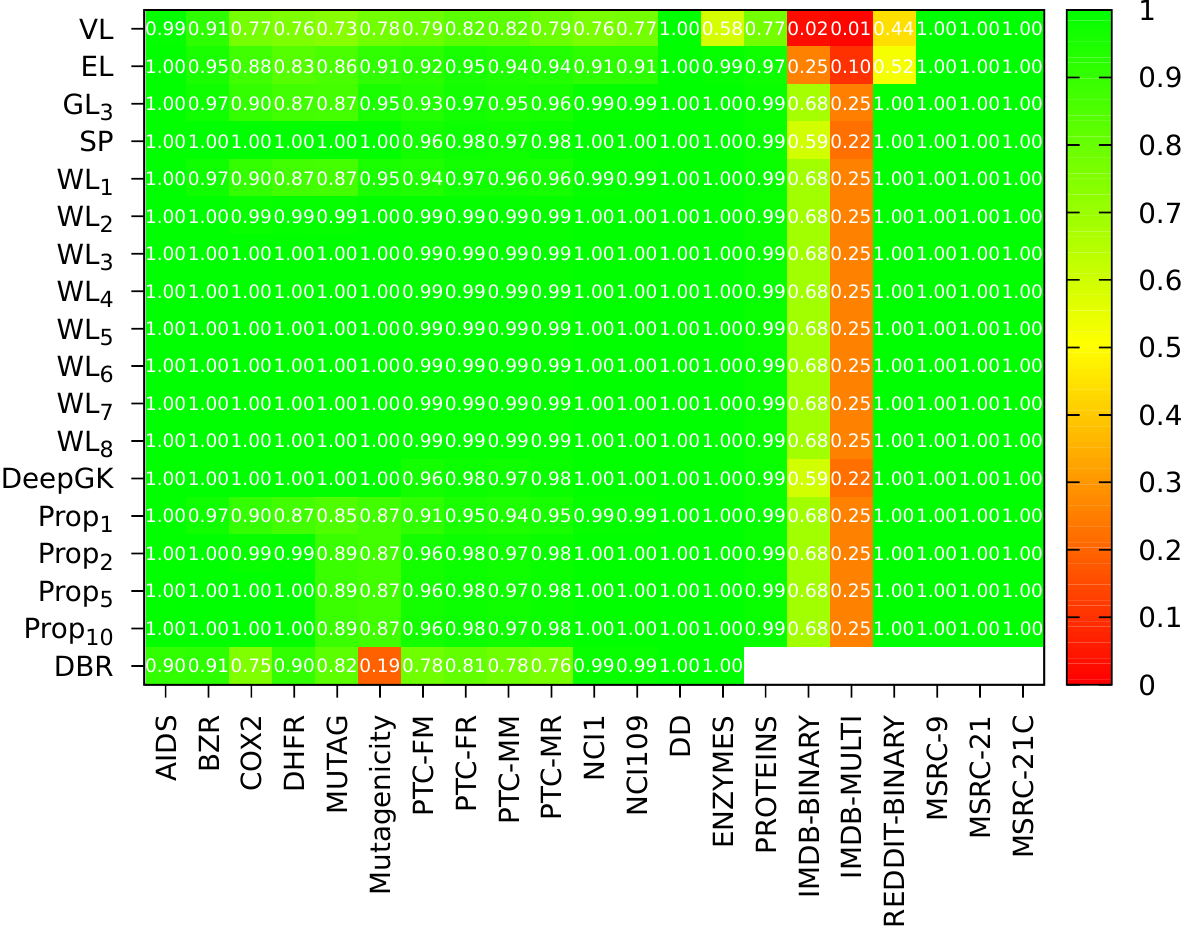}
	\caption{Label completeness ratio.}\label{fig:completeness_label}
\end{figure}

Figure~\ref{fig:completeness} shows the completeness ratio of various permutation
invariant graph kernels with different parameters on the datasets as a heatmap.
The WL-OA kernels achieved the same results as the WL kernels and are therefore
not depicted. As expected, VL achieves only a weak completeness ratio, since it
ignores the graph structure completely. To a lesser extent, this also applies
to EL and GL$_3$.
The SP and the WL$_h$ kernels with $h\geq 2$ provide a high completeness ratio
close to one for most datasets.
However, for the IMDB-BINARY dataset shortest-paths appear to be less powerful
features than small local graphlets. This indicates structural differences
between this dataset and the molecular graph datasets, where SP consistently
achieves better results than GL$_3$. As expected DeepGK performs similar to the
SP kernel. WL and Prop are both based on a neighborhood aggregation mechanism, but
WL achieves a higher completeness ratio on several datasets. This is explained
by the fact that Prop does not support edge labels and does not employ a relabeling
function after each propagation step.
DBR does not take labels into account and consequently fails to distinguish many
graphs of the datasets, for which vertex labels are informative.
The difficulty of distinguishing the graphs in a dataset varies strongly based on
the type of graphs. The computer vision graphs are almost perfectly distinguished
by just considering the vertex label multiplicities, molecular graphs often require
multiple iterations of Weisfeiler-Lehman or global features such as shortest paths.
For social networks, the REDDIT-BINARY graphs are also effectively distinguished
by Weisfeiler-Lehman refinement, while this is not possible for the two IMDB
datasets. However, we observed that all the graphs in these two datasets that
cannot be distinguished by WL$_1$ are in fact isomorphic. Therefore, a higher
completeness ratio cannot be achieved by any permutation-invariant graph kernel.

We now consider the label completeness ratio depicted in Figure~\ref{fig:completeness_label}.
The label completion ratio generally shows the same trends, but higher values close
to one are reached as expected.
For the datasets IMDB-BINARY and IMDB-MULTI we have already observed that WL$_1$
distinguishes all non-isomorphic graphs. As we see in Figure~\ref{fig:completeness_label}
these datasets contain a large number of isomorphic graphs that actually belong
to different classes. Apparently, the information contained in the dataset is
not sufficient to allow perfect classification. A general observations from
the heatmaps is that WL (just as WL-OA) effectively distinguish most graphs after
only few iterations of refinement. For some non-challenging datasets even VL and
EL are sufficient expressive. Therefore, these kernels are interesting baselines
for accuracy experiments.
In order to effectively learn with a graph kernel, it is not sufficient to just
distinguish graphs, which may lead to strong overfitting, but to provide a smooth
similarity measure that allows the classifier to generalize to unseen data.

\paragraph{\ref{h:rbf} Non-linear decision boundaries.}
We discuss the accuracy results of the classification experiments summarized in Tables~\ref{tab:results1} and~\ref{tab:results2}.
The classification accuracy of the simple kernels VL and EL can be drastically
improved by combining them with the Gaussian RBF kernel for several datasets.
A clear improvement is also achieved for GL3 on an average.
For WL and WL-OA the Gaussian RBF kernel only leads to minor changes in classification
accuracy for most datasets. However, a strong improvement is observed for WL
and the dataset ENZYMES, even lifting the accuracy above the value reached by
WL-OA on the same dataset.
However, for the dataset REDDIT-BINARY the accuracy of WL is improved, but still
far below the accuracy obtained by WL-OA, which is based on the histogram
intersection kernel applied to the WL feature vectors.
A surprising result is that the trivial EL kernel combined with the Gaussian RBF kernel
performs competitive to many sophisticated graph kernels. On an average it
provides a higher accuracy than the (unmodified) SP, GL3 and PM kernel.
The DBR kernel does not take labels into account and performs poorly on most datasets.

The application of the Gaussian RBF kernel introduces the hyper-parameter~$\sigma$,
which must be optimized, e.g., via grid search and cross-validation.
This is computational demanding for large datasets, in particular, when the
graph kernel also requires parameters that must be optimized. Therefore,
we suggest to combine VL, EL and GL3 with a Gaussian RBF kernel as a base line.
For WL and WL-OA the parameter $h$ needs to be optimized and the accuracy gain
is minor for most datasets, in particular for WL-OA.
Therefore, their combination with an Gaussian RBF kernel cannot be generally recommended.
Note that the combination with an Gaussian RBF kernel also complicates the application
of fast linear classifiers, which are advisable for large datasets.

\newcommand{\win}[1]{$\hspace{-0.3mm}$\textbf{#1}}
\newcommand{\sd}[1]{\scriptsize{$\pm$#1}}

\begin{sidewaystable*}[p]

	\begin{center}	
		\caption{Classification accuracy and standard deviation for several kernels and their variant when plugged into the Gaussian RBF kernel.}
		\label{tab:results1}
		\rowcolors{4}{gray!15}{white}
		\renewcommand{\arraystretch}{1.3}
		\resizebox{1.1\textwidth}{!}{%
		\begin{tabular}{lcc|cc|cc|cc|cc|cc}\toprule
			\multirow{2}{*}{\textbf{Dataset}}  &\multicolumn{2}{c|}{\textbf{VL}}&\multicolumn{2}{c|}{\textbf{EL}}&\multicolumn{2}{c|}{\textbf{SP}}&\multicolumn{2}{c|}{\textbf{WL}}&\multicolumn{2}{c|}{\textbf{WL-OA}}&\multicolumn{2}{c}{\textbf{GL3}}\\\cmidrule{2-13}
			                & $K_\text{lin}$     & $K_\text{RBF}$     & $K_\text{lin}$     & $K_\text{RBF}$     & $K_\text{lin}$ & $K_\text{RBF}$     & $K_\text{lin}$     & $K_\text{RBF}$     & $K_\text{lin}$     & $K_\text{RBF}$     & $K_\text{lin}$     & $K_\text{RBF}$ \\\midrule
             \textsf{NCI1} &  64.6\sd{0.1} &  67.2\sd{2.8} &  66.3\sd{0.1} &  71.8\sd{0.3} &  73.2\sd{0.3} &  79.3\sd{0.4} &  85.9\sd{0.1} &  86.2\sd{0.1} &  86.2\sd{0.2} &  \win{86.6}\sd{0.2} &  70.5\sd{0.2} &  76.5\sd{0.4}\\
           \textsf{NCI109} &  63.6\sd{0.2} &  68.9\sd{1.4} &  64.9\sd{0.1} &  71.4\sd{0.5} &  72.7\sd{0.3} &  77.6\sd{0.3} &  85.9\sd{0.3} &  86.0\sd{0.3} &  86.2\sd{0.2} &  \win{86.4}\sd{0.2} &  69.3\sd{0.2} &  76.0\sd{0.4}\\
           \textsf{PTC-FR} &  67.9\sd{0.4} &  66.9\sd{0.5} &  66.8\sd{0.5} &  65.2\sd{1.2} &  67.1\sd{2.0} &  63.7\sd{2.0} &  67.1\sd{1.2} &  66.8\sd{1.5} &  67.8\sd{1.1} &  67.0\sd{1.3} &  65.5\sd{0.9} &  65.0\sd{1.4}\\
           \textsf{PTC-MR} &  57.8\sd{0.9} &  59.4\sd{1.4} &  56.7\sd{1.6} &  60.5\sd{1.8} &  58.8\sd{2.2} &  62.0\sd{1.8} &  60.4\sd{1.5} &  \win{62.7}\sd{2.0} &  62.6\sd{1.5} &  \win{62.7}\sd{1.0} &  57.4\sd{1.6} &  60.4\sd{1.6}\\
           \textsf{PTC-FM} &  63.9\sd{0.5} &  62.6\sd{0.9} &  64.5\sd{0.4} &  60.5\sd{1.4} &  62.7\sd{1.0} &  60.2\sd{1.3} &  62.8\sd{1.2} &  60.9\sd{0.8} &  61.6\sd{1.2} &  61.7\sd{1.2} &  60.2\sd{3.0} &  60.7\sd{0.8}\\
           \textsf{PTC-MM} &  66.6\sd{0.8} &  64.7\sd{0.4} &  64.1\sd{1.0} &  62.7\sd{1.6} &  63.3\sd{1.2} &  63.2\sd{0.8} &  \win{67.8}\sd{2.1} &  67.7\sd{1.3} &  66.4\sd{1.1} &  66.3\sd{1.7} &  61.4\sd{1.7} &  61.3\sd{1.4}\\
            \textsf{MUTAG} &  85.4\sd{0.7} &  82.9\sd{1.0} &  83.6\sd{1.0} &  88.4\sd{2.2} &  83.1\sd{1.3} &  85.2\sd{1.4} &  86.6\sd{0.6} &  87.9\sd{1.0} &  87.5\sd{2.1} &  87.3\sd{1.7} &  87.2\sd{1.1} &  87.8\sd{1.1}\\
     \textsf{Mutagenicity} &  67.0\sd{0.2} &  73.9\sd{0.3} &  72.4\sd{0.1} &  80.3\sd{0.3} &  77.4\sd{0.2} &  80.1\sd{0.2} &  83.6\sd{0.2} &  84.5\sd{0.3} &  84.2\sd{0.2} &  \win{84.7}\sd{0.4} &  79.8\sd{0.2} &  82.7\sd{0.3}\\
             \textsf{AIDS} &  \win{99.7}\sd{0.0} &  \win{99.7}\sd{0.0} &  99.5\sd{0.0} &  99.4\sd{0.0} &  99.6\sd{0.0} &  \win{99.7}\sd{0.0} &  \win{99.7}\sd{0.0} &  \win{99.7}\sd{0.0} &  \win{99.7}\sd{0.0} &  \win{99.7}\sd{0.0} &  99.2\sd{0.1} &  99.3\sd{0.1}\\
              \textsf{BZR} &  78.8\sd{0.1} &  86.0\sd{0.2} &  79.1\sd{0.5} &  86.3\sd{0.3} &  86.5\sd{0.9} &  88.1\sd{0.5} &  \win{88.5}\sd{0.7} &  87.9\sd{0.8} &  88.2\sd{0.4} &  88.0\sd{0.5} &  81.6\sd{0.7} &  85.4\sd{1.0}\\
             \textsf{COX2} &  78.2\sd{0.0} &  80.6\sd{0.3} &  82.0\sd{0.6} &  \win{83.9}\sd{0.7} &  80.6\sd{0.9} &  81.7\sd{0.8} &  81.2\sd{1.0} &  81.7\sd{0.7} &  80.4\sd{0.9} &  80.8\sd{1.3} &  81.3\sd{0.7} &  81.9\sd{0.5}\\
             \textsf{DHFR} &  60.9\sd{0.2} &  74.8\sd{1.2} &  67.9\sd{0.6} &  73.2\sd{0.9} &  77.5\sd{0.6} &  80.7\sd{0.7} &  82.7\sd{0.4} &  \win{83.5}\sd{0.6} &  83.0\sd{1.0} &  83.3\sd{0.6} &  74.7\sd{0.6} &  81.2\sd{1.0}\\
               \textsf{DD} &  78.2\sd{0.4} &  80.1\sd{0.4} &  77.5\sd{0.6} &  78.7\sd{0.7} &  79.5\sd{0.6} &  74.5\sd{0.2} &  78.9\sd{0.4} &  80.9\sd{0.3} &  79.2\sd{0.4} &  79.9\sd{0.5} &  79.7\sd{0.7} &  79.1\sd{0.6}\\
         \textsf{PROTEINS} &  71.9\sd{0.4} &  74.7\sd{0.4} &  73.4\sd{0.3} &  75.2\sd{0.5} &  75.9\sd{0.4} &  74.0\sd{0.3} &  75.5\sd{0.3} &  73.9\sd{0.7} &  76.2\sd{0.4} &  75.9\sd{0.6} &  72.7\sd{0.6} &  73.0\sd{0.6}\\
          \textsf{ENZYMES} &  23.4\sd{1.1} &  41.7\sd{1.1} &  27.7\sd{0.7} &  45.1\sd{1.2} &  41.9\sd{1.7} &  59.5\sd{1.3} &  53.7\sd{1.5} &  62.6\sd{1.2} &  59.9\sd{1.0} &  62.3\sd{1.1} &  30.4\sd{1.1} &  58.6\sd{1.0}\\
      \textsf{IMDB-BINARY} &  46.3\sd{0.9} &  56.5\sd{0.6} &  46.0\sd{0.9} &  62.6\sd{1.2} &  57.3\sd{0.6} &  70.2\sd{0.8} &  72.9\sd{0.6} &  71.3\sd{1.0} &  73.1\sd{0.7} &  \win{73.5}\sd{0.6} &  59.4\sd{0.4} &  70.1\sd{0.8}\\
       \textsf{IMDB-MULTI} &  31.9\sd{0.5} &  39.5\sd{0.9} &  30.8\sd{0.9} &  46.9\sd{0.6} &  39.6\sd{0.2} &  46.1\sd{0.7} &  50.3\sd{0.4} &  50.7\sd{0.6} &  50.4\sd{0.5} &  50.7\sd{0.5} &  40.6\sd{0.4} &  47.1\sd{0.5}\\
    \textsf{REDDIT-BINARY} &  75.3\sd{0.1} &  77.6\sd{0.2} &  75.1\sd{0.1} &  79.4\sd{0.1} &  81.7\sd{0.2} &  67.8\sd{0.2} &  80.9\sd{0.4} &  83.9\sd{0.5} &  \win{89.3}\sd{0.2} &  88.9\sd{0.1} &  60.1\sd{0.2} &  73.6\sd{0.1}\\
           \textsf{MSRC-9} &  88.4\sd{1.3} &  87.7\sd{1.0} &  \win{92.6}\sd{0.9} &  90.2\sd{0.7} &  91.4\sd{0.8} &  89.2\sd{1.0} &  90.1\sd{0.8} &  89.1\sd{0.9} &  90.7\sd{0.8} &  90.1\sd{0.7} &  91.6\sd{0.7} &  91.6\sd{0.9}\\
          \textsf{MSRC-21} &  89.4\sd{0.3} &  90.0\sd{0.5} &  89.5\sd{0.3} &  87.3\sd{0.4} &  89.4\sd{0.6} &  37.4\sd{1.2} &  89.3\sd{0.6} &  89.8\sd{0.4} &  90.0\sd{0.6} &  90.5\sd{0.4} &  90.5\sd{0.7} &  85.1\sd{0.6}\\
         \textsf{MSRC-21C} &  81.2\sd{1.2} &  80.8\sd{1.7} &  84.5\sd{0.8} &  81.8\sd{1.3} &  83.8\sd{1.2} &  78.3\sd{1.3} &  81.9\sd{0.9} &  82.1\sd{1.1} &  84.9\sd{0.8} &  84.5\sd{1.0} &  84.0\sd{1.7} &  82.6\sd{1.0}\\
         \textsf{Tox21-AR} &  95.9\sd{0.0} &  96.4\sd{0.0} &  95.9\sd{0.0} &  97.5\sd{0.0} &  97.1\sd{0.0} &  97.5\sd{0.0} &  97.9\sd{0.0} &  \win{98.0}\sd{0.0} &  \win{98.0}\sd{0.0} &  \win{98.0}\sd{0.0} &  96.4\sd{0.0} &  97.6\sd{0.0}\\
        \textsf{Tox21-MMP} &  84.3\sd{0.0} &  86.5\sd{0.1} &  84.5\sd{0.0} &  89.7\sd{0.2} &  86.4\sd{0.1} &  90.7\sd{0.2} &  92.5\sd{0.1} &  \win{93.0}\sd{0.2} &  92.7\sd{0.1} &  92.8\sd{0.1} &  87.3\sd{0.1} &  91.2\sd{0.2}\\
        \textsf{Tox21-AHR} &  88.4\sd{0.0} &  89.1\sd{0.2} &  88.6\sd{0.1} &  91.4\sd{0.2} &  88.4\sd{0.0} &  91.9\sd{0.1} &  93.4\sd{0.1} &  \win{93.7}\sd{0.1} &  93.5\sd{0.1} &  93.6\sd{0.1} &  89.7\sd{0.1} &  92.8\sd{0.2}\\
\midrule
\textbf{Average}  &  71.2\phantom{\sd{0.0}} &  74.5\phantom{\sd{0.0}} &  72.2\phantom{\sd{0.0}} &  76.2\phantom{\sd{0.0}} &  75.6\phantom{\sd{0.0}} &  74.9\phantom{\sd{0.0}} &  79.6\phantom{\sd{0.0}} &  80.2\phantom{\sd{0.0}} &  \win{80.5}\phantom{\sd{0.0}} &  \win{80.6}\phantom{\sd{0.0}} &  73.8\phantom{\sd{0.0}} &  77.5\phantom{\sd{0.0}}\\
			\bottomrule
		\end{tabular}}
	\end{center}
\end{sidewaystable*}

\newcommand{\mkil}{MK-IL}
\newcommand{\mkl}{MK-L}
\newcommand{\pmk}{PM}
\newcommand{\deep}{DeepGK}
\newcommand{\dbr}{DBR}
\newcommand{\propk}{Prop}

\begin{sidewaystable*}[p]
	\begin{center}
		\caption{Classification accuracy and standard deviation for several kernels and their variant when plugged into the Gaussian RBF kernel. }
		\label{tab:results2}
		\rowcolors{4}{gray!15}{white}
		\renewcommand{\arraystretch}{1.3}
			\resizebox{1.1\textwidth}{!}{%
		\begin{tabular}{lcc|cc|cc|cc|cc|cc}\toprule
			\multirow{2}{*}{\textbf{Dataset}}  &\multicolumn{2}{c|}{\textbf{\mkil}}&\multicolumn{2}{c|}{\textbf{\mkl}}&\multicolumn{2}{c|}{\textbf{\pmk}}&\multicolumn{2}{c|}{\textbf{\deep}}&\multicolumn{2}{c|}{\textbf{\dbr}\footnote{Computation of the DBR kernel did not finish within 48h on some datasets, as indicated by a line (---). The DBR kernel does not make use of label information.}}&\multicolumn{2}{c}{\textbf{\propk}}\\\cmidrule{2-13}
			                & $K_\text{lin}$ & $K_\text{RBF}$ & $K_\text{lin}$ & $K_\text{RBF}$ & $K_\text{lin}$ & $K_\text{RBF}$ & $K_\text{lin}$ & $K_\text{RBF}$ & $K_\text{lin}$ & $K_\text{RBF}$ & $K_\text{lin}$ & $K_\text{RBF}$ \\\midrule
             \textsf{NCI1} &  76.8\sd{0.3} &  78.1\sd{0.3} &  72.8\sd{0.3} &  75.5\sd{0.3} &  73.3\sd{0.3} &  80.0\sd{0.3} &  74.9\sd{0.2} &  78.4\sd{0.3} &  67.4\sd{0.3} &  76.0\sd{0.2} &  84.6\sd{0.2} &  85.6\sd{0.2}\\
           \textsf{NCI109} &  75.4\sd{0.3} &  75.9\sd{0.3} &  71.9\sd{0.3} &  74.1\sd{0.4} &  71.1\sd{0.3} &  78.8\sd{0.2} &  73.3\sd{0.2} &  77.7\sd{0.3} &  66.6\sd{0.2} &  75.3\sd{0.2} &  84.1\sd{0.2} &  84.7\sd{0.2}\\
           \textsf{PTC-FR} &  \win{69.1}\sd{0.6} &  68.6\sd{0.7} &  68.4\sd{0.7} &  68.1\sd{0.9} &  65.9\sd{0.8} &  65.0\sd{1.1} &  66.4\sd{1.2} &  63.8\sd{1.5} &  65.3\sd{0.5} &  64.0\sd{0.7} &  66.0\sd{1.5} &  65.6\sd{1.0}\\
           \textsf{PTC-MR} &  60.6\sd{1.0} &  60.5\sd{1.1} &  59.1\sd{1.7} &  59.7\sd{1.4} &  61.5\sd{1.3} &  59.5\sd{2.0} &  59.9\sd{1.5} &  60.9\sd{1.7} &  53.7\sd{1.3} &  55.1\sd{1.9} &  59.9\sd{1.6} &  61.3\sd{2.1}\\
           \textsf{PTC-FM} &  58.6\sd{1.8} &  \win{64.7}\sd{0.6} &  60.4\sd{0.9} &  61.4\sd{1.6} &  59.7\sd{1.4} &  62.2\sd{0.8} &  62.6\sd{0.9} &  60.9\sd{1.1} &  56.2\sd{1.5} &  59.8\sd{1.5} &  60.9\sd{1.6} &  61.7\sd{1.6}\\
           \textsf{PTC-MM} &  62.1\sd{1.6} &  65.0\sd{1.4} &  63.8\sd{1.2} &  63.1\sd{0.7} &  62.9\sd{1.3} &  62.2\sd{1.0} &  63.3\sd{0.9} &  61.8\sd{1.4} &  59.4\sd{1.1} &  63.5\sd{1.0} &  63.9\sd{1.0} &  64.6\sd{1.9}\\
            \textsf{MUTAG} &  82.8\sd{1.4} &  83.7\sd{0.8} &  83.1\sd{1.3} &  83.5\sd{1.0} &  84.9\sd{1.2} &  86.7\sd{0.8} &  85.1\sd{1.5} &  84.4\sd{0.7} &  86.2\sd{1.6} &  84.6\sd{0.7} &  \win{90.3}\sd{0.9} &  86.1\sd{1.1}\\
     \textsf{Mutagenicity} &  70.7\sd{0.3} &  75.2\sd{0.2} &  70.9\sd{0.3} &  75.3\sd{0.2} &  72.1\sd{0.2} &  75.5\sd{0.3} &  79.4\sd{0.3} &  80.2\sd{0.2} &  66.2\sd{0.1} &  66.8\sd{0.5} &  67.5\sd{0.2} &  76.7\sd{0.4}\\
             \textsf{AIDS} &  99.6\sd{0.0} &  99.6\sd{0.1} &  99.6\sd{0.0} &  99.6\sd{0.0} &  \win{99.7}\sd{0.0} &  \win{99.7}\sd{0.0} &  99.6\sd{0.0} &  99.6\sd{0.0} &  99.3\sd{0.1} &  \win{99.7}\sd{0.0} &  \win{99.7}\sd{0.0} &  \win{99.7}\sd{0.0}\\
              \textsf{BZR} &  88.1\sd{0.8} &  88.2\sd{0.8} &  88.1\sd{0.6} &  87.8\sd{0.7} &  84.5\sd{1.0} &  85.5\sd{0.7} &  86.5\sd{0.6} &  87.8\sd{0.6} &  82.8\sd{0.9} &  84.1\sd{0.8} &  87.1\sd{0.5} &  87.7\sd{1.0}\\
             \textsf{COX2} &  81.2\sd{1.0} &  81.1\sd{0.5} &  80.5\sd{0.8} &  80.5\sd{0.7} &  80.7\sd{0.5} &  80.3\sd{0.7} &  80.4\sd{1.1} &  81.4\sd{0.6} &  78.1\sd{0.1} &  77.3\sd{0.7} &  81.7\sd{0.8} &  81.5\sd{0.9}\\
             \textsf{DHFR} &  81.5\sd{0.8} &  82.1\sd{0.3} &  79.2\sd{0.8} &  80.0\sd{0.7} &  75.3\sd{0.7} &  78.1\sd{0.8} &  80.7\sd{1.0} &  81.0\sd{0.8} &  75.1\sd{0.5} &  78.3\sd{0.7} &  82.8\sd{0.6} &  83.2\sd{0.7}\\
               \textsf{DD} &  78.3\sd{0.3} &  78.2\sd{0.3} &  77.3\sd{0.4} &  77.3\sd{0.4} &  78.7\sd{0.3} &  79.2\sd{0.9} &  79.4\sd{0.4} &  71.0\sd{0.2} &  78.8\sd{0.6} &  78.2\sd{0.6} &  78.9\sd{0.3} &  \win{81.6}\sd{0.5}\\
         \textsf{PROTEINS} &  76.6\sd{0.6} &  \win{76.8}\sd{0.4} &  75.1\sd{0.2} &  74.8\sd{0.5} &  74.5\sd{0.4} &  74.6\sd{0.5} &  75.7\sd{0.3} &  74.2\sd{0.4} &           --- &           --- &  74.3\sd{0.5} &  74.6\sd{0.5}\\
          \textsf{ENZYMES} &  \win{64.1}\sd{1.3} &  63.5\sd{1.1} &  61.6\sd{1.2} &  62.0\sd{1.2} &  40.2\sd{1.0} &  49.3\sd{1.1} &  42.3\sd{1.0} &  58.9\sd{1.1} &  37.6\sd{0.7} &  39.5\sd{1.3} &  49.0\sd{1.6} &  62.6\sd{0.9}\\
      \textsf{IMDB-BINARY} &  69.4\sd{0.6} &  69.9\sd{0.5} &  70.6\sd{0.5} &  70.1\sd{0.4} &  70.7\sd{0.6} &  71.1\sd{0.9} &  60.5\sd{0.3} &  70.2\sd{0.7} &           --- &           --- &  \win{73.5}\sd{0.3} &  71.2\sd{0.7}\\
       \textsf{IMDB-MULTI} &  46.1\sd{0.7} &  47.0\sd{0.5} &  47.1\sd{0.6} &  47.6\sd{0.4} &  47.8\sd{0.6} &  47.8\sd{0.6} &  40.8\sd{1.1} &  46.1\sd{0.7} &           --- &           --- &  49.8\sd{0.6} &  \win{51.0}\sd{0.7}\\
    \textsf{REDDIT-BINARY} &           --- &           --- &           --- &           --- &  82.3\sd{0.2} &  82.7\sd{0.4} &  82.4\sd{0.1} &  67.8\sd{0.2} &           --- &           --- &  78.2\sd{0.3} &  85.5\sd{0.3}\\
           \textsf{MSRC-9} &  90.9\sd{1.0} &  90.4\sd{0.5} &  90.4\sd{1.2} &  90.4\sd{0.8} &  90.4\sd{1.4} &  90.1\sd{1.0} &  91.8\sd{0.8} &  88.2\sd{1.2} &           --- &           --- &  89.4\sd{1.3} &  89.7\sd{0.9}\\
          \textsf{MSRC-21} &  89.0\sd{0.6} &  89.0\sd{0.8} &  89.3\sd{0.5} &  89.3\sd{0.5} &  \win{91.3}\sd{0.5} &  91.2\sd{0.5} &  89.9\sd{0.5} &  27.5\sd{1.2} &           --- &           --- &  88.6\sd{0.5} &  89.8\sd{0.6}\\
         \textsf{MSRC-21C} &  \win{85.7}\sd{0.6} &  85.6\sd{0.9} &  85.6\sd{0.9} &  84.8\sd{1.1} &  84.4\sd{0.9} &  84.6\sd{0.9} &  85.1\sd{1.4} &  76.8\sd{1.3} &           --- &           --- &  81.4\sd{1.1} &  81.8\sd{1.1}\\
         \textsf{Tox21-AR} &  97.7\sd{0.0} &  97.7\sd{0.0} &  97.4\sd{0.0} &  97.4\sd{0.0} &  97.6\sd{0.0} &  97.7\sd{0.0} &  97.0\sd{0.0} &  97.6\sd{0.0} &           --- &           --- &  97.8\sd{0.0} &  97.8\sd{0.0}\\
        \textsf{Tox21-MMP} &  86.9\sd{0.1} &  87.2\sd{0.1} &  86.8\sd{0.1} &  87.2\sd{0.1} &  86.6\sd{0.2} &  89.7\sd{0.1} &  86.3\sd{0.1} &  90.4\sd{0.1} &           --- &           --- &  84.7\sd{0.1} &  89.6\sd{0.2}\\
        \textsf{Tox21-AHR} &  89.6\sd{0.0} &  89.6\sd{0.0} &  89.4\sd{0.1} &  89.5\sd{0.0} &  89.4\sd{0.1} &  91.7\sd{0.1} &  88.7\sd{0.1} &  91.8\sd{0.1} &           --- &           --- &  89.2\sd{0.0} &  91.1\sd{0.1}\\
\midrule
\textbf{Average}  &  77.4\phantom{\sd{0.0}} &  78.2\phantom{\sd{0.0}} &  76.9\phantom{\sd{0.0}} &  77.3\phantom{\sd{0.0}} &  76.1\phantom{\sd{0.0}} &  77.6\phantom{\sd{0.0}} &  76.3\phantom{\sd{0.0}} &  74.1\phantom{\sd{0.0}} &  69.5\phantom{\sd{0.0}} &  71.6\phantom{\sd{0.0}} &  77.6\phantom{\sd{0.0}} &  78.1\phantom{\sd{0.0}}\\
			\bottomrule
		\end{tabular}}
	\end{center}
\end{sidewaystable*}

\paragraph{\ref{h:sup} Accuracy.}
Tables~\ref{tab:results1} and~\ref{tab:results2} show that for almost every kernel
there is at least one dataset, for which it provides the best accuracy.
This is even true for the trivial kernels VL and EL on the datasets
AIDS and MSRC-9; and also COX2 when combined with an Gaussian RBF kernel.
Moreover, VL combined with the Gaussian RBF kernel almost reaches the accuracy of the best kernels for DD.
The dataset AIDS is almost perfectly classified by VL, which suggests that
this dataset is not an adequate benchmark dataset for graph kernel comparison.
For the other two datasets (MSRC-9 and COX2), there are two possible reasons for the observed results. Either these datasets can be classified optimally without taking the graph structure into account, making them not adequate for graph kernel comparison. This would mean that the remaining error is dominated by irreducible error (label noise). Alternatively, current state-of-the-art kernels are not able to benefit from their structure; the remaining error is due to bias. If the second reason is true, these datasets are particularly challenging. In practice, for a finite dataset, it is hard to distinguish bias from noise conclusively, and it is likely that the full explanation is a combination of the two.

The kernels WL and WL-OA provide the best accuracy results for most datasets.
WL-OA achieves the highest accuracy on an average even without combining it with
the Gaussian RBF kernel. Since these kernels are also efficiently computed, they
represent a suitable first approach when classifying new datasets.
We suggest to use WL-OA for small and medium-sized datasets with kernel support
vector machines and WL for large datasets with linear support vector machines.

The analysis of the label completeness ration depicted in Figure~\ref{fig:completeness_label}
suggests that VL cannot perform well on ENZYMES, IMDB-BINARY, IMDB-MULTI and
REDDIT-BINARY.  EL shows weaknesses on IMDB-BINARY, IMDB-MULTI and
REDDIT-BINARY and DBR on Mutagenicity.
The WL and WL-OA kernels can effectively distinguish most non-isomorphic
benchmark graphs.
These observations are in accordance with the accuracy results observed.
However, there is no clear relation between the label completeness ratio and the
prediction accuracy. This suggests that the ability of graph kernels to take
features into account that allow to effectively distinguish graphs is only a
minor issue for current benchmark datasets. Instead taking the features into
account that allow the classifier to generalize to unseen data appears to be
most relevant.

\paragraph{\ref{h:qsim} Agreement.}
The sheer number and variety of existing graph kernels suggest that there may be groups of kernels that are more similar to each other than to other kernels. In this section, we attempt to discover such groups by a qualitative comparison of the predictions (and errors) made by different kernels for a fixed set of graphs. Additionally, we examine the heterogeneity in errors made for the same set of graphs to assess the overall agreement between rivalling kernels.

\begin{figure}
	\centering
	\includegraphics[width=.97\textwidth]{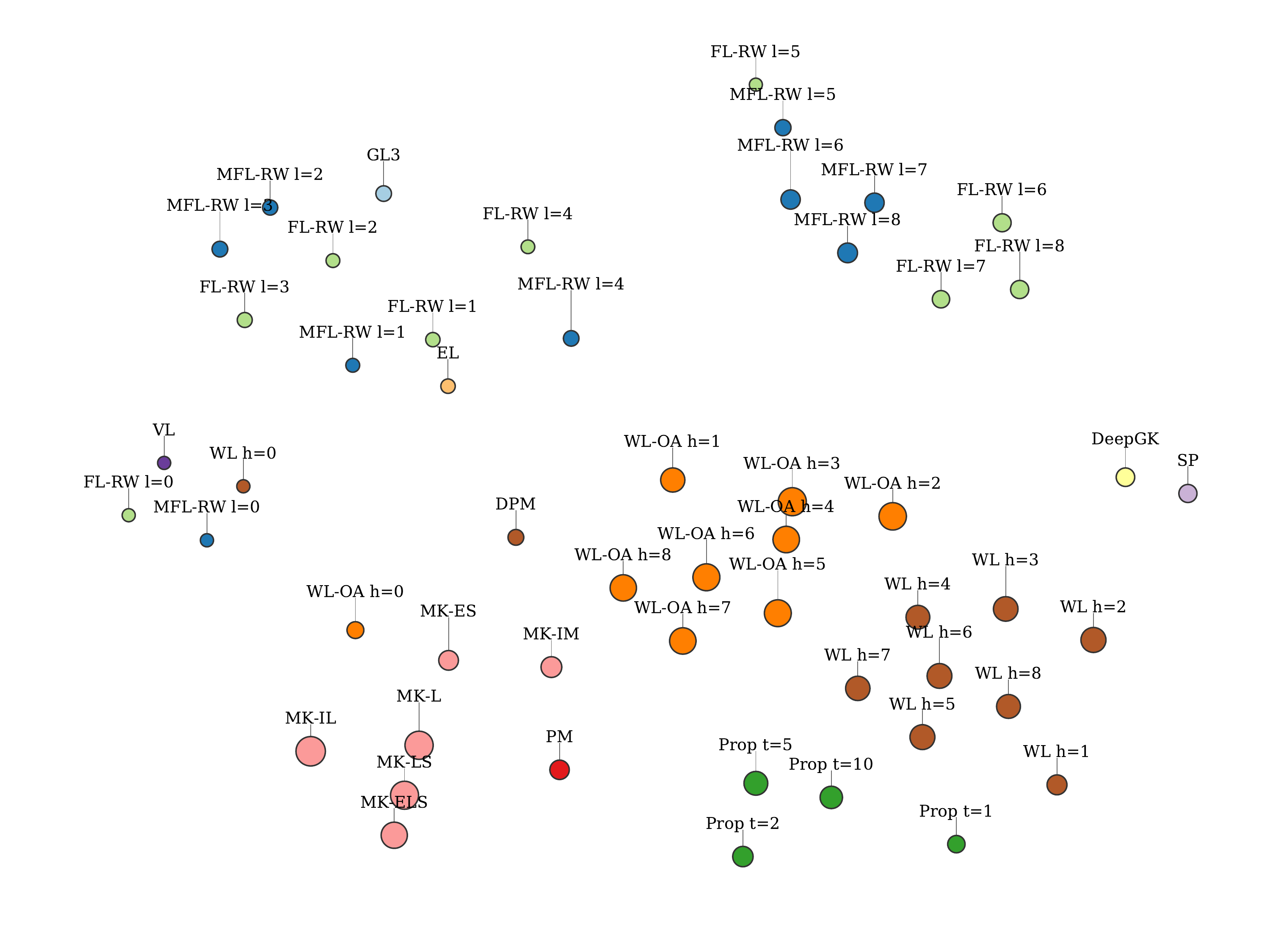}
	\caption{Graph kernels embedded in 2D by tSNE projection of their predictions on MUTAG, ENZYMES and PTC-MR. The results illustrate the similarities among, for example, short-length RW kernels (FL-RW $l\leq4$) and small-graphlet GK kernels (GL3), as well as WL and Prop kernels.}\label{fig:all_tsne}
\end{figure}

We embed each kernel into a common geometric space based on their predictions on a set of benchmark graphs. Let each kernel $k_1, ..., k_m$ and each graph $G_1, ..., G_n$ in a dataset $D$ index the rows and columns of a matrix $\vec{P}^D \in \mathbb{R}^{m\times n}$, respectively. Then, let $P^D_{ij}$ represent the prediction made by kernel $k_i$ on graph $G_j$ after being trained on other graphs from $D$. We construct such matrices $\vec{P}^l$ for multiple datasets $\{D_l\}_{l=1}^N$ and concatenate them to form $\vec{P} = [\vec{P}^1, ..., \vec{P}^N]$, a high-dimensional representation of the features captured by each kernel. Similarly, we construct matrices $\{\vec{E}^l\}_{l=1}^N$ and $\vec{E} = [\vec{P}^1, ..., \vec{P}^N]$, representing the prediction errors made by different kernels on different graphs. Specifically, we let $E^l_{ij} = \mathds{1}[P_{ij} \neq y_l(G_j)]$, where $y_l(G_j)$ is the class label of $G_j$. Here, we construct $\vec{P}$ and $\vec{E}$ from the predictions made by a large set of kernels and parameter settings (see \cref{fig:all_tsne} for a list) applied to the datasets MUTAG, ENZYMES and PTC-MR.

In \cref{fig:all_tsne}, we illustrate the predictions of different kernels by projecting the rows of the prediction matrix $\vec{P}$ to $\mathbb{R}^2$ using t-SNE~\citep{maaten2008visualizing}. The position of each dot represents a projection of the predictions made by a single kernel. The color represents the kernel family and the size represents the average accuracy of the kernel in the considered datasets. For comparison, we include two additional variants of the RW kernel: one comparing only walks of a fixed length $l$ (FL-RW), and one defined as the sum of such kernels up to a fixed length $l$ (MFL-RW).
We see that WL optimal assignment (WL-OA) and matching kernels (MK) predict similarly, compared to for example short-length RW kernels. However, despite small random walks and WL-OA with $h=0$ representing very local features, they predict qualitatively different. We also see that RW kernels that sum up kernels of length $l<L$ walks are very similar to kernels based on just length $L$ walks and that EL, GL3 and short-length RW kernels predict similarly, as expected from their local scope.

Similarity between two rows $e_i = E_{i\cdot}, e_{i'} = E_{i'\cdot}$ of the error matrix $\vec{E}$ indicate that kernels $k_{i}$ and $k_{i'}$ make similar predictive errors on the considered datasets. To assess the overall extent to which particular graphs are ``easy'' or ``hard'' for many kernels, we studied the variance of the columns of $E$. We find that the average zero-one loss across kernels on MUTAG (0.14), ENZYMES (0.57) and PTC-MR (0.42) correlates strongly with the mean absolute deviation around the median across kernels (0.07, 0.26, 0.23). The latter may be interpreted as the fraction of instances for which kernels disagree with the majority vote. We also evaluated the average \emph{inter-agreement} between kernels as measured using \new{Fleiss' kappa}~\citep{fleiss1971measuring}. A high value of Fleiss' kappa indicates that different raters agree significantly more often than random raters with the same marginal label probabiltiy. On MUTAG, ENZYMES and PTC-MR, the kappa measure shows a trend similar (but inverse) to the standard deviation with values of (0.60, 0.28, 0.36).

We conclude that, on these examples, the more difficult the classification task, the more varied the predictive errors. Indeed, if the average error across kernels was 0.0, all models would agree everywhere. However, if different kernels had similar biases, the reverse would not necessarily be true. Instead, these results confirm our intuition that different kernels encode different biases and may be appropriate for different datasets as a result.

\paragraph{\ref{h:cont} Continuous attributes.} As can be seen in~\cref{er}, on all datasets, excluding the FRANKENSTEIN dataset, one variant of the hash graph kernel framework achieves state-of-the-art results. This is in line with the theoretical results outlined in~\cite{Mor+2016}, i.e., they show how to approximate well-known graph kernels for graphs with vertex attributes up to some arbitrarily small error (depending on the number of iterations). However, the results are already achieved with a small number of iterations. This is likely a property of the employed datasets, i.e., a coarse-grained comparison of the attributes is sufficient.
Moreover, together with the propagation kernel, the instances of the hash graph kernel framework achieve a much lower running time compared to the other implicit approaches. The lower performance of the hash graph kernel instances on the FRANKENSTEIN dataset is likely due to the high-dimensional vertex attributes, which are hard to compare using hash functions.

\begin{table}
	\setlength{\tabcolsep}{5pt}
	\begin{center}

		\caption{Classification accuracies in percent and standard deviations (Number of iterations for \textsc{HGK-WL} and \textsc{HGK-SP}: 20 (100 for \textsc{Synthie}), \textsc{OOM}--- Out of Memory.}
		\rowcolors{4}{gray!15}{white}
		\renewcommand{\arraystretch}{1.3}
			\resizebox{1.\textwidth}{!}{%
			\begin{tabular}{lc|c|c|c|cc}\toprule
				\multirow{2}{*}{\textbf{ Kernel}}&\multicolumn{5}{c}{\textbf{Dataset}} & \multirow{2}{*}{\textbf{ Average}} \\\cmidrule{2-6}
				                & \textsf{ENZYMES}   & \textsf{FRANKENSTEIN} & \textsf{PROTEINS}  & \textsf{SyntheticNew} & \textsf{Synthie}
				\\\midrule
				\textsf{SP+RBF} & 71.0\sd{0.8}            & 72.8\sd{0.2}               & 76.6\sd{0.5}            & 96.2\sd{0.4}               &  52.8\sd{1.8}    & 73.9 \\
				\textsf{HGK-SP} & \win{71.3}\sd{0.9} & 70.1\sd{0.3}          & \win{77.5}\sd{0.4} & 96.5\sd{0.6}          & 94.3\sd{0.5}                         & 81.9 \\
				\textsf{HGK-WL} & 67.6\sd{1.0}       & 73.6\sd{0.4}          & 76.7\sd{0.4}       & \win{98.8}\sd{0.3}    & \win{96.8}\sd{0.5}                   & \win{82.7} \\
				\textsf{GH}     & 68.8\sd{1.0}       & 68.5\sd{0.3}          & 72.3\sd{0.3}       & 85.1\sd{1.0}          & 73.2\sd{0.8}                         & 73.6 \\
				\textsf{GI}     & \win{71.7}\sd{0.8} & \win{76.3}\sd{0.3}    & 76.9\sd{0.5}       & 83.1\sd{1.1}          & 95.8\sd{0.5}                         & 80.8 \\
				\textsf{P2K}    & 69.2\sd{0.4}       & \textsf{OOM}          & 73.5\sd{0.5}       & 91.7\sd{0.9}          & 50.2\sd{1.9}                         & 71.2 \\
				\bottomrule

			\end{tabular}}
		\label{er}
	\end{center}
\end{table}

\subsection{A practitioner's guide}
\label{sec:guide}
Because of the limited theoretical knowledge we have about the expressivity of different kernels and the challenge of assessing this a priori, it is difficult to predict which kernel will perform well for a given problem. Nevertheless, it is often the case that some of the kernels in the literature are less or more well suited to the problem at hand. For example, kernels with high time complexity w.r.t.\@ vertex count are expensive to compute for very large graphs; kernels that do not support vertex attributes are ill-suited in learning problems where these are highly significant.

\begin{sidewaysfigure}
	\centering
	\includegraphics[width=1\textwidth]{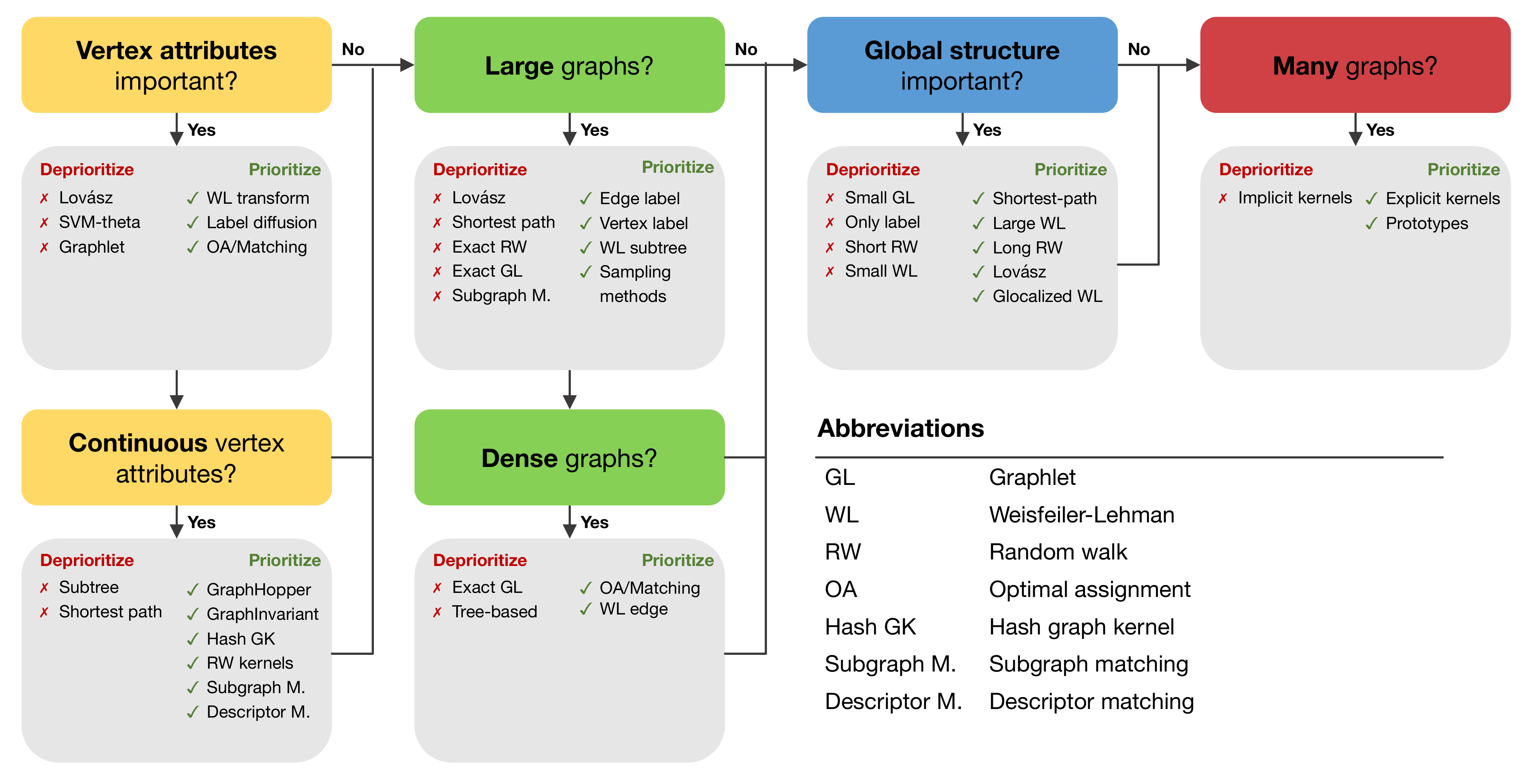}
	\caption{Guidelines for prioritizing kernels for consideration based on known properties of the graph learning problem. In Section~\ref{sec:guide}, we justify these recommendations based on the graph kernel literature.}
	\label{fig:flowchart}
\end{sidewaysfigure}

Below, we give and motivate general guidelines for prioritizing and deprioritizing kernels based on four properties of the problem at hand: the importance and nature of vertex attributes, the size and density of graphs, the importance of global structure, and the number of graphs in the available dataset. Examples of appropriate and unappropriate kernels are given for  extreme cases of each property, and the resulting guidelines are illustrated in Figure~\ref{fig:flowchart}.
The chosen set of properties is certainly a subset of those that may be predictive of a kernel's performance in a given task. For example, the density and number of vertices of a graph are very crude measures of the graph's structure. On the other hand, these features are generally applicable and easy to compute for any sets of graphs. In some fixed domain, more specific structural properties such as girth or diameter may be important and could guide the choice of kernel further. In this work, however, we limit ourselves to the more general case.

\paragraph{Vertex attributes} Almost all established benchmarks for graph classification contain vertex labels and almost all graph kernels support the use of them in some way. In fact, any kernel can be made sensitive to vertex and edge attribute through multiplication by a label kernel, although this approach will not take into account the dependencies between labels and structure. Hence, one of the great contributions of the Weisfeiler-Lehman~\citep{She+2011} and related kernels (e.g. Propagation kernels~\citep{Neu+2016}) is that they capture such dependencies in transformed graphs that are beneficial to all kernels that support vertex labels. It has therefore become standard practice to perform a WL-like transform on labeled graphs before application of other kernels. For this reason, we consider WL-kernels a first choice for applications where vertex labels are important. Propagation kernels also naturally couple structure and attributes, but are generally more expensive to compute. The assignment step of OA kernels matches  vertices based on both structure and attribute, depending on implementation.
In contrast, the original Lov\'{a}sz, SVM-theta and graphlet kernels have no standard way of incorporating vertex labels. The graphlet kernel may be modified to do so by considering subgraph patterns as different if they have different labels. An important special-case of attributed graphs is graphs with non-discrete vertex attributes; these require special consideration. The GraphHopper, GraphInvariant and Hash Graph kernels as well as neural network-based approaches excel at making use of such attributes. In contrast, subtree kernels and shortest-path kernels become prohibitively expensive to compute when combined with continuous attributes.

\paragraph{Large graphs} Early graph kernels such as the RW and SP kernels were plagued by worst-case running time complexities that were prohibitively high for large graphs: $\mathcal{O}(n^6)$ and $\mathcal{O}(n^4)$ for pairs of graphs with $n$ the largest number of vertices. Also expensive to compute, the subgraph matching kernel has complexity $\mathcal{O}(kn^{2(k+1)})$ where $k$ is the size of the considered subgraphs. In practice, even a complexity quadratic in the number of vertices is too high for large-scale learning---the goal is often to achieve complexity linear in the largest number of edges, $m$. This goal puts fundamental limitations on expressivity, as linear complexity is unachievable if the attributes of each edge of one graph has to be independently compared to those of each edge in another. However, when speed is of utmost importance, we recommend using efficient alternatives such as fast subtree kernels with complexity $\mathcal{O}(hm)$ where $h$ the depth of the deepest subtree. Additionally, a single WL iteration may be computed in $\mathcal{O}(m)$ time and the WL label propagation may be used as-is with an already fast kernel at a constant multiplicative cost $h$, equal to the number of WL iterations. As a result, to improve a kernel's sensitivity to vertex label structure is often relatively cheap. Finally, for settings when a particular kernel is preferred for its expressivity but not for its running time, authors have proposed approximation schemes that reduce running time based on sampling or approximate optimization. For example, the time to compute the $k$-graphlet spectrum for a graph, with worst-case complexity $\mathcal{O}(nd^{k-1})$ and $d$ the maximum degree, may be significantly reduced for dense graphs by sampling subgraphs to produce an unbiased estimate of the kernel; The Lovász kernel, with complexity $\mathcal{O}(n^6)$, was approximated with the SVM-theta kernel with  $\mathcal{O}(n^2)$; The random walk kernel may be approximated by the $p$-random walk kernel where walks are limited to length $p$. Similar approximations may be derived also for other kernels. For very large graphs, simple alternatives like the edge label and vertex label kernels may be useful baselines but neglect the graph structure completely.

\paragraph{Global structure} Global properties of graphs are properties that are not well described by statistics of (small) subgraphs~\citep{Joh+2014}. It has been shown, for example, that there exist graphs for which all small subgraphs are trees, but the overall graph has high girth and high chromatic number~\citep{alon2004probabilistic}.  Although the graph kernel literature has often left the precise interpretation of ``global'' to the reader, kernels such as the Lov\'{a}sz kernels and the Glocalized WL kernel, have been proposed with guarantees of capturing specific properties that are considered global by the authors (see \cref{sec:other}). Beside these kernels, if domain knowledge suggests that global structure is important to the task at hand, we recommend prioritizing kernels that compute features from larger subgraph patterns, walks or paths. This rules out the use of Graphlet kernels, since counting large graphlets is often prohibitively expensive, and (small) neighborhood aggregation methods such as the Weisfeiler-Lehman kernel for small numbers of iterations. On the other hand, the shortest-path kernel, long-walk RW and high-iteration WL kernels compute features based on patterns spanning large portions of graphs.

\paragraph{Large datasets} A drawback of kernel methods in general is that they require computation and storage of the full $N\times N$ kernel matrix for each pair of instances in a dataset of $N$ graphs. This can be alleviated significantly if the chosen kernel admits an explicit $d$-dimensional representation with $d \ll N$, such as the vertex label, Weisfeiler-Lehman and graphlet kernels. In this case, only the $N \times d$ feature matrix is necessary for learning. Thus, if many graphs are available to learn from, we recommend starting with kernels that admit an explicit feature representation, such as the WL, GL and subtree kernels. However, this is not always possible, such as when continuous vertex attributes are important, and vertices are compared with a distance metric. Instead, computations using implicit kernels may be approximated using the prototypes method described in \cref{def:oakernel} in which a subset of $d$ graphs are selected and compared to each instance in the dataset. Under certain conditions on the prototype selection, this gives an unbiased estimator of the kernel matrix which can be used in place of its exact version.
Finally, in most cases, more efficient learning methods are applicable when explicit feature representations are available. For classification with support vector machines, for example, the software package LIBSVM~\citep{Cha+2011} is commonly used for learning with (implicit) kernels. When explicit feature representations are available, the software LIBLINEAR~\citep{Fan2008}, which scales to very large datasets, can be used as an alternative.

\section{Conclusion}
We gave an overview over the graph kernel literature. We hope that this survey will spark further progress in the area of graph kernel design and graph classification in general. Moreover, we hope that this article is valuable for the practitioner applying graph classification methods to solve real-world problems.

\section*{Acknowledgements}
We thank Pinar Yanardag, Lu Bai, Giannis Nikolentzos, Marion Neumann, and Franceso Orsini for providing their graph kernel source code.

\section*{Funding}
This work has been supported by the German Research Foundation (DFG) within
the Collaborative Research Center SFB 876 ``Providing Information by
Resource-Constrained Data Analysis'', project A6 ``Resource-efficient Graph Mining''.

\bibliographystyle{unsrtnat}
\bibliography{bibliography}

\begin{thebibliography}{133}
\providecommand{\natexlab}[1]{#1}
\providecommand{\url}[1]{\texttt{#1}}
\expandafter\ifx\csname urlstyle\endcsname\relax
  \providecommand{\doi}[1]{doi: #1}\else
  \providecommand{\doi}{doi: \begingroup \urlstyle{rm}\Url}\fi

\bibitem[Ghosh et~al.(2018)Ghosh, Das, Gonçalves, Quaresma, and
  Kundu]{Gho+2018}
S.~Ghosh, N.~Das, T.~Gonçalves, P.~Quaresma, and M.~Kundu.
\newblock The journey of graph kernels through two decades.
\newblock \emph{Computer Science Review}, 27:\penalty0 88 -- 111, 2018.

\bibitem[Zhang et~al.(2018{\natexlab{a}})Zhang, Wang, and Wang]{Zha+2018b}
Y.~Zhang, L.~Wang, and L.~Wang.
\newblock A comprehensive evaluation of graph kernels for unattributed graphs.
\newblock \emph{Entropy}, 20\penalty0 (12):\penalty0 984, 2018{\natexlab{a}}.

\bibitem[Vishwanathan et~al.(2010)Vishwanathan, Schraudolph, Kondor, and
  Borgwardt]{Vis+2010}
S.~V.~N. Vishwanathan, N.~N. Schraudolph, R.~Kondor, and K.~M. Borgwardt.
\newblock Graph kernels.
\newblock \emph{Journal of Machine Learning Research}, 11:\penalty0 1201--1242,
  2010.

\bibitem[Borgwardt(2007)]{Bor+2007}
K.~M. Borgwardt.
\newblock \emph{Graph kernels}.
\newblock Phd thesis, Ludwig Maximilians University Munich, 2007.

\bibitem[Kriege(2015)]{Kri2015}
N.~M. Kriege.
\newblock \emph{Comparing Graphs: {A}lgorithms \& Applications}.
\newblock Phd~thesis, TU Dortmund University, 2015.

\bibitem[Neumann(2015)]{Neu+2015}
M.~Neumann.
\newblock \emph{Learning with Graphs using Kernels from Propagated
  Information}.
\newblock Phd~thesis, University of Bonn, 2015.

\bibitem[Shervashidze(2012)]{She+2012}
N.~Shervashidze.
\newblock \emph{Scalable graph kernels}.
\newblock Phd~thesis, 2012.

\bibitem[Sch\"{o}lkopf and Smola(2001)]{Scholkopf2001}
Bernhard Sch\"{o}lkopf and Alexander~J. Smola.
\newblock \emph{Learning with Kernels: Support Vector Machines, Regularization,
  Optimization, and Beyond}.
\newblock MIT Press, Cambridge, MA, USA, 2001.
\newblock ISBN 0262194759.

\bibitem[Shawe-Taylor and Cristianini(2004)]{Shawe-Taylor2004}
John Shawe-Taylor and Nello Cristianini.
\newblock \emph{Kernel Methods for Pattern Analysis}.
\newblock Cambridge University Press, New York, NY, USA, 2004.
\newblock ISBN 0521813972.

\bibitem[Sch{\"o}lkopf et~al.(1997)Sch{\"o}lkopf, Smola, and
  M{\"u}ller]{scholkopf1997kernel}
Bernhard Sch{\"o}lkopf, Alexander Smola, and Klaus-Robert M{\"u}ller.
\newblock Kernel principal component analysis.
\newblock In \emph{International Conference on Artificial Neural Networks},
  pages 583--588. Springer, 1997.

\bibitem[Mohri et~al.(2012)Mohri, Rostamizadeh, and Talwalkar]{Moh+2012}
M.~Mohri, A.~Rostamizadeh, and A.~Talwalkar.
\newblock \emph{Foundations of Machine Learning}.
\newblock MIT Press, 2012.

\bibitem[Cortes and Vapnik(1995)]{cortes1995support}
Corinna Cortes and Vladimir Vapnik.
\newblock Support-vector networks.
\newblock \emph{Machine learning}, 20\penalty0 (3):\penalty0 273--297, 1995.

\bibitem[Rasmussen(2004)]{rasmussen2004gaussian}
Carl~Edward Rasmussen.
\newblock Gaussian processes in machine learning.
\newblock In \emph{Advanced lectures on machine learning}, pages 63--71.
  Springer, 2004.

\bibitem[Silverman(1986)]{silverman2018density}
Bernard~W Silverman.
\newblock \emph{Density estimation for statistics and data analysis}.
\newblock Monographs on Statistics and Applied Probability. Chapman \& Hall,
  London, 1986.

\bibitem[Kriege et~al.(2014)Kriege, Neumann, Kersting, and Mutzel]{Kri+2014}
N.~Kriege, M.~Neumann, K.~Kersting, and M.~Mutzel.
\newblock Explicit versus implicit graph feature maps: {A} computational phase
  transition for walk kernels.
\newblock In \emph{{IEEE} International Conference on Data Mining}, pages
  881--886, 2014.

\bibitem[Kriege et~al.(2019)Kriege, Neumann, Morris, Kersting, and
  Mutzel]{Kriege2019dami}
Nils~M. Kriege, Marion Neumann, Christopher Morris, Kristian Kersting, and
  Petra Mutzel.
\newblock A unifying view of explicit and implicit feature maps of graph
  kernels.
\newblock \emph{Data Mining and Knowledge Discovery}, 33\penalty0 (6):\penalty0
  1505--1547, Nov 2019.
\newblock \doi{10.1007/s10618-019-00652-0}.

\bibitem[Johnson(2005)]{Johnson2005}
David~S. Johnson.
\newblock The {NP}-completeness column.
\newblock \emph{ACM Transctions on Algorithms}, 1\penalty0 (1):\penalty0
  160--176, July 2005.
\newblock ISSN 1549-6325.
\newblock \doi{10.1145/1077464.1077476}.

\bibitem[Haussler(1999)]{Hau1999}
D.~Haussler.
\newblock Convolution kernels on discrete structures.
\newblock Technical Report UCS-CRL-99-10, University of California at Santa
  Cruz, 1999.

\bibitem[Yanardag and Vishwanathan(2015{\natexlab{a}})]{Yan+2015}
P.~Yanardag and S.~V.~N. Vishwanathan.
\newblock Deep graph kernels.
\newblock In \emph{ACM SIGKDD International Conference on Knowledge Discovery
  and Data Mining}, pages 1365--1374, 2015{\natexlab{a}}.

\bibitem[Aiolli et~al.(2015)Aiolli, Donini, Navarin, and Sperduti]{Aio+2015}
F.~Aiolli, M.~Donini, N.~Navarin, and A.~Sperduti.
\newblock Multiple graph-kernel learning.
\newblock In \emph{{IEEE Symposium Series on Computational Intelligence}},
  pages 1607--1614, 2015.

\bibitem[Shin and Kuboyama(2008)]{shin2008generalization}
Kilho Shin and Tetsuji Kuboyama.
\newblock A generalization of haussler's convolution kernel: mapping kernel.
\newblock In \emph{International conference on Machine learning}, pages
  944--951. ACM, 2008.

\bibitem[G\"{a}rtner et~al.(2003)G\"{a}rtner, Flach, and Wrobel]{Gae+2003}
T.~G\"{a}rtner, P.~Flach, and S.~Wrobel.
\newblock On graph kernels: Hardness results and efficient alternatives.
\newblock In \emph{Learning Theory and Kernel Machines}, pages 129--143.
  Springer, 2003.

\bibitem[Kashima et~al.(2003)Kashima, Tsuda, and Inokuchi]{Kas+2003}
H.~Kashima, K.~Tsuda, and A.~Inokuchi.
\newblock Marginalized kernels between labeled graphs.
\newblock In \emph{International Conference on Machine Learning}, pages
  321--328, 2003.

\bibitem[Borgwardt and Kriegel(2005)]{Borgwardt2005}
K.~M. Borgwardt and H.-P. Kriegel.
\newblock Shortest-path kernels on graphs.
\newblock In \emph{IEEE International Conference on Data Mining}, pages 74--81,
  2005.

\bibitem[Hermansson et~al.(2015)Hermansson, Johansson, and Watanabe]{Her+2015}
L.~Hermansson, F.~D. Johansson, and O.~Watanabe.
\newblock Generalized shortest path kernel on graphs.
\newblock In \emph{Discovery Science: International Conference}, pages 78--85,
  2015.

\bibitem[Shervashidze et~al.(2009)Shervashidze, Vishwanathan, Petri, Mehlhorn,
  and Borgwardt]{She+2009}
N.~Shervashidze, S.~V.~N. Vishwanathan, T.~H. Petri, K.~Mehlhorn, and K.~M.
  Borgwardt.
\newblock Efficient graphlet kernels for large graph comparison.
\newblock In \emph{International Conference on Artificial Intelligence and
  Statistics}, pages 488--495, 2009.

\bibitem[Horv\'{a}th et~al.(2004)Horv\'{a}th, G\"{a}rtner, and
  Wrobel]{Hor+2004}
T.~Horv\'{a}th, T.~G\"{a}rtner, and S.~Wrobel.
\newblock Cyclic pattern kernels for predictive graph mining.
\newblock In \emph{ACM SIGKDD International Conference on Knowledge Discovery
  and Data Mining}, pages 158--167, 2004.

\bibitem[Ramon and G{\"a}rtner(2003)]{Ram+2003}
J.~Ramon and T.~G{\"a}rtner.
\newblock Expressivity versus efficiency of graph kernels.
\newblock In \emph{International Workshop on Mining Graphs, Trees and
  Sequences}, pages 65--74, 2003.

\bibitem[Mah\'{e} and Vert(2009)]{Mah+2009}
P.~Mah\'{e} and J.-P. Vert.
\newblock Graph kernels based on tree patterns for molecules.
\newblock \emph{Machine Learning}, 75\penalty0 (1):\penalty0 3--35, 2009.

\bibitem[Da~San~Martino et~al.(2012{\natexlab{a}})Da~San~Martino, Navarin, and
  Sperduti]{Mar+2012}
G.~Da~San~Martino, N.~Navarin, and A.~Sperduti.
\newblock A tree-based kernel for graphs.
\newblock In \emph{SIAM Conference of Data Mining}, pages 975--986,
  2012{\natexlab{a}}.

\bibitem[Da~San~Martino et~al.(2012{\natexlab{b}})Da~San~Martino, Navarin, and
  Sperduti]{Mar+2012a}
G.~Da~San~Martino, N.~Navarin, and A.~Sperduti.
\newblock A memory efficient graph kernel.
\newblock In \emph{{International Joint Conference on Neural Networks}}, pages
  1--7, 2012{\natexlab{b}}.

\bibitem[Feragen et~al.(2013)Feragen, Kasenburg, Petersen, Bruijne, and
  M.]{Fer+2013}
A.~Feragen, N.~Kasenburg, J.~Petersen, M.~D. Bruijne, and Borgwardt~K. M.
\newblock Scalable kernels for graphs with continuous attributes.
\newblock In \emph{Advances in Neural Information Processing Systems}, pages
  216--224, 2013.
\newblock Erratum available at
  \url{http://image.diku.dk/aasa/papers/graphkernels_nips_erratum.pdf}.

\bibitem[Orsini et~al.(2015)Orsini, Frasconi, and De~Raedt]{Ors+2015}
F.~Orsini, P.~Frasconi, and L.~De~Raedt.
\newblock Graph invariant kernels.
\newblock In \emph{International Joint Conference on Artificial Intelligence},
  pages 3756--3762, 2015.

\bibitem[Kriege and Mutzel(2012)]{Kri+2012}
N.~Kriege and P.~Mutzel.
\newblock Subgraph matching kernels for attributed graphs.
\newblock In \emph{International Conference on Machine Learning}, 2012.

\bibitem[Shervashidze et~al.(2011)Shervashidze, Schweitzer, van Leeuwen,
  Mehlhorn, and Borgwardt]{She+2011}
N.~Shervashidze, P.~Schweitzer, E.~J. van Leeuwen, K.~Mehlhorn, and K.~M.
  Borgwardt.
\newblock Weisfeiler-{L}ehman graph kernels.
\newblock \emph{Journal of Machine Learning Research}, 12:\penalty0 2539--2561,
  2011.

\bibitem[Morris et~al.(2017)Morris, Kersting, and Mutzel]{Mor+2017}
C.~Morris, K.~Kersting, and P.~Mutzel.
\newblock Glocalized {Weisfeiler-Lehman} kernel: Global-local feature maps of
  graphs.
\newblock In \emph{IEEE International Conference on Data Mining}, 2017.

\bibitem[Hido and Kashima(2009)]{Hid+2009}
S.~Hido and H.~Kashima.
\newblock A linear-time graph kernel.
\newblock In \emph{IEEE International Conference on Data Mining}, pages
  179--188, 2009.

\bibitem[Neumann et~al.(2016)Neumann, Garnett, Bauckhage, and
  Kersting]{Neu+2016}
M.~Neumann, R.~Garnett, C.~Bauckhage, and K.~Kersting.
\newblock Propagation kernels: {E}fficient graph kernels from propagated
  information.
\newblock \emph{Machine Learning}, 102\penalty0 (2):\penalty0 209--245, 2016.

\bibitem[Costa and De~Grave(2010)]{Cos+2010}
F.~Costa and K.~De~Grave.
\newblock Fast neighborhood subgraph pairwise distance kernel.
\newblock In \emph{International Conference on Machine Learning}, pages
  255--262. Omnipress, 2010.

\bibitem[Mah\'{e} et~al.(2004)Mah\'{e}, Ueda, Akutsu, Perret, and
  Vert]{Mah+2004}
P.~Mah\'{e}, N.~Ueda, T.~Akutsu, J.-L. Perret, and J.-P. Vert.
\newblock Extensions of marginalized graph kernels.
\newblock In \emph{International Conference on Machine Learning}, pages
  552--559, 2004.

\bibitem[Sugiyama and Borgwardt(2015)]{Sug+2015}
M.~Sugiyama and K.~M. Borgwardt.
\newblock Halting in random walk kernels.
\newblock In \emph{Advances in Neural Information Processing Systems}, pages
  1639--1647, 2015.

\bibitem[Kang et~al.(2012)Kang, Tong, and Sun]{Kan+2012}
U.~Kang, H.~Tong, and J.~Sun.
\newblock Fast random walk graph kernel.
\newblock In \emph{SIAM International Conference on Data Mining}, pages
  828--838, 2012.

\bibitem[Fr\"{o}hlich et~al.(2005)Fr\"{o}hlich, Wegner, Sieker, and
  Zell]{Fro+2005}
H.~Fr\"{o}hlich, J.~K. Wegner, F.~Sieker, and A.~Zell.
\newblock Optimal assignment kernels for attributed molecular graphs.
\newblock In \emph{International Conference on Machine learning}, pages
  225--232, 2005.

\bibitem[Kriege et~al.(2016)Kriege, Giscard, and Wilson]{Kri+2016}
N.~M. Kriege, P.-L. Giscard, and R.~C. Wilson.
\newblock On valid optimal assignment kernels and applications to graph
  classification.
\newblock In \emph{Advances in Neural Information Processing Systems}, pages
  1615--1623, 2016.

\bibitem[Nikolentzos et~al.(2017{\natexlab{a}})Nikolentzos, Meladianos, and
  Vazirgiannis]{Nik+2017}
G.~Nikolentzos, P.~Meladianos, and M.~Vazirgiannis.
\newblock Matching node embeddings for graph similarity.
\newblock In \emph{{AAAI Conference on Artificial Intelligence}}, pages
  2429--2435, 2017{\natexlab{a}}.

\bibitem[Johansson and Dubhashi(2015)]{Joh+2015}
F.~D. Johansson and D.~Dubhashi.
\newblock Learning with similarity functions on graphs using matchings of
  geometric embeddings.
\newblock In \emph{ACM SIGKDD International Conference on Knowledge Discovery
  and Data Mining}, pages 467--476, 2015.

\bibitem[Su et~al.(2016)Su, Han, Harang, and Yan]{Su+2016}
Y.~Su, F.~Han, R.~E. Harang, and X.~Yan.
\newblock A fast kernel for attributed graphs.
\newblock In \emph{SIAM International Conference on Data Mining}, pages
  486--494, 2016.

\bibitem[Kondor et~al.(2009)Kondor, Shervashidze, and Borgwardt]{Kon+2009}
R.~Kondor, N.~Shervashidze, and K.~M. Borgwardt.
\newblock The graphlet spectrum.
\newblock In \emph{International Conference on Machine Learning}, pages
  529--536, 2009.

\bibitem[Kondor and Pan(2016)]{Kon+2016}
R.~Kondor and H.~Pan.
\newblock The multiscale laplacian graph kernel.
\newblock In \emph{Advances in Neural Information Processing Systems}, pages
  2982--2990, 2016.

\bibitem[Johansson et~al.(2014)Johansson, Jethava, Dubhashi, and
  Bhattacharyya]{Joh+2014}
F.~D. Johansson, V.~Jethava, D.~P. Dubhashi, and C.~Bhattacharyya.
\newblock Global graph kernels using geometric embeddings.
\newblock In \emph{International Conference on Machine Learning}, pages
  694--702, 2014.

\bibitem[Yanardag and Vishwanathan(2015{\natexlab{b}})]{Yan+2015a}
P.~Yanardag and S.~V.~N. Vishwanathan.
\newblock A structural smoothing framework for robust graph comparison.
\newblock In \emph{Advances in Neural Information Processing Systems}, pages
  2134--2142, 2015{\natexlab{b}}.

\bibitem[Morris et~al.(2016)Morris, Kriege, Kersting, and Mutzel]{Mor+2016}
C.~Morris, N.~M. Kriege, K.~Kersting, and P.~Mutzel.
\newblock Faster kernel for graphs with continuous attributes via hashing.
\newblock In \emph{IEEE International Conference on Data Mining}, pages
  1095--1100, 2016.

\bibitem[Bai et~al.(2014)Bai, Ren, Bai, and Hancock]{bai2014graph}
Lu~Bai, Peng Ren, Xiao Bai, and Edwin~R Hancock.
\newblock A graph kernel from the depth-based representation.
\newblock In \emph{Joint IAPR International Workshops on Statistical Techniques
  in Pattern Recognition and Structural and Syntactic Pattern Recognition},
  pages 1--11, 2014.

\bibitem[Bai et~al.(2015)Bai, Rossi, Zhang, and Hancock]{Bai+2015}
Lu~Bai, Luca Rossi, Zhihong Zhang, and Edwin~R. Hancock.
\newblock An aligned subtree kernel for weighted graphs.
\newblock In \emph{International Conference on Machine Learning}, pages 30--39,
  2015.

\bibitem[Adamson and Bush(1973)]{Adamson1973}
George~W. Adamson and Judith~A. Bush.
\newblock A method for the automatic classification of chemical structures.
\newblock \emph{Information Storage and Retrieval}, 9\penalty0 (10):\penalty0
  561 -- 568, 1973.
\newblock ISSN 0020-0271.
\newblock \doi{http://dx.doi.org/10.1016/0020-0271(73)90059-4}.

\bibitem[Willett and Winterman(1986)]{Willett1986}
Peter Willett and Vivienne Winterman.
\newblock A comparison of some measures for the determination of
  inter-molecular structural similarity measures of inter-molecular structural
  similarity.
\newblock \emph{Quantitative Structure-Activity Relationships}, 5\penalty0
  (1):\penalty0 18--25, 1986.
\newblock ISSN 1521-3838.
\newblock \doi{10.1002/qsar.19860050105}.

\bibitem[Rogers and Hahn(2010)]{Rogers2010}
David Rogers and Mathew Hahn.
\newblock Extended-connectivity fingerprints.
\newblock \emph{Journal of Chemical Information and Modeling}, 50\penalty0
  (5):\penalty0 742--754, May 2010.
\newblock URL \url{http://dx.doi.org/10.1021/ci100050t}.

\bibitem[Ralaivola et~al.(2005)Ralaivola, Swamidass, Saigo, and
  Baldi]{Ralaivola2005}
Liva Ralaivola, Sanjay~Joshua Swamidass, Hiroto Saigo, and Pierre Baldi.
\newblock Graph kernels for chemical informatics.
\newblock \emph{Neural Networks}, 18\penalty0 (8):\penalty0 1093 -- 1110, 2005.
\newblock ISSN 0893-6080.
\newblock \doi{DOI: 10.1016/j.neunet.2005.07.009}.
\newblock {Neural Networks and Kernel Methods for Structured Domains}.

\bibitem[Merkwirth and Lengauer(2005)]{Mer+2005}
C.~Merkwirth and T.~Lengauer.
\newblock Automatic generation of complementary descriptors with molecular
  graph networks.
\newblock \emph{Journal of Chemical Information and Modeling}, 45\penalty0
  (5):\penalty0 1159--1168, 2005.

\bibitem[Duvenaud et~al.(2015)Duvenaud, Maclaurin, Iparraguirre, Bombarell,
  Hirzel, Aspuru-Guzik, and Adams]{Duv+2015}
D.~K. Duvenaud, D.~Maclaurin, J.~Iparraguirre, R.~Bombarell, T.~Hirzel,
  A.~Aspuru-Guzik, and R.~P. Adams.
\newblock Convolutional networks on graphs for learning molecular fingerprints.
\newblock In \emph{Advances in Neural Information Processing Systems}, pages
  2224--2232, 2015.

\bibitem[Kipf and Welling(2017)]{Kip+2017}
T.~N. Kipf and M.~Welling.
\newblock Semi-supervised classification with graph convolutional networks.
\newblock In \emph{International Conference on Learning Representations}, 2017.

\bibitem[Gilmer et~al.(2017)Gilmer, Schoenholz, Riley, Vinyals, and
  Dahl]{Gil+2017}
J.~Gilmer, S.~S. Schoenholz, P.~F. Riley, O.~Vinyals, and G.~E. Dahl.
\newblock Neural message passing for quantum chemistry.
\newblock In \emph{International Conference on Machine Learning}, volume~70 of
  \emph{Proceedings of Machine Learning Research}, pages 1263--1272. PMLR,
  2017.

\bibitem[Hamilton et~al.(2017)Hamilton, Ying, and Leskovec]{Ham+2017}
W.~L. Hamilton, R.~Ying, and J.~Leskovec.
\newblock Inductive representation learning on large graphs.
\newblock In \emph{Advances in Neural Information Processing Systems}, pages
  1025--1035, 2017.

\bibitem[Fey et~al.(2018)Fey, Lenssen, Weichert, and M{\"{u}}ller]{Fey2018}
Matthias Fey, Jan~Eric Lenssen, Frank Weichert, and Heinrich M{\"{u}}ller.
\newblock {SplineCNN}: Fast geometric deep learning with continuous b-spline
  kernels.
\newblock In \emph{{IEEE} Conference on Computer Vision and Pattern
  Recognition}, pages 869--877, 2018.

\bibitem[Morris et~al.(2019)Morris, Ritzert, Fey, Hamilton, Lenssen, Rattan,
  and Grohe]{Mor+2019}
C.~Morris, M.~Ritzert, M.~Fey, W.~L. Hamilton, Jan~Eric Lenssen, G.~Rattan, and
  M.~Grohe.
\newblock Weisfeiler and {Leman} go neural: Higher-order graph neural networks.
\newblock In \emph{AAAI Conference on Artificial Intelligence}, page TBD, 2019.

\bibitem[Babai and Kucera(1979)]{Bab+1979}
L.~Babai and L.~Kucera.
\newblock Canonical labelling of graphs in linear average time.
\newblock In \emph{Annual Symposium on Foundations of Computer Science}, pages
  39--46, 1979.

\bibitem[Vert(2008)]{Ver2008}
J.{-}P. Vert.
\newblock The optimal assignment kernel is not positive definite.
\newblock \emph{CoRR}, abs/0801.4061, 2008.
\newblock URL \url{http://arxiv.org/abs/0801.4061}.

\bibitem[Loosli et~al.(2015)Loosli, Canu, and Ong]{Loosli2015}
G.~Loosli, S.~Canu, and C.~S. Ong.
\newblock Learning svm in krein spaces.
\newblock \emph{IEEE Transactions on Pattern Analysis and Machine
  Intelligence}, PP\penalty0 (99):\penalty0 1--1, 2015.
\newblock ISSN 0162-8828.
\newblock \doi{10.1109/TPAMI.2015.2477830}.

\bibitem[Woźnica et~al.(2010)Woźnica, Kalousis, and Hilario]{Woznica2010}
Adam Woźnica, Alexandros Kalousis, and Melanie Hilario.
\newblock Adaptive matching based kernels for labelled graphs.
\newblock In \emph{Advances in Knowledge Discovery and Data Mining}, volume
  6119 of \emph{Lecture Notes in Computer Science}, pages 374--385. 2010.
\newblock \doi{10.1007/978-3-642-13672-6\_37}.

\bibitem[Ramon and Bruynooghe(2001)]{Ramon2001}
Jan Ramon and Maurice Bruynooghe.
\newblock A polynomial time computable metric between point sets.
\newblock \emph{Acta Informatica}, 37\penalty0 (10):\penalty0 765--780, Jul
  2001.
\newblock ISSN 1432-0525.
\newblock \doi{10.1007/PL00013304}.
\newblock URL \url{https://doi.org/10.1007/PL00013304}.

\bibitem[Balcan et~al.(2008)Balcan, Blum, and Srebro]{balcan2008theory}
Maria-Florina Balcan, Avrim Blum, and Nathan Srebro.
\newblock A theory of learning with similarity functions.
\newblock \emph{Machine Learning}, 72\penalty0 (1-2):\penalty0 89--112, 2008.

\bibitem[Kriege(2019)]{Kriege2019}
Nils~M. Kriege.
\newblock Deep {Weisfeiler-Lehman} assignment kernels via multiple kernel
  learning.
\newblock In \emph{27th European Symposium on Artificial Neural Networks,
  {ESANN} 2019}, 2019.

\bibitem[Pachauri et~al.(2013)Pachauri, Kondor, and Singh]{Pachauri2013}
Deepti Pachauri, Risi Kondor, and Vikas Singh.
\newblock Solving the multi-way matching problem by permutation
  synchronization.
\newblock In \emph{Advances in Neural Information Processing Systems}, pages
  1860--1868. 2013.

\bibitem[Schiavinato et~al.(2015)Schiavinato, Gasparetto, and
  Torsello]{Schiavinato2015}
Michele Schiavinato, Andrea Gasparetto, and Andrea Torsello.
\newblock Transitive assignment kernels for structural classification.
\newblock In \emph{Similarity-Based Pattern Recognition: Third International
  Workshop}, pages 146--159, 2015.
\newblock ISBN 978-3-319-24261-3.
\newblock \doi{10.1007/978-3-319-24261-3_12}.

\bibitem[Grauman and Darrell(2007{\natexlab{a}})]{grauman2007pyramid}
Kristen Grauman and Trevor Darrell.
\newblock The pyramid match kernel: Efficient learning with sets of features.
\newblock \emph{Journal of Machine Learning Research}, 8\penalty0
  (Apr):\penalty0 725--760, 2007{\natexlab{a}}.

\bibitem[Johansson et~al.(2015)Johansson, Frost, Retzner, and
  Dubhashi]{johansson2015classifying}
Fredrik~D Johansson, Otto Frost, Carl Retzner, and Devdatt Dubhashi.
\newblock Classifying large graphs with differential privacy.
\newblock In \emph{Modeling Decisions for Artificial Intelligence}, pages
  3--17. Springer, 2015.

\bibitem[Ahmed et~al.(2016)Ahmed, Willke, and Rossi]{Ahm+2016}
N.~K. Ahmed, T.~Willke, and R.~A. Rossi.
\newblock Estimation of local subgraph counts.
\newblock In \emph{IEEE International Conference on Big Data}, pages 1--10,
  2016.

\bibitem[Chen and Lui(2016)]{Che+2016}
X.~Chen and John C.~S. Lui.
\newblock Mining graphlet counts in online social networks.
\newblock In \emph{{IEEE} International Conference on Data Mining}, pages
  71--80, 2016.

\bibitem[Bressan et~al.(2017)Bressan, Chierichetti, Kumar, Leucci, and
  Panconesi]{Bre+2017}
M.~Bressan, F.~Chierichetti, R.~Kumar, S.~Leucci, and A.~Panconesi.
\newblock Counting graphlets: Space vs time.
\newblock In \emph{ACM International Conference on Web Search and Data Mining},
  pages 557--566, 2017.

\bibitem[Wale et~al.(2008)Wale, Watson, and Karypis]{Wale2008a}
Nikil Wale, Ian~A. Watson, and George Karypis.
\newblock Comparison of descriptor spaces for chemical compound retrieval and
  classification.
\newblock \emph{Knowledge Information Systems}, 14\penalty0 (3):\penalty0
  347--375, 2008.

\bibitem[McKay and Piperno(2014)]{McKay2014}
Brendan~D. McKay and Adolfo Piperno.
\newblock Practical graph isomorphism, {II}.
\newblock \emph{Journal of Symbolic Computation}, 60\penalty0 (0):\penalty0 94
  -- 112, 2014.
\newblock ISSN 0747-7171.
\newblock \doi{http://dx.doi.org/10.1016/j.jsc.2013.09.003}.

\bibitem[Mah{\'{e}} et~al.(2005)Mah{\'{e}}, Ueda, Akutsu, Perret, and
  Vert]{Mah+2005}
P.~Mah{\'{e}}, N.~Ueda, T.~Akutsu, J.-L. Perret, and J.-P. Vert.
\newblock Graph kernels for molecular structure-activity relationship analysis
  with support vector machines.
\newblock \emph{Journal of Chemical Information and Modeling}, 45\penalty0
  (4):\penalty0 939--951, 2005.

\bibitem[Borgwardt et~al.(2005)Borgwardt, Ong, Sch\"onauer, Vishwanathan,
  Smola, and Kriegel]{Bor+2005a}
K.~M. Borgwardt, C.~S. Ong, S.~Sch\"onauer, S.~V.~N. Vishwanathan, A.~J. Smola,
  and H.-P. Kriegel.
\newblock Protein function prediction via graph kernels.
\newblock \emph{Bioinformatics}, 21\penalty0 (Supplement 1):\penalty0 i47--i56,
  2005.

\bibitem[Harchaoui and Bach(2007)]{Har+2007}
Z.~Harchaoui and F.~Bach.
\newblock Image classification with segmentation graph kernels.
\newblock In \emph{IEEE Conference on Computer Vision and Pattern Recognition},
  pages 1--8, 2007.

\bibitem[Zhang et~al.(2018{\natexlab{b}})Zhang, Wang, Xiang, Huang, and
  Nehorai]{Zha+2018RetGK}
Z.~Zhang, M.~Wang, Y.~Xiang, Y.~Huang, and A.~Nehorai.
\newblock Retgk: Graph kernels based on return probabilities of random walks.
\newblock In \emph{Advances in Neural Information Processing Systems}, pages
  3964--3974, 2018{\natexlab{b}}.

\bibitem[Grauman and Darrell(2007{\natexlab{b}})]{Gra+2006}
K.~Grauman and T.~Darrell.
\newblock Approximate correspondences in high dimensions.
\newblock In \emph{Advances in Neural Information Processing Systems}, pages
  505--512, 2007{\natexlab{b}}.

\bibitem[Lovász(2006)]{Lov+1979}
L.~Lovász.
\newblock On the shannon capacity of a graph.
\newblock \emph{IEEE Transactions on Information Theory}, 25\penalty0
  (1):\penalty0 1--7, 2006.

\bibitem[Li et~al.(2015)Li, Tong, Xiao, and Fan]{Li+2015}
L.~Li, H.~Tong, Y.~Xiao, and W.~Fan.
\newblock \emph{Cheetah}: Fast graph kernel tracking on dynamic graphs.
\newblock In \emph{{SIAM} International Conference on Data Mining}, pages
  280--288, 2015.

\bibitem[Li et~al.(2012)Li, Zhu, Chi, and Zhang]{Li+2012}
B.~Li, X.~Zhu, L.~Chi, and C.~Zhang.
\newblock Nested subtree hash kernels for large-scale graph classification over
  streams.
\newblock In \emph{IEEE International Conference on Data Mining}, pages
  399--408, 2012.

\bibitem[Massimo et~al.(2016)Massimo, Navarin, and Sperduti]{Massimo2016}
Carlo~M. Massimo, Nicol{\`o} Navarin, and Alessandro Sperduti.
\newblock Hyper-parameter tuning for graph kernels via multiple kernel
  learning.
\newblock In \emph{Advances in Neural Information Processing}, pages 214--223,
  2016.

\bibitem[Nikolentzos et~al.(2018)Nikolentzos, Meladianos, Limnios, and
  Vazirgiannis]{Nikolentzos2018}
Giannis Nikolentzos, Polykarpos Meladianos, Stratis Limnios, and Michalis
  Vazirgiannis.
\newblock A degeneracy framework for graph similarity.
\newblock In \emph{{International Joint Conference on Artificial Intelligenc}},
  pages 2595--2601. ijcai.org, 2018.

\bibitem[Mikolov et~al.(2013)Mikolov, Chen, Corrado, and Dean]{Mik+2013}
T.~Mikolov, K.~Chen, G.~Corrado, and J.~Dean.
\newblock Efficient estimation of word representations in vector space.
\newblock \emph{CoRR}, abs/1301.3781, 2013.

\bibitem[Kriege et~al.(2018)Kriege, Morris, Rey, and Sohler]{Kri+2018}
Nils~M. Kriege, Christopher Morris, Anja Rey, and Christian Sohler.
\newblock A property testing framework for the theoretical expressivity of
  graph kernels.
\newblock In \emph{International Joint Conference on Artificial Intelligence},
  pages 2348--2354, 2018.
\newblock \doi{10.24963/ijcai.2018/325}.

\bibitem[Oneto et~al.(2017)Oneto, Navarin, Donini, Sperduti, Aiolli, and
  Anguita]{One+2017}
L.~Oneto, N.~Navarin, M.~Donini, A.~Sperduti, F.~Aiolli, and D.~Anguita.
\newblock Measuring the expressivity of graph kernels through statistical
  learning theory.
\newblock \emph{Neurocomputing}, 268\penalty0 (Supplement C):\penalty0 4--16,
  2017.

\bibitem[Dwork et~al.(2014)Dwork, Roth, et~al.]{dwork2014algorithmic}
Cynthia Dwork, Aaron Roth, et~al.
\newblock The algorithmic foundations of differential privacy.
\newblock \emph{Foundations and Trends{\textregistered} in Theoretical Computer
  Science}, 9\penalty0 (3--4):\penalty0 211--407, 2014.

\bibitem[Brown(2009)]{Brown2009}
Nathan Brown.
\newblock Chemoinformatics -- an introduction for computer scientists.
\newblock \emph{ACM Computing Surveys}, 41\penalty0 (2), 2009.

\bibitem[Horv\'{a}th et~al.(2010)Horv\'{a}th, Ramon, and Wrobel]{Horvath2010a}
Tam\'{a}s Horv\'{a}th, Jan Ramon, and Stefan Wrobel.
\newblock Frequent subgraph mining in outerplanar graphs.
\newblock \emph{Data Mining and Knowledge Discovery}, 21:\penalty0 472--508,
  2010.
\newblock ISSN 1384-5810.
\newblock \doi{10.1007/s10618-009-0162-1}.

\bibitem[Yamaguchi et~al.(2003)Yamaguchi, Aoki, and Mamitsuka]{Yamaguchi2003a}
Atsuko Yamaguchi, Kiyoko~F. Aoki, and Hiroshi Mamitsuka.
\newblock Graph complexity of chemical compounds in biological pathways.
\newblock \emph{Genome Informatics}, 14:\penalty0 376--377, 2003.

\bibitem[Swamidass et~al.(2005)Swamidass, Chen, Bruand, Phung, Ralaivola, and
  Baldi]{Swa+2005}
S.~Joshua Swamidass, Jonathan Chen, Jocelyne Bruand, Peter Phung, Liva
  Ralaivola, and Pierre Baldi.
\newblock Kernels for small molecules and the prediction of mutagenicity,
  toxicity and anti-cancer activity.
\newblock \emph{Bioinformatics}, 21\penalty0 (Suppl 1):\penalty0 i359--i368,
  2005.

\bibitem[Ceroni et~al.(2007)Ceroni, Costa, and Frasconi]{Ceroni2007}
Alessio Ceroni, Fabrizio Costa, and Paolo Frasconi.
\newblock Classification of small molecules by two- and three-dimensional
  decomposition kernels.
\newblock \emph{Bioinformatics}, 23\penalty0 (16):\penalty0 2038--2045, Aug
  2007.
\newblock URL \url{http://dx.doi.org/10.1093/bioinformatics/btm298}.

\bibitem[Mah\'{e} et~al.(2006)Mah\'{e}, Ralaivola, Stoven, and Vert]{Mah+2006}
P.~Mah\'{e}, L.~Ralaivola, V.~Stoven, and J.-P. Vert.
\newblock The pharmacophore kernel for virtual screening with support vector
  machines.
\newblock \emph{Journal of Chemical Information and Modeling}, 46\penalty0
  (5):\penalty0 2003--2014, 2006.

\bibitem[Daylight(2008)]{Daylight2008}
Chemical Information~Systems Daylight.
\newblock Daylight theory manual v4.9.
\newblock http://www.daylight.com/dayhtml/doc/theory, January 2008.

\bibitem[Durant et~al.(2002)Durant, Leland, Henry, and Nourse]{Durant2002}
Joseph~L. Durant, Burton~A. Leland, Douglas~R. Henry, and James~G. Nourse.
\newblock Reoptimization of mdl keys for use in drug discovery.
\newblock \emph{Journal of Chemical Information and Computer Sciences},
  42\penalty0 (5):\penalty0 1273--1280, 2002.

\bibitem[Borgwardt et~al.(2007)Borgwardt, Kriegel, Vishwanathan, and
  Schraudolphs]{borgwardt2007graph}
Karsten~M Borgwardt, Hans-Peter Kriegel, SVN Vishwanathan, and Nicol~N
  Schraudolphs.
\newblock Graph kernels for disease outcome prediction from protein-protein
  interaction networks.
\newblock In \emph{Biocomputing 2007}, pages 4--15. World Scientific, 2007.

\bibitem[{Vega-Pons} et~al.(2014){Vega-Pons}, {Avesani}, {Andric}, and
  {Hasson}]{Vega-Pons2014}
S.~{Vega-Pons}, P.~{Avesani}, M.~{Andric}, and U.~{Hasson}.
\newblock Classification of inter-subject fmri data based on graph kernels.
\newblock In \emph{International Workshop on Pattern Recognition in
  Neuroimaging}, pages 1--4, June 2014.
\newblock \doi{10.1109/PRNI.2014.6858549}.

\bibitem[Takerkart et~al.(2014)Takerkart, Auzias, Thirion, and
  Ralaivola]{Takerkart2014}
Sylvain Takerkart, Guillaume Auzias, Bertrand Thirion, and Liva Ralaivola.
\newblock Graph-based inter-subject pattern analysis of fmri data.
\newblock \emph{PLOS ONE}, 9\penalty0 (8):\penalty0 1--14, 08 2014.
\newblock \doi{10.1371/journal.pone.0104586}.

\bibitem[Vega-Pons and Avesani(2013)]{Vega-Pons2013}
Sandro Vega-Pons and Paolo Avesani.
\newblock Brain decoding via graph kernels.
\newblock In \emph{Proceedings of the 2013 International Workshop on Pattern
  Recognition in Neuroimaging}, PRNI '13, pages 136--139, Washington, DC, USA,
  2013. IEEE Computer Society.
\newblock ISBN 978-0-7695-5061-9.
\newblock \doi{10.1109/PRNI.2013.43}.
\newblock URL \url{http://dx.doi.org/10.1109/PRNI.2013.43}.

\bibitem[Wang et~al.(2016)Wang, Wilson, and Hancock]{Wang2016}
Jianjia Wang, Richard~C. Wilson, and Edwin~R. Hancock.
\newblock fmri activation network analysis using bose-einstein entropy.
\newblock In Antonio Robles-Kelly, Marco Loog, Battista Biggio, Francisco
  Escolano, and Richard Wilson, editors, \emph{Structural, Syntactic, and
  Statistical Pattern Recognition}, pages 218--228, Cham, 2016. Springer
  International Publishing.
\newblock ISBN 978-3-319-49055-7.

\bibitem[Jie et~al.(2016)Jie, Liu, Jiang, and Zhang]{Jie2016}
Biao Jie, Mingxia Liu, Xi~Jiang, and Daoqiang Zhang.
\newblock Sub-network based kernels for brain network classification.
\newblock In \emph{ACM International Conference on Bioinformatics,
  Computational Biology, and Health Informatics}, pages 622--629, 2016.
\newblock ISBN 978-1-4503-4225-4.
\newblock \doi{10.1145/2975167.2985687}.

\bibitem[Nikolentzos et~al.(2017{\natexlab{b}})Nikolentzos, Meladianos,
  Rousseau, Stavrakas, and Vazirgiannis]{Nik+2017a}
G.~Nikolentzos, P.~Meladianos, F.~Rousseau, Y.~Stavrakas, and M.~Vazirgiannis.
\newblock Shortest-path graph kernels for document similarity.
\newblock In \emph{Empirical Methods in Natural Language Processing}, pages
  1890--1900, 2017{\natexlab{b}}.

\bibitem[Hermansson et~al.(2013)Hermansson, Kerola, Johansson, Jethava, and
  Dubhashi]{hermansson2013entity}
Linus Hermansson, Tommi Kerola, Fredrik Johansson, Vinay Jethava, and Devdatt
  Dubhashi.
\newblock Entity disambiguation in anonymized graphs using graph kernels.
\newblock In \emph{ACM International Conference on Information \& Knowledge
  Management}, pages 1037--1046, 2013.

\bibitem[Li et~al.(2016)Li, Saidi, Sanchez, Sch{\"{a}}f, and
  Schweitzer]{Li+2016}
W.~Li, H.~Saidi, H.~Sanchez, M.~Sch{\"{a}}f, and P.~Schweitzer.
\newblock Detecting similar programs via the {Weisfeiler-Leman} graph kernel.
\newblock In \emph{{International Conference on Software Reuse}}, pages
  315--330, 2016.

\bibitem[de~Vries(2013)]{Vri+2013}
Gerben K.~D. de~Vries.
\newblock A fast approximation of the {Weisfeiler-Lehman} graph kernel for rdf
  data.
\newblock In \emph{European Conference on Machine Learning \& Principles and
  Practice of Knowledge Discovery in Databases}, pages 606--621, 2013.

\bibitem[Wu et~al.(2014)Wu, Yuan, and Hu]{Wu2014}
B.~Wu, C.~Yuan, and W.~Hu.
\newblock Human action recognition based on context-dependent graph kernels.
\newblock In \emph{IEEE Conference on Computer Vision and Pattern Recognition},
  pages 2609--2616, June 2014.
\newblock \doi{10.1109/CVPR.2014.334}.

\bibitem[Neumann et~al.(2013)Neumann, Moreno, Antanas, Garnett, and
  Kersting]{Neumann2013a}
M.~Neumann, P.~Moreno, L.~Antanas, R.~Garnett, and K.~Kersting.
\newblock Graph kernels for object category prediction in task{--}dependent
  robot grasping.
\newblock In L.~Adamic, L.~Getoor, B.~Huang, J.~Leskovec, and J.~McAuley,
  editors, \emph{Working Notes of the International Workshop on Mining and
  Learning with Graphs at KDD 2013}, Chicago, IL, USA, August 11 2013.

\bibitem[Chang and Lin(2011)]{Cha+2011}
C.-C. Chang and C.-J. Lin.
\newblock {LIBSVM:} {A} library for support vector machines.
\newblock \emph{ACM Transactions on Intelligent Systems and Technology},
  2\penalty0 (3):\penalty0 27:1--27:27, 2011.

\bibitem[Nikolentzos and Vazirgiannis(2018)]{Nikolentzos2018a}
Giannis Nikolentzos and Michalis Vazirgiannis.
\newblock Enhancing graph kernels via successive embeddings.
\newblock In \emph{ACM International Conference on Information and Knowledge
  Management}, pages 1583--1586, 2018.
\newblock \doi{10.1145/3269206.3269289}.

\bibitem[Kersting et~al.(2016)Kersting, Kriege, Morris, Mutzel, and
  Neumann]{KKMMN2016}
K.~Kersting, N.~M. Kriege, C.~Morris, P.~Mutzel, and M.~Neumann.
\newblock Benchmark data sets for graph kernels, 2016.
\newblock URL \url{http://graphkernels.cs.tu-dortmund.de}.

\bibitem[Riesen and Bunke(2008)]{Riesen2008}
Kaspar Riesen and Horst Bunke.
\newblock {IAM} graph database repository for graph based pattern recognition
  and machine learning.
\newblock In \emph{Structural, Syntactic, and Statistical Pattern Recognition:
  Joint IAPR International Workshop}, pages 287--297, 2008.
\newblock ISBN 978-3-540-89689-0.
\newblock \doi{10.1007/978-3-540-89689-0\_33}.

\bibitem[Sutherland et~al.(2003)Sutherland, O'Brien, and
  Weaver]{Sutherland2003}
Jeffrey~J. Sutherland, Lee~A. O'Brien, and Donald~F. Weaver.
\newblock Spline-fitting with a genetic algorithm: a method for developing
  classification structure-activity relationships.
\newblock \emph{J Chem Inf Comput Sci}, 43\penalty0 (6):\penalty0 1906--1915,
  2003.
\newblock \doi{10.1021/ci034143r}.

\bibitem[Dobson and Doig(2003)]{Dob+2003}
P.~D. Dobson and A.~J. Doig.
\newblock Distinguishing enzyme structures from non-enzymes without alignments.
\newblock \emph{Journal of Molecular Biology}, 330\penalty0 (4):\penalty0
  771--783, 2003.

\bibitem[Schomburg et~al.(2004)Schomburg, Chang, Ebeling, Gremse, Heldt, Huhn,
  and Schomburg]{Schomburg2004}
Ida Schomburg, Antje Chang, Christian Ebeling, Marion Gremse, Christian Heldt,
  Gregor Huhn, and Dietmar Schomburg.
\newblock Brenda, the enzyme database: updates and major new developments.
\newblock \emph{Nucleic Acids Research}, 32\penalty0 (Database-Issue):\penalty0
  431--433, 2004.
\newblock \doi{10.1093/nar/gkh081}.

\bibitem[Kazius et~al.(2005)Kazius, Mc{G}uire, and Bursi]{Kaz+2005}
J.~Kazius, R.~Mc{G}uire, and R.~Bursi.
\newblock Derivation and validation of toxicophores for mutagenicity
  prediction.
\newblock \emph{Journal Medicinal Chemistry}, 48\penalty0 (13):\penalty0
  312--320, 2005.

\bibitem[Debnath et~al.(1991)Debnath, Lopez~de Compadre, Debnath, Shusterman,
  and Hansch]{Deb+1991}
A.~K. Debnath, R.~L. Lopez~de Compadre, G.~Debnath, A.~J. Shusterman, and
  C.~Hansch.
\newblock Structure-activity relationship of mutagenic aromatic and
  heteroaromatic nitro compounds. correlation with molecular orbital energies
  and hydrophobicity.
\newblock \emph{Journal of Medicinal Chemistry}, 34\penalty0 (2):\penalty0
  786--797, 1991.

\bibitem[Helma et~al.(2001)Helma, King, Kramer, and Srinivasan]{Helma2001}
C.~Helma, R.~D. King, S.~Kramer, and A.~Srinivasan.
\newblock The predictive toxicology challenge 2000–2001.
\newblock \emph{Bioinformatics}, 17\penalty0 (1):\penalty0 107--108, 2001.
\newblock \doi{10.1093/bioinformatics/17.1.107}.

\bibitem[Tox21 Data Challenge()]{Tox21}
Tox21 Data Challenge, 2014.
\newblock URL \url{https://tripod.nih.gov/tox21/challenge/data.jsp}.

\bibitem[Nikolentzos(2016)]{Nik+2017code}
G~Nikolentzos.
\newblock Pyramid match kernel.
\newblock \url{http://www.db-net.aueb.gr/nikolentzos/code/matchingnodes.zip},
  2016.

\bibitem[Yanardag(2015)]{Yan+2015code}
P~Yanardag.
\newblock Deep graph kernels (code).
\newblock \url{http://www.mit.edu/~pinary/kdd/DEEP_GRAPH_KERNELS_CODE.tar.gz},
  2015.

\bibitem[Neumann(2016)]{Neu+2016code}
M.~Neumann.
\newblock Propagation kernel (code).
\newblock \url{https://github.com/marionmari/propagation_kernels.git}, 2016.

\bibitem[Maaten and Hinton(2008)]{maaten2008visualizing}
Laurens van~der Maaten and Geoffrey Hinton.
\newblock Visualizing data using t-sne.
\newblock \emph{Journal of machine learning research}, 9\penalty0
  (Nov):\penalty0 2579--2605, 2008.

\bibitem[Fleiss(1971)]{fleiss1971measuring}
Joseph~L Fleiss.
\newblock Measuring nominal scale agreement among many raters.
\newblock \emph{Psychological bulletin}, 76\penalty0 (5):\penalty0 378, 1971.

\bibitem[Alon and Spencer(2004)]{alon2004probabilistic}
Noga Alon and Joel~H Spencer.
\newblock \emph{The probabilistic method}.
\newblock John Wiley \& Sons, 2004.

\bibitem[Fan et~al.(2008)Fan, Chang, Hsieh, Wang, and Lin]{Fan2008}
Rong-En Fan, Kai-Wei Chang, Cho-Jui Hsieh, Xiang-Rui Wang, and Chih-Jen Lin.
\newblock Liblinear: A library for large linear classification.
\newblock \emph{J. Mach. Learn. Res.}, 9:\penalty0 1871--1874, June 2008.
\newblock ISSN 1532-4435.

\end{thebibliography}

\end{document}